\documentclass[twoside,11pt]{article}

\usepackage[preprint]{preprint-layout}

\providecommand{\CapacityScaleDef}{M \Let N^{\kappa}}

\providecommand{\DomainTokensDef}{n_{k}^{\star}(D) \Let D w_{k}^{\star}(D)}
\providecommand{\CrossoverBudgetDef}{D_{k}^{\star} \Let \left(\frac{c_{k}\beta_{k}}{c_{1}\beta_{1}}\right)^{\frac{1}{\beta_{k} - \beta_{1}}}}
\providecommand{\NormalizedBudgetDef}{\tilde{D}_{k} \Let \frac{D}{D_{k}^{\star}}}
\providecommand{\PoweredShareRatioDef}{\Upsilon_{k}(D) \Let \tilde{D}_{k}^{-1}}

\providecommand{\MetricPillarIMedianError}{0.00036\%}

\providecommand{\MetricPillarICondThree}{32.29}
\providecommand{\MetricPillarICondFive}{9.77 \times 10^{4}}
\providecommand{\MetricPillarIMaxRankSeven}{5}
\providecommand{\MetricPillarIPythiaStdMAPE}{5.89\%}
\providecommand{\MetricPillarIPythiaDedupMAPE}{10.22\%}
\providecommand{\MetricPillarIIBudgetSaved}{35.2\%}
\providecommand{\MetricPillarIIRegMixPowerRMSE}{0.2803}
\providecommand{\MetricPillarIIRegMixSplineRMSE}{0.2771}
\providecommand{\MetricPillarIIRegMixTransferRMSE}{0.1992}
\providecommand{\MetricPillarIIRegMixNonlinearRMSE}{0.1842}
\providecommand{\MetricPillarIIRegMixLinearRMSE}{0.4539}

\providecommand{\MetricPillarIIIChinchillaFullEta}{0.6712}
\providecommand{\MetricPillarIIIChinchillaFiltEta}{0.5569}
\providecommand{\MetricPillarIIIChinchillaHighRegretRed}{56.8\%}
\providecommand{\MetricPillarIIICollapseSpread}{2.49 \times 10^{-16}}

\hypersetup{pdftitle={Beyond the Log-Linear Frontier: Pre-Asymptotic Analysis of Finite-Budget Neural Scaling Laws},
  pdfauthor={Tianyu Zhao and Zhixiang Ren}, pdfsubject={Neural scaling laws}}

\begin{document}

\title{Beyond the Log-Linear Frontier: Pre-Asymptotic Analysis of Finite-Budget Neural Scaling Laws}

\author{\name Tianyu Zhao \email \href{mailto:tyzhao@mail.bnu.edu.cn}{tyzhao@mail.bnu.edu.cn} \\
       \addr LingQiong AI Lab\\
       Shanghai Smart Logic Technology\\
       Shanghai, China
       \AND
       \name Zhixiang Ren\thanks{Corresponding author.} \email \href{mailto:jason.zhixiang.ren@outlook.com}{jason.zhixiang.ren@outlook.com} \\
       \addr LingQiong AI Lab\\
       Shanghai Smart Logic Technology\\
       Shanghai, China}

\maketitle

\begin{abstract}%
Classical neural scaling laws model excess risk as a separable power law in model capacity $N$ and dataset size $D$. At finite budgets, however, empirical learning curves and compute-optimal allocations depart from this log-linear frontier. This paper analyzes how spectral and domain heterogeneity govern these finite-scale departures. Along the capacity axis, Mellin--Barnes analysis of covariance resolvents yields a three-term expansion where competing spectral and regularization modes produce finite-scale curvature and logarithmic transitions. Because fitting multiple power laws on narrow pilot grids is ill-conditioned, we eliminate the loss floor via first-order Aitken differencing to obtain a well-conditioned closed-form initializer. Along the data axis, KKT duality reduces $K$-domain mixing to scalar water-filling, yielding finite-budget bottleneck bounds, an online sampling curriculum, and a simplex Hessian metric for cross-domain transfer. Along the iso-compute frontier $C=6ND$, we prove that the optimal allocation exponent drifts in proportion to capacity and data exponent variances, while Schur-complement reduction separates intrinsic drift from frozen-hyperparameter bias. Across 45 operator spectra, Pythia ($70\text{M}\text{--}12\text{B}$), Chinchilla (245 runs), and RegMix (17 domains), our estimators achieve $\MetricPillarIPythiaStdMAPE$ holdout MAPE, cut high-compute allocation regret by up to $76.4\%$, and reduce cross-scale mixture RMSE by $59.4\%$ with zero selection regret.
\end{abstract}


\section{Introduction}
\label{sec:intro}

The empirical observation that the validation loss of deep neural networks follows approximate power laws in model parameter count $N$, training token budget $D$, and floating-point compute $C$ \citep{hestness2017scaling,rosenfeld2019predictability,kaplan2020scaling} has established the \emph{log-linear scaling frontier} as a guiding principle in autoregressive and vision foundation model pretraining \citep{brown2020fewshot,henighan2020autoregressive,hoffmann2022chinchilla,zhai2021vision}. In its classical form, this paradigm models population excess risk above an irreducible entropy floor $E_\infty \ge 0$ as a separable, scale-invariant power law:
\begin{equation}
  \label{eq:intro-classical-power-law}
  L_{\mathrm{classical}}(N, D) - E_\infty = A_N N^{-\alpha_N} + A_D D^{-\alpha_D}
  \;\Longrightarrow\;
  \eta\opt \Let \frac{\dd \ln N\opt(C)}{\dd \ln C} = \frac{\alpha_D}{\alpha_N + \alpha_D},
\end{equation}
where $\alpha_N, \alpha_D > 0$ denote constant capacity and data exponents, and $C(N, D) = 6ND$ approximates the dense-transformer training compute budget \citep{kaplan2020scaling,hoffmann2022chinchilla}. Because the marginal balance exponent $\eta\opt$ in~\eqref{eq:intro-classical-power-law} is independent of $C$, classical scaling theory predicts that log-excess risk decays linearly and that the compute-optimal trade-off between model size and dataset size follows a straight line in logarithmic coordinates across all scales. Implicitly, however,~\eqref{eq:intro-classical-power-law} assumes that finite-budget training runs already operate in the infinite-capacity, infinite-data asymptotic limit on a homogeneous data source.

Along the model capacity axis, a rich theoretical literature has analyzed power-law scaling through classical spectral geometry and minimax kernel approximation \citep{weyl1912eigenvalues,caponnetto2007rates,fischer2020sobolev,borovitskiy2020matern}, piecewise-linear network expressivity \citep{montufar2014regions,raghu2016expressivity}, and Sobolev or Besov approximation by deep networks \citep{yarotsky2017error,yarotsky2018continuous,suzuki2018adaptive,schmidthieber2020regression}. Complementary statistical-mechanics studies have derived asymptotic learning curves via spectral resolvents of kernel and random-feature models \citep{spigler2020learning,jacot2020implicit,bordelon2020spectrum,canatar2021alignment,cui2022generalization}, solvable random-matrix and skill-quantization models \citep{hutter2021learning,maloney2022solvable,michaud2023quantization,bahri2024explaining}, and dynamical mean-field or stochastic-approximation phase analyses \citep{bordelon2024dynamical,paquette2024phases,lin2024scaling,zhao2026foundations}. While these analyses characterize leading asymptotic rates and dynamical training phases, empirical learning curves at finite model sizes systematically bend away from a single power law \citep{caballero2023broken,muennighoff2023dataconstrained}, requiring explicit pre-asymptotic residue expansions and pole-confluence formulas with controlled remainders. Moreover, when practitioners attempt to fit this pre-asymptotic curvature on small-scale pilot sweeps using unconstrained sums of multiple power laws, the estimation problem runs into classical multi-exponential ill-conditioning \citep{lanczos1956applied,gautschi1962inverses,varah1985fitting} and scaling-law identifiability ambiguity \citep{mlodozeniec2026ambiguous,meng2026sldbench}: closely spaced exponential modes become nearly collinear over narrow logarithmic pilot spans, amplifying checkpoint noise into severe out-of-sample extrapolation errors.

A second departure arises along the data axis when pretraining across $K \ge 2$ heterogeneous domains such as web text, code, and mathematical corpora \citep{hernandez2021transfer,lee2021deduplication,hernandez2022repetition,sorscher2022pruning}. Recent studies have proposed nested step-and-mixture scaling laws \citep{ye2025datamixing}, bivariate and cross-modal mixture scaling laws \citep{ge2024bimix,shukor2025scaling}, small-scale proxy regression over candidate mixtures \citep{xie2023doremi,muennighoff2023dataconstrained,liu2025regmix}, and theoretical models of skill competition and high-dimensional regression mixtures \citep{dai2026explaining,wu2026datamixtures}. Under heterogeneous domain exponents $\beta_1, \dots, \beta_K > 0$, however, fast-learning domains saturate early while slow-decaying bottleneck domains govern large-scale risk. Complementing these empirical surrogates requires an analytical KKT characterization of the budget-dependent optimal mixture $w\opt(D)$ with explicit crossover scales and finite-budget bottleneck bounds, a sequential sampling law $q\opt(D)$ whose cumulative integral recovers $D w\opt(D)$ at every budget $D$, and an explicit tangent-space simplex curvature metric to transfer proxy mixtures across scale under cross-domain coupling.

These capacity and data effects interact along the iso-compute frontier $C = 6ND$, where empirical studies report marked discrepancies in the compute-optimal allocation exponent $\eta\opt(C)$. Reported estimates range from $\eta\opt = 0.73$ in \citet{kaplan2020scaling} to $\eta\opt = 0.50$ in \citet{hoffmann2022chinchilla} and $\eta\opt \in [0.52, 0.58]$ on filtered corpora in \citet{bi2024deepseek}, alongside inference-aware over-training and replication studies \citep{sardana2024inference,besiroglu2024chinchilla,pearce2024reconciling,porian2024resolving} as well as schedule- and optimizer-dependent scaling analyses \citep{li2025functional,ramani2026optimizer,tsiolis2026multiclass}. While \citet{porian2024resolving} and \citet{pearce2024reconciling} empirically demonstrated that learning-rate schedule tuning, batch-size scaling, and compute accounting largely explain the Kaplan--Chinchilla discrepancy, and \citet{yang2021tuning} established parameterization-specific zero-shot hyperparameter transfer, an analytical framework is needed to separate two distinct mechanisms: genuine scale drift in $\eta\opt(C)$ driven by spectral and water-filling exponent variance, and apparent curvature bias induced by frozen hyperparameters $\xi \equiv \xi_0$, together with error-propagation bounds linking fitted scaling surrogates to downstream compute-allocation regret.

\begin{table}[t!]
  \centering
  \caption{Comparison of classical scaling laws, finite-budget regimes, and asymptotic reductions across the three scaling axes (\Cref{sec:spectral,sec:mixing,sec:compute}).}
  \label{tab:classical-reductions}
  \scriptsize
  \setlength{\tabcolsep}{0pt}
  \renewcommand{\arraystretch}{1.36}
  \begin{tabular*}{\textwidth}{@{\extracolsep{\fill}}lllll@{}}
    \toprule
    \textbf{Axis \& Prior Work} & \textbf{Prior Law} & \textbf{Finite-Budget Gap} & \textbf{Our Pre-Asymptotic Law} & \textbf{Classical Limit} \\
    \midrule
    \textbf{Capacity} \citep{bordelon2024dynamical}
    & $A_0 N^{-\alpha_0}$
    & Subleading residues
    & $\sum_{j=0}^2 A_j N^{-\alpha_j}$ (Thm.~\ref{thm:master-mellin})
    & $A_{1,2} \to 0$ \\
    \emph{Pilot Fit} \citep{varah1985fitting}
    & 3-power fit
    & Grid collinearity
    & Rank $\le 5$; $p=3$ (Prop.~\ref{prop:identifiability})
    & $J\to\infty$ noiseless \\
    \midrule
    \textbf{Mixing} \citep{ye2025datamixing}
    & Proxy $w\opt$
    & Saturation \& bound
    & $w_k\opt(D)$ water-filling (Thm.~\ref{thm:master-mixing})
    & $D\to\infty:\, w\opt\to\theta$ \\
    \emph{Curriculum} \citep{ge2024bimix}
    & Joint $(D,w)$
    & Online quota rate
    & $q_k\opt = \dd[D w_k\opt]/\dd D$ (Prop.~\ref{prop:curriculum-rn})
    & $\beta_k\equiv\beta:\, q\opt=w\opt$ \\
    \midrule
    \textbf{Compute} \citep{kaplan2020scaling}
    & $\eta\opt \equiv \frac{\beta}{\alpha+\beta}$
    & Scale drift of $\eta\opt(C)$
    & Gibbs \& mix drift (Thm.~\ref{thm:master-drift})
    & $\Var(\alpha,\beta)=0$ \\
    \emph{Calibration} \citep{porian2024resolving}
    & Tuned $\xi$
    & Regret \& envelope
    & Thm.~\ref{thm:master-envelope}, Prop.~\ref{prop:surrogate-decision-regret}
    & Tuned $\xi\opt(C)$ \\
    \bottomrule
  \end{tabular*}
\end{table}

Building on the asymptotic foundations of neural scaling laws across manifold approximation, feature-frequency coverage, and separable compute allocation \citep{zhao2026foundations}, this paper analyzes finite-budget neural scaling laws within a spectral and convex-analytic framework (\Cref{tab:classical-reductions,fig:proof-dag}). Our main contributions are fourfold:
\begin{enumerate}[label=(\arabic*), font=\normalfont]
  \item \emph{Pre-asymptotic capacity approximation (\Cref{sec:spectral}).} Mellin--Barnes analysis of covariance resolvents yields a three-term capacity expansion with competing spectral and filter residues (\Cref{lem:resolvent-trace,lem:bregman-fisher,thm:master-mellin,cor:slope-collapse}) and logarithmic transitions at pole confluence (\Cref{prop:pole-collision,prop:gf-holomorphic}). On narrow pilot grids where multi-power fitting is ill-conditioned, first-order Aitken differencing eliminates the loss floor $E_\infty$ to restore linear conditioning and closed-form initialization (\Cref{prop:identifiability,cor:aitken-identifiability}).
  \item \emph{Multi-domain sample allocation (\Cref{sec:mixing}).} KKT duality on the simplex reduces $K$-domain mixing to scalar water-filling with an invariant share-ratio law and finite-budget bottleneck bounds (\Cref{prop:water-filling,cor:powered-share-ratio,thm:master-mixing}). Differentiating the cumulative quota trajectory yields an online sampling curriculum whose distribution is dual to the inverse simplex Hessian governing cross-domain transfer (\Cref{prop:curriculum-rn,prop:simplex-hessian,thm:fisher-transfer}).
  \item \emph{Compute-constrained resource allocation (\Cref{sec:compute}).} Along the iso-compute frontier, monotone loss wrappers preserve a constant allocation exponent (\Cref{thm:invariance-counterexample}), whereas multi-mode capacity and data spectra drive $\eta\opt(C)$ in proportion to exponent variance (\Cref{prop:curvature-ode,thm:master-drift,prop:water-filling-drift,cor:logistic-transition}). Eliminating auxiliary hyperparameters via Schur-complement reduction separates intrinsic scale drift from frozen-schedule bias and bounds downstream allocation regret (\Cref{thm:master-envelope,cor:allocation-regret,prop:surrogate-decision-regret}).
  \item \emph{Numerical and empirical evaluation (\Cref{sec:experiments}).} Across 45 operator spectra and three public benchmarks, our estimators achieve $\MetricPillarIPythiaStdMAPE$ holdout MAPE on Pythia ($70\text{M}\text{--}12\text{B}$, $\MetricPillarIPythiaDedupMAPE$ on Deduped), calibrate pilot exponents ($\MetricPillarIIIChinchillaFullEta \to \MetricPillarIIIChinchillaFiltEta \to 0.5288$) to cut high-compute allocation regret by $\MetricPillarIIIChinchillaHighRegretRed$ (\emph{Frontier}) and $76.4\%$ (\emph{Surface}) on Chinchilla (245 runs), and reduce cross-scale mixture RMSE by up to $59.4\%$ with zero $1\text{M}\to 1\text{B}$ selection regret across 17 RegMix domains.
\end{enumerate}

The remainder of the paper is organized as follows. \Cref{sec:preliminaries} formulates the Hilbert-space resolvent setup, the local Fisher quadratic reduction of cross-entropy risk (\Cref{lem:resolvent-trace,lem:bregman-fisher}), and \Cref{assump:spectral,assump:mixing,assump:compute} (with notation summarized in \Cref{tab:notation-summary} of \Cref{app:preliminaries}), while \Cref{sec:conclusion} discusses implications for pretraining practice and \Cref{app:preliminaries,app:spectral,app:mixing,app:compute} provide complete proofs.

\section{Problem Setup and Preliminaries}
\label{sec:preliminaries}

We begin by formulating the underlying probability space, the Hilbert-space covariance and resolvent operators, the local Fisher quadratic reduction of multi-class cross-entropy, and the structural assumptions that link capacity approximation, domain mixing, and compute allocation.

\subsection{Data Distribution and Covariance Operators}
\label{subsec:prob-hilbert}

Let $(\Omega, \mc{F}, \PP)$ be a complete probability space. Let $(\mc{X}, \mc{B}(\mc{X}))$ be a measurable input space and $(\mc{Y}, \mc{B}(\mc{Y}))$ be a measurable target space. Consider a $K$-domain mixture setting where $I \in \{1, \dots, K\}$ is a discrete domain-index random variable with baseline evaluation prior $\pi = (\pi_1, \dots, \pi_K) \in \Delta_{K-1}^{\circ}$, where
\begin{equation}
  \Delta_{K-1} \;\Let\; \left\{ w \in \R_+^K : \sum_{k=1}^K w_k = 1 \right\}, \qquad
  \Delta_{K-1}^{\circ} \;\Let\; \left\{ w \in (0, 1)^K : \sum_{k=1}^K w_k = 1 \right\}.
\end{equation}
Conditioned on $\{I = k\}$, the input-target pair $(X, Y)$ is drawn from the domain-specific conditional probability measure $\PP_k$ with marginal input measure $\mu_k$. Target performance is evaluated under the fixed evaluation prior $\pi$ (with reference marginal $\mu_\pi$), whereas training samples are drawn under a controllable mixture $w = (w_1, \dots, w_K) \in \Delta_{K-1}^{\circ}$, inducing the domain-conditional, reference, and training mixture laws
\begin{gather}
  \PP_k(\dd x, \dd y) \;\Let\; \PP(X \in \dd x, Y \in \dd y \mid I = k), \qquad
  \mu_k(\dd x) \;\Let\; \PP_k(X \in \dd x), \nonumber\\
  \mu_\pi \;\Let\; \sum_{k=1}^K \pi_k \mu_k, \qquad
  \PP_w \;\Let\; \sum_{k=1}^K w_k \PP_k, \qquad
  \mu_w \;\Let\; \sum_{k=1}^K w_k \mu_k.
\end{gather}

Let $(\mc{H}, \ip{\cdot}{\cdot}_{\mc{H}})$ be a separable real Hilbert space equipped with a measurable feature map $\psi : \mc{X} \to \mc{H}$ satisfying $\E_{\mu_\pi}\!\left[\norm{\psi(X)}_{\mc{H}}^2\right] < \infty$. Under the fixed reference measure $\mu_\pi$, the evaluation sampling operator $\mc{I}_0 : \mc{H} \to L_2(\mc{X}, \mu_\pi)$, defined by $(\mc{I}_0 g)(x) \Let \ip{g}{\psi(x)}_{\mc{H}}$, is Hilbert--Schmidt, and its adjoint yields the fixed self-adjoint, positive, trace-class reference covariance operator $\mc{T} \Let \mc{I}_0^* \mc{I}_0 : \mc{H} \to \mc{H}$:
\begin{gather}
  \mc{T} \;\Let\; \int_{\mc{X}} (\psi(x) \otimes_{\mc{H}} \psi(x)) \,\mu_\pi(\dd x), \nonumber\\
  \ip{g}{\mc{T} h}_{\mc{H}} \;=\; \E_{\mu_\pi}\!\left[ \ip{g}{\psi(X)}_{\mc{H}} \ip{\psi(X)}{h}_{\mc{H}} \right]
  \qquad \forall g, h \in \mc{H}.
\end{gather}
Assuming without loss of generality that $\ker(\mc{T}) = \{0\}$, the Hilbert--Schmidt spectral theorem yields a complete orthonormal basis $\{e_m\}_{m=1}^{\infty}$ of $\mc{H}$ with non-increasing eigenvalues $\lambda_1 \ge \lambda_2 \ge \dots > 0$. Following the standard operator-theoretic framework for kernel and ridge regression \citep{zhang2005effective,caponnetto2007rates,fischer2020sobolev,bartlett2020benign}, for any regularization scale $\mu > 0$, the spectral decomposition of $\mc{T}$, its trace, its regularized resolvent $\mc{R}(\mu)$, and the resulting effective spectral dimension $d_{\mathrm{eff}}(\mu)$ are given by
\begin{gather}
  \mc{T} \;=\; \sum_{m=1}^{\infty} \lambda_m (e_m \otimes_{\mc{H}} e_m), \qquad
  \tr(\mc{T}) \;=\; \E_{\mu_\pi}\!\left[\norm{\psi(X)}_{\mc{H}}^2\right] \;=\; \sum_{m=1}^{\infty} \lambda_m \;<\; \infty, \nonumber\\
  \mc{R}(\mu) \;\Let\; (\mc{T} + \mu \mc{I})^{-1}, \qquad
  d_{\mathrm{eff}}(\mu) \;\Let\; \tr\!\left(\mc{T} \mc{R}(\mu)\right) \;=\; \sum_{m=1}^{\infty} \frac{\lambda_m}{\lambda_m + \mu}.
  \label{eq:resolvent-deff-def}
\end{gather}
When the eigenvalues of $\mc{T}$ decay algebraically at rate $q > 1$, integral comparison with Euler's Beta--Gamma reflection identity yields explicit two-sided bounds on $d_{\mathrm{eff}}(\mu)$.

\begin{lemma}[Effective Spectral Dimension Bounds]
\label{lem:resolvent-trace}
Suppose there exist constants $0 < c_- \le c_+ < \infty$ and a spectral decay exponent $q > 1$ such that the eigenvalues $\{\lambda_m\}_{m=1}^{\infty}$ of $\mc{T}$ obey
\begin{equation}
  c_- m^{-q} \;\le\; \lambda_m \;\le\; c_+ m^{-q} \qquad \forall m \in \N_{\ge 1}.
\end{equation}
For any regularization parameter $\mu > 0$, define the spectral cutoff scales $M_{\mu,\pm}$ and Euler's reflection constant $\Lambda_q$ by
\begin{equation}
  M_{\mu,\pm} \;\Let\; \left(\frac{c_{\pm}}{\mu}\right)^{1/q}, \qquad
  \Lambda_q \;\Let\; \frac{\pi / q}{\sin(\pi / q)} \;>\; 1.
\end{equation}
Then the effective spectral dimension $d_{\mathrm{eff}}(\mu)$ in~\eqref{eq:resolvent-deff-def} satisfies the two-sided bound
\begin{equation}
  \Lambda_q M_{\mu,-} - 1
  \;=\; \int_0^{\infty} \frac{\dd u}{1 + (u / M_{\mu,-})^q} - 1
  \;\le\; d_{\mathrm{eff}}(\mu)
  \;\le\; \int_0^{\infty} \frac{\dd u}{1 + (u / M_{\mu,+})^q}
  \;=\; \Lambda_q M_{\mu,+}.
  \label{eq:deff-sandwich}
\end{equation}
\end{lemma}
\noindent\ProofRef{lem:resolvent-trace}

\subsection{Quadratic Reduction of Cross-Entropy Risk}
\label{subsec:bregman-reduction}

The spectral calculus in \Cref{subsec:prob-hilbert} is formulated for quadratic risk in $\mc{H}$, whereas autoregressive language modeling and multi-class classification minimize conditional cross-entropy over a finite vocabulary $\mc{V}$ of size $V = \abs{\mc{V}} < \infty$. To connect the Hilbert space $\mc{H}$ to vocabulary logits in $\R^V$, let $\bm{\Psi}(x) : \mc{H} \to \R^V$ be a uniformly bounded linear feature operator, and let $\theta, \theta\opt \in \mc{H}$ denote a candidate parameter and the Bayes-optimal parameter:
\begin{equation}
  z_{\theta}(x) \;\Let\; \bm{\Psi}(x)\theta \in \R^V, \qquad
  z\opt(x) \;\Let\; \bm{\Psi}(x)\theta\opt \in \R^V, \qquad
  \sup_{x \in \mc{X}} \norm{\bm{\Psi}(x)}_{\mathrm{op}} \;\le\; \kappa_{\Psi} < \infty.
  \label{eq:linear-logit-model}
\end{equation}
Over the logit space $\R^V$, the multinomial log-partition function $A(z)$ generates the softmax predictive distribution $p(z) \in \Delta_{V-1}^{\circ}$ and expresses the pointwise Kullback--Leibler divergence as a Bregman divergence \citep{wainwright2008graphical,amari2016information}:
\begin{gather}
  A(z) \;\Let\; \ln \sum_{v=1}^V e^{z_v}, \qquad
  p(z) \;\Let\; \nabla A(z) \in \Delta_{V-1}^{\circ}, \nonumber\\
  \KL{p(z\opt)}{p(z)} \;=\; A(z) - A(z\opt) - \ip{p(z\opt)}{z - z\opt}_{\R^V}.
  \label{eq:softmax-bregman-kl}
\end{gather}
Because the linear term in~\eqref{eq:softmax-bregman-kl} cancels the first derivative $\nabla A(z\opt) = p(z\opt)$, the second-order Taylor expansion of $\KL{p(z\opt)}{p(z)}$ around $z\opt(x)$ is governed by the conditional Fisher information matrix $\mc{I}_F(z) \succeq 0$ on $\R^V$. Pulling $\mc{I}_F(z\opt(X))$ back along $\bm{\Psi}(X)$ under the reference evaluation law $\mu_\pi$ defines the self-adjoint Fisher covariance operator $\mc{T}_F : \mc{H} \to \mc{H}$ and links it to the population excess cross-entropy risk $\mc{E}_{\mathrm{KL}}(\theta)$ \citep{martens2020natural,tsiolis2026multiclass}:
\begin{align}
  \mc{I}_F(z) &\;\Let\; \nabla^2 A(z) \;=\; \diag(p(z)) - p(z)p(z)^{\top}, \nonumber\\
  \mc{T}_F &\;\Let\; \E_{\mu_\pi}\!\left[\bm{\Psi}(X)^* \mc{I}_F(z\opt(X)) \bm{\Psi}(X)\right], \nonumber\\
  \mc{E}_{\mathrm{KL}}(\theta) &\;\Let\; \E_{\mu_\pi}\!\left[ \KL{p(z\opt(X))}{p(z_{\theta}(X))} \right].
  \label{eq:fisher-pullback-operator}
\end{align}
Uniformly bounding the third derivative $\nabla^3 A(z)$ over $\R^V$ via self-concordant cumulant control \citep{bach2010selfconcordant} shows that $\mc{T}_F$ is trace-class and controls $\mc{E}_{\mathrm{KL}}(\theta)$ up to a cubic remainder.

\begin{lemma}[Local Fisher Quadratic Approximation]
\label{lem:bregman-fisher}
Under the bounded feature condition in~\eqref{eq:linear-logit-model}, the pulled-back Fisher covariance operator $\mc{T}_F$ in~\eqref{eq:fisher-pullback-operator} is a positive trace-class operator satisfying
\begin{equation}
  \tr(\mc{T}_F) \;\le\; \left(1 - \frac{1}{V}\right)\kappa_{\Psi}^2 \;<\; \infty.
\end{equation}
Furthermore, for every $\theta \in \mc{H}$ with logit residual $\Delta z(x) \Let z_{\theta}(x) - z\opt(x)$, the population excess cross-entropy risk $\mc{E}_{\mathrm{KL}}(\theta)$ obeys the cubic remainder bound
\begin{gather}
  \left\lvert \mc{E}_{\mathrm{KL}}(\theta) - \frac{1}{2}\ip{\theta - \theta\opt}{\mc{T}_F(\theta - \theta\opt)}_{\mc{H}} \right\rvert
  \;\le\; \frac{C_3}{6}\,\E_{\mu_\pi}\!\left[ \norm{\Delta z(X)}_2^3 \right]
  \;\le\; \frac{C_3 \kappa_{\Psi}^3}{6}\norm{\theta - \theta\opt}_{\mc{H}}^3, \nonumber\\
  C_3 \;\Let\; \sup_{z \in \R^V} \norm{\nabla^3 A(z)}_{\mathrm{op}} \;\le\; \frac{1}{\sqrt{2}} \;<\; 1.
  \label{eq:bregman-sandwich}
\end{gather}
\end{lemma}
\noindent\ProofRef{lem:bregman-fisher}

\subsection{Structural Assumptions and Risk Decomposition}
\label{subsec:structural-assumptions}

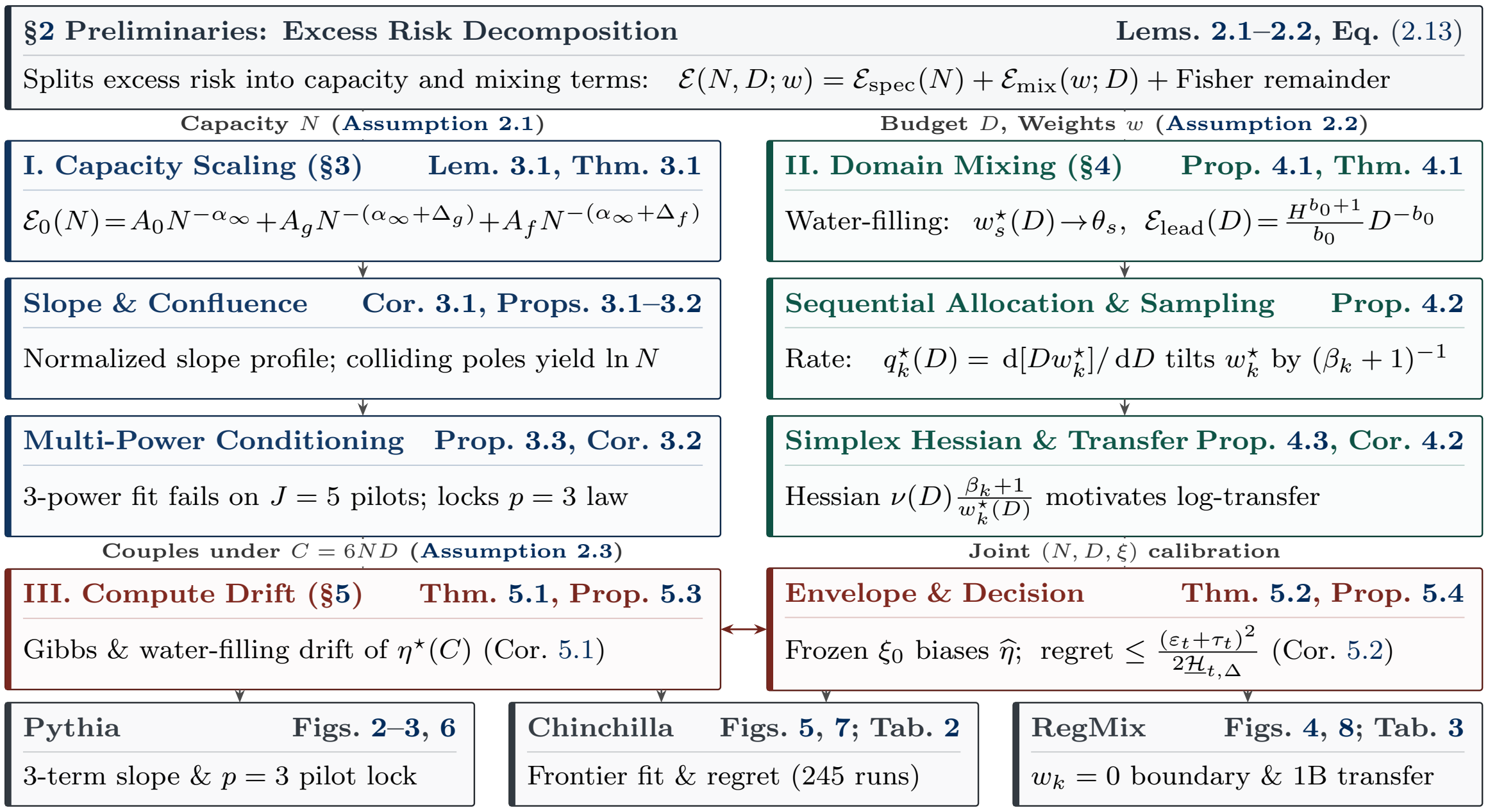
\begin{figure}[t!]
  \centering
  \begin{tikzpicture}[
    x=1cm, y=1cm,
    font=\scriptsize,
    >= {Stealth[length=3.8pt, width=3.0pt]},
    preprintcard/.style={
      draw=#1!65!black,
      fill=#1!2,
      line width=0.45pt,
      rounded corners=0.8pt,
      inner xsep=5.2pt,
      inner ysep=3.2pt,
      outer sep=0pt,
      align=left
    },
    rootcard/.style={
      preprintcard=pRoot,
      fill=pRoot!4,
      text width=14.873cm,
      minimum width=15.24cm,
      minimum height=0.92cm
    },
    pillarcard/.style={
      preprintcard=#1,
      text width=7.013cm,
      minimum width=7.38cm,
      minimum height=1.12cm
    },
    expcard/.style={
      preprintcard=pExp,
      fill=pExp!3,
      text width=4.473cm,
      minimum width=4.84cm,
      minimum height=0.86cm
    },
    edgelbl/.style={
      font=\tiny\bfseries,
      fill=white,
      inner xsep=2.0pt,
      inner ysep=1.0pt,
      outer sep=0pt,
      text=black!80
    },
    arr/.style={
      ->,
      line width=0.50pt,
      draw=black!70
    }
  ]
    \definecolor{pRoot}{RGB}{38,52,68}
    \definecolor{pOne}{RGB}{20,62,112}
    \definecolor{pTwo}{RGB}{16,96,80}
    \definecolor{pThree}{RGB}{142,46,36}
    \definecolor{pExp}{RGB}{60,68,76}

    \newcommand{\cardbody}[5]{%
      \hbox to #1{\strut\textcolor{#2!90!black}{\textbf{#3}}\hfil\textcolor{#2!85!black}{\textbf{#4}}}\vspace{1.6pt}%
      {\color{#2!28!white}\hrule height 0.35pt}\vspace{2.2pt}%
      \hbox to #1{\strut\vphantom{$\tfrac{1}{2}$}#5\hfil}%
    }
    \newcommand{\pillarbody}[5]{%
      \hbox to #1{\strut\textcolor{#2!90!black}{\textbf{#3}}\hfil\textcolor{#2!85!black}{\textbf{#4}}}\vspace{1.6pt}%
      {\color{#2!28!white}\hrule height 0.35pt}\vspace{2.2pt}%
      \hbox to #1{\strut\vphantom{$\tfrac{H^{b_0+1}}{2\underline{\mc{H}}_{t,\Delta}}$}#5\hfil}%
    }

    \node[rootcard] (root) at (0, 0) {%
      \cardbody{14.873cm}{pRoot}%
        {\S\ref{sec:preliminaries} Preliminaries: Excess Risk Decomposition}%
        {Lems.~\ref{lem:resolvent-trace}--\ref{lem:bregman-fisher}, Eq.~\eqref{eq:unified-risk-decomposition}}%
        {Splits excess risk into capacity and mixing terms:\quad $\mc{E}(N,D;w) = \mc{E}_{\mathrm{spec}}(N) + \mc{E}_{\mathrm{mix}}(w; D) + \text{Fisher remainder}$}%
    };

    \node[pillarcard=pOne] (p1_main) at (-3.93, -1.48) {%
      \pillarbody{7.013cm}{pOne}%
        {I.\ Capacity Scaling (\S\ref{sec:spectral})}%
        {Lem.~\ref{lem:mellin-kernel}, Thm.~\ref{thm:master-mellin}}%
        {$\mc{E}_0(N)\!=\!A_0 N^{-\alpha_\infty}\!+\!A_g N^{-(\alpha_\infty+\Delta_g)}\!+\!A_f N^{-(\alpha_\infty+\Delta_f)}$}%
    };
    \node[pillarcard=pTwo] (p2_main) at (3.93, -1.48) {%
      \pillarbody{7.013cm}{pTwo}%
        {II.\ Domain Mixing (\S\ref{sec:mixing})}%
        {Prop.~\ref{prop:water-filling}, Thm.~\ref{thm:master-mixing}}%
        {Water-filling:\; $w_s\opt(D)\!\to\!\theta_s$,\; $\mc{E}_{\mathrm{lead}}(D)\!=\!\tfrac{H^{b_0+1}}{b_0}D^{-b_0}$}%
    };

    \node[pillarcard=pOne] (p1_reg) at (-3.93, -2.90) {%
      \pillarbody{7.013cm}{pOne}%
        {Slope \& Confluence}%
        {Cor.~\ref{cor:slope-collapse}, Props.~\ref{prop:pole-collision}--\ref{prop:gf-holomorphic}}%
        {Normalized slope profile; colliding poles yield $\ln N$}%
    };
    \node[pillarcard=pTwo] (p2_seq) at (3.93, -2.90) {%
      \pillarbody{7.013cm}{pTwo}%
        {Sequential Allocation \& Sampling}%
        {Prop.~\ref{prop:curriculum-rn}}%
        {Rate:\;\; $q_k\opt(D) = \dd[D w_k\opt]/\dd D$ tilts $w_k\opt$ by $(\beta_k+1)^{-1}$}%
    };

    \node[pillarcard=pOne] (p1_id) at (-3.93, -4.32) {%
      \pillarbody{7.013cm}{pOne}%
        {Multi-Power Conditioning}%
        {Prop.~\ref{prop:identifiability}, Cor.~\ref{cor:aitken-identifiability}}%
        {3-power fit fails on $J=5$ pilots; locks $p=3$ law}%
    };
    \node[pillarcard=pTwo] (p2_hess) at (3.93, -4.32) {%
      \pillarbody{7.013cm}{pTwo}%
        {Simplex Hessian \& Transfer}%
        {Prop.~\ref{prop:simplex-hessian}, Cor.~\ref{thm:fisher-transfer}}%
        {Hessian $\nu(D)\tfrac{\beta_k+1}{w_k\opt(D)}$ motivates log-transfer}%
    };

    \node[pillarcard=pThree] (p3_drift) at (-3.93, -5.90) {%
      \pillarbody{7.013cm}{pThree}%
        {III.\ Compute Drift (\S\ref{sec:compute})}%
        {Thm.~\ref{thm:master-drift}, Prop.~\ref{prop:water-filling-drift}}%
        {Gibbs \& water-filling drift of $\eta\opt(C)$ (Cor.~\ref{cor:logistic-transition})}%
    };
    \node[pillarcard=pThree] (p3_env) at (3.93, -5.90) {%
      \pillarbody{7.013cm}{pThree}%
        {Envelope \& Decision}%
        {Thm.~\ref{thm:master-envelope}, Prop.~\ref{prop:surrogate-decision-regret}}%
        {Frozen $\xi_0$ biases $\widehat{\eta}$;\; regret $\le \tfrac{(\varepsilon_t+\tau_t)^2}{2\underline{\mc{H}}_{t,\Delta}}$ (Cor.~\ref{cor:allocation-regret})}%
    };

    \node[expcard] (exp1) at (-5.20, -7.19) {%
      \cardbody{4.473cm}{pExp}%
        {Pythia}%
        {Figs.~\ref{fig:pillar1-mellin}--\ref{fig:pillar1-spectrum-grid}, \ref{fig:pythia-extrapolation}}%
        {3-term slope \& $p=3$ pilot lock}%
    };
    \node[expcard] (exp3) at (0.00, -7.19) {%
      \cardbody{4.473cm}{pExp}%
        {Chinchilla}%
        {Figs.~\ref{fig:pillar3-coupling}, \ref{fig:chinchilla-regret}; Tab.~\ref{tab:chinchilla-calibration}}%
        {Frontier fit \& regret ($245\text{ runs}$)}%
    };
    \node[expcard] (exp2) at (5.20, -7.19) {%
      \cardbody{4.473cm}{pExp}%
        {RegMix}%
        {Figs.~\ref{fig:pillar2-mixing}, \ref{fig:regmix-transfer}; Tab.~\ref{tab:regmix-transfer}}%
        {$w_k=0$ boundary \& $1\text{B}$ transfer}%
    };

    \draw[line width=1.8pt, pRoot!85!black] ([xshift=0.9pt]root.north west) -- ([xshift=0.9pt]root.south west);
    \foreach \n in {p1_main, p1_reg, p1_id} {
      \draw[line width=1.8pt, pOne!85!black] ([xshift=0.9pt]\n.north west) -- ([xshift=0.9pt]\n.south west);
    }
    \foreach \n in {p2_main, p2_seq, p2_hess} {
      \draw[line width=1.8pt, pTwo!85!black] ([xshift=0.9pt]\n.north west) -- ([xshift=0.9pt]\n.south west);
    }
    \foreach \n in {p3_drift, p3_env} {
      \draw[line width=1.8pt, pThree!85!black] ([xshift=0.9pt]\n.north west) -- ([xshift=0.9pt]\n.south west);
    }
    \foreach \n in {exp1, exp3, exp2} {
      \draw[line width=1.8pt, pExp!85!black] ([xshift=0.9pt]\n.north west) -- ([xshift=0.9pt]\n.south west);
    }

    \draw[arr] (root.south -| p1_main.north) -- node[edgelbl] {Capacity $N$ (\Cref{assump:spectral})} (p1_main.north);
    \draw[arr] (root.south -| p2_main.north) -- node[edgelbl] {Budget $D$, Weights $w$ (\Cref{assump:mixing})} (p2_main.north);

    \draw[arr] (p1_main.south) -- (p1_reg.north);
    \draw[arr] (p1_reg.south) -- (p1_id.north);

    \draw[arr] (p2_main.south) -- (p2_seq.north);
    \draw[arr] (p2_seq.south) -- (p2_hess.north);

    \draw[arr] (p1_id.south) -- node[edgelbl] {Couples under $C = 6ND$ (\Cref{assump:compute})} (p3_drift.north);
    \draw[arr] (p2_hess.south) -- node[edgelbl] {Joint $(N, D, \xi)$ calibration} (p3_env.north);
    \draw[<->, line width=0.55pt, draw=pThree!80!black] (p3_drift.east) -- (p3_env.west);

    \draw[arr] (p3_drift.south -| exp1.north) -- (exp1.north);
    \draw[arr] ([xshift=-0.85cm]exp3.north |- p3_drift.south) -- ([xshift=-0.85cm]exp3.north);
    \draw[arr] ([xshift=0.85cm]exp3.north |- p3_env.south) -- ([xshift=0.85cm]exp3.north);
    \draw[arr] (p3_env.south -| exp2.north) -- (exp2.north);
  \end{tikzpicture}
  \caption{\textbf{Conceptual Architecture and Reading Roadmap.} The finite-budget risk decomposition (\Cref{sec:preliminaries}) branches into I.\ Capacity Scaling (\Cref{sec:spectral}) and II.\ Multi-Domain Mixing (\Cref{sec:mixing}), couples along $C = 6ND$ in III.\ Compute-Optimal Allocation (\Cref{sec:compute}), and is evaluated in \Cref{sec:experiments}.}
  \label{fig:proof-dag}
\end{figure}

Building on the Hilbert-space resolvent calculus of \Cref{subsec:prob-hilbert} and the local Fisher quadratic reduction of \Cref{subsec:bregman-reduction}, we model the baseline finite-budget excess risk of a network of size $N$ trained on $D$ tokens under domain mixture $w \in \Delta_{K-1}^{\circ}$ at the fixed reference covariance operator $\mc{T}$ and target evaluation prior $\pi$ via an additive decomposition into a capacity-limited approximation error $\mc{E}_{\mathrm{spec}}(N)$ and a sample-limited estimation error $\mc{E}_{\mathrm{mix}}(w; D)$ \citep{caponnetto2007rates,fischer2020sobolev}:
\begin{equation}
  \mc{E}(N, D; w) \;=\; \mc{E}_{\mathrm{spec}}(N) + \mc{E}_{\mathrm{mix}}(w; D)
  \quad\Longrightarrow\quad
  \mc{E}\opt(C) \;\Let\; \min_{\substack{\kappa_c N D = C \\ w \in \Delta_{K-1}}} \mc{E}(N, D; w).
  \label{eq:unified-risk-decomposition}
\end{equation}
Cross-distribution transfer couplings $\bm{\Gamma}$ when $\mu_w \neq \mu_\pi$ are analyzed in \Cref{subsec:simplex-hessian-metric}, nonlinear monotone wrappers in \Cref{subsec:invariance-counterexample}, joint hyperparameter surfaces in \Cref{subsec:envelope-regret}, and non-separable slice perturbations in \Cref{prop:surrogate-decision-regret}. We now specify the three baseline components of~\eqref{eq:unified-risk-decomposition} through \Cref{assump:spectral,assump:mixing,assump:compute} (\Cref{fig:proof-dag}, with core notation summarized in \Cref{tab:notation-summary} in \Cref{app:preliminaries}).

To formulate the scalar spectral surrogate model for $\mc{E}_{\mathrm{spec}}(N)$ under the fixed reference operator $\mc{T} = \mc{I}_0^*\mc{I}_0$, let $\{\phi_m\}_{m=1}^{\infty}$ denote the $L_2(\mc{X}, \mu_\pi)$-orthonormal eigenfunctions spanning the closure $\overline{\operatorname{ran}(\mc{I}_0)}$ associated with the eigenbasis $\{e_m\}_{m=1}^{\infty}$ of $\mc{T}$, and let $g_\mu \in L_2(\mc{X}, \mu_\pi)$ denote the Tikhonov-regularized predictor of a target function offset $g\opt \in \overline{\operatorname{ran}(\mc{I}_0)} \subset L_2(\mc{X}, \mu_\pi)$ at regularization scale $\mu > 0$:
\begin{gather}
  \phi_m \;\Let\; \lambda_m^{-1/2}\mc{I}_0 e_m, \qquad
  c_m \;\Let\; \abs{\ip{g\opt}{\phi_m}_{L_2(\mu_\pi)}}^2, \qquad
  \norm{g\opt}_{L_2(\mu_\pi)}^2 \;=\; \sum_{m=1}^{\infty} c_m \;<\; \infty, \nonumber\\
  g_\mu \;\Let\; \mc{I}_0(\mc{T} + \mu\mc{I})^{-1}\mc{I}_0^* g\opt, \qquad
  \norm{g\opt - g_\mu}_{L_2(\mu_\pi)}^2
  \;=\; \sum_{m=1}^{\infty} c_m\left(\frac{\mu}{\lambda_m + \mu}\right)^2.
  \label{eq:tikhonov-residual-bridge}
\end{gather}
Identity~\eqref{eq:tikhonov-residual-bridge} factors each mode's quadratic residual into a target energy $c_m$ and a bounded attenuation factor $(\mu / (\lambda_m + \mu))^2 \in (0, 1)$. To connect this scalar surrogate to the multi-class cross-entropy reduction of \Cref{lem:bregman-fisher} for a realizable target $g\opt = \mc{I}_0\theta\opt$ and Tikhonov iterate $\theta_\mu \Let (\mc{T} + \mu\mc{I})^{-1}\mc{T}\theta\opt$ (with $g_\mu = \mc{I}_0\theta_\mu$), we distinguish two operator regimes:
\begin{itemize}
  \item \emph{Single-contrast Fisher identity ($\mc{T} = \mc{T}_F$):} When the logit residual $\Delta z(x)$ lies in a single scalar contrast subspace with $\mc{T} = \mc{T}_F$, the local Fisher quadratic form equals one half of~\eqref{eq:tikhonov-residual-bridge}, and under uniformly bounded eigenfunctions $\sup_m\norm{\phi_m}_\infty \le C_\phi$ with $r > 2$ and $\mu = c_0 M^{-q}$, $L_2$--$L_\infty$ interpolation bounds the scalar cubic residual by
  \begin{align}
    \frac{1}{2}\ip{\theta_\mu - \theta\opt}{\mc{T}_F(\theta_\mu - \theta\opt)}_{\mc{H}}
    &\;=\; \frac{1}{2}\norm{g\opt - g_\mu}_{L_2(\mu_\pi)}^2, \nonumber\\
    \norm{g\opt - g_\mu}_{L_3(\mu_\pi)}^3
    &\;\le\; \norm{g\opt - g_\mu}_{L_2(\mu_\pi)}^2\,\norm{g\opt - g_\mu}_{\infty} \nonumber\\
    &\;=\;
    \begin{cases}
      \mc{O}(M^{2 - 3r/2}), & 2 < r < 2q + 1, \\
      \mc{O}\!\left(M^{-2q - \min\{r/2 - 1,\,q\}}\ln M\right), & r \ge 2q + 1,
    \end{cases}
  \end{align}
  scaling exact residue amplitudes by $\frac{1}{2}$ whenever any orthogonal logit components and the cubic remainder in~\eqref{eq:bregman-sandwich} are of strictly higher order than the retained subleading pole (e.g., $\frac{3}{2}r - 2 > 2q$).
  \item \emph{Two-sided spectral equivalence ($c_F\mc{T} \preceq \mc{T}_F \preceq C_F\mc{T}$):} For general $0 < c_F \le C_F < \infty$, the sandwich~\eqref{eq:bregman-sandwich} transfers two-sided quadratic risk rate bounds between $\mc{T}$ and $\mc{T}_F$.
\end{itemize}
Within the scalar surrogate~\eqref{eq:tikhonov-residual-bridge}, algebraic mode decay $c_m = c\,m^{-r}(1 + b\,m^{-\delta})$ and $\lambda_m = c_0 m^{-q}$ induce three nested regularity regimes \citep{fischer2020sobolev}:
\begin{itemize}
  \item \emph{Finite $L_2(\mu_\pi)$ target energy ($r > 1$, $q > 0$):} The mode series $\sum_{m=1}^{\infty} c_m < \infty$ and the attenuated residual risk~\eqref{eq:tikhonov-residual-bridge} converge absolutely for every $\mu > 0$.
  \item \emph{Trace-class covariance ($q > 1$):} The eigenvalue sum $\tr(\mc{T}) < \infty$ is finite, as required by \Cref{lem:resolvent-trace,lem:bregman-fisher}.
  \item \emph{$\mc{H}$-realizable source condition ($r > q + 1$):} The target belongs to the reproducing subspace $g\opt = \mc{I}_0 f\opt$ for $f\opt \in \mc{H}$ with finite Hilbert norm $\sum_{m=1}^{\infty} c_m/\lambda_m < \infty$ and mode weights $c_m = \lambda_m\abs{\ip{f\opt}{e_m}_{\mc{H}}}^2$.
\end{itemize}
Because $(\mu / (\lambda_m + \mu))^2 \in (0, 1)$, the spectral risk series~\eqref{eq:tikhonov-residual-bridge} remains well-defined and absolutely convergent throughout the full range $q > 0$ and $r > 1$, which our first assumption parameterizes at finite model capacity $N$.

\begin{assumption}[Spectral Source Decay and Resolvent Filter]
\label{assump:spectral}
A network of parameter count $N \in \R_{>0}$ resolves spectral modes up to an effective cutoff scale $\CapacityScaleDef$ ($\kappa > 0$). Under eigenvalue decay $\lambda_m = c_0 m^{-q}$ ($q > 0$) and resolution floor $\mu = c_0 M^{-q}$, each mode's attenuation factor satisfies $(\mu / (\lambda_m + \mu))^2 = \psi_q(m/M)$, and the target mode energies $\{c_m\}_{m=1}^{\infty}$ and capacity-limited spectral excess risk $\mc{E}_{\mathrm{spec}}(N)$ obey the two-term Weyl expansion and rational filter law
\begin{align}
  c_m &\;=\; c\,m^{-r}\!\left(1 + b\,m^{-\delta}\right),
  \label{eq:weyl-mode-energy} \\
  \mc{E}_{\mathrm{spec}}(N) &\;\Let\; \sum_{m=1}^{\infty} c_m\,\psi_q\!\left(\frac{m}{M}\right), \qquad
  \psi_q(u) \;\Let\; \left(1 + u^{-q}\right)^{-2},
  \label{eq:rational-filter}
\end{align}
where $c > 0$ is the leading spectral amplitude, $r > 1$ is the primary source decay exponent, $b \ge 0$ is the relative subleading amplitude, and $\delta = 1/d > 0$ is the geometric correction exponent on an effective $d$-dimensional data manifold.
\end{assumption}

\Cref{assump:spectral} combines the classical two-term Weyl eigenvalue and manifold-regularity asymptotics of differential and Mat\'ern covariance operators \citep{weyl1912eigenvalues,hormander1968spectral,borovitskiy2020matern} with the capacity-scaling exponents of deep piecewise-linear and adaptive neural networks \citep[e.g.,][]{yarotsky2017error,schmidthieber2020regression,zhao2026foundations}. Rather than hard Eckart--Young truncation $\Ind{m > M}$ \citep{eckart1936approximation} (whose Mellin transform carries no left-half-plane filter poles), projecting an infinite-dimensional feature map onto an $M_{\mathrm{raw}}$-dimensional random subspace induces a self-consistent Marchenko--Pastur resolution floor $\mu_{\mathrm{eff}}(N)$ on $\mc{T}_F$ \citep{marchenko1967distribution,jacot2020implicit,bordelon2020spectrum,maloney2022solvable,bahri2024explaining}:
\begin{equation}
  M \;=\; \frac{M_{\mathrm{raw}}}{\Lambda_q}, \qquad
  \mu_{\mathrm{eff}}(N) \;=\; c_0 M^{-q}
  \quad\Longrightarrow\quad
  \left(\frac{\mu_{\mathrm{eff}}(N)}{\lambda_m + \mu_{\mathrm{eff}}(N)}\right)^2 \;=\; \psi_q\!\left(\frac{m}{M}\right),
\end{equation}
where $\Lambda_q = (\pi/q)/\sin(\pi/q)$ is Euler's effective-dimension constant from \Cref{lem:resolvent-trace}. To complement this capacity-side approximation model, we next specify how finite training tokens are distributed across heterogeneous data domains.

\begin{assumption}[Separable Multi-Domain Excess Risk]
\label{assump:mixing}
Under a total training budget of $D > 0$ tokens allocated across $K \ge 2$ domains via mixture weights $w \in \Delta_{K-1}^{\circ}$, the cumulative token quota assigned to domain $k$ is $\DomainTokensDef$. After subtracting the irreducible entropy floor $\mc{E}_{\infty}$, the separable domain-mixture excess risk over $\Delta_{K-1}^{\circ}$ \citep{ge2024bimix,ye2025datamixing,liu2025regmix} takes the form
\begin{equation}
  \mc{E}_{\mathrm{mix}}(w; D) \;\Let\; \sum_{k=1}^K \pi_k B_k \left(w_k D\right)^{-\beta_k} \;=\; \sum_{k=1}^K c_k \left(w_k D\right)^{-\beta_k},
  \label{eq:simplex-risk-def}
\end{equation}
where $\pi \in \Delta_{K-1}^{\circ}$ is the target evaluation prior, $B_k > 0$ is the domain difficulty scale, $c_k \Let \pi_k B_k > 0$ is the effective domain weight, and $\beta_k > 0$ is the domain sample-scaling exponent. When cross-domain representation transfer is active \citep{hernandez2021transfer,wu2026datamixtures}, the excess risk includes a convex quadratic transfer interaction $\frac{1}{2}\ip{w}{\bm{\Gamma} w}_{\R^K}$ with symmetric positive semidefinite coupling matrix $\bm{\Gamma} \in \R^{K \times K}$ ($\bm{\Gamma} \succeq 0$).
\end{assumption}

Finally, to couple model capacity $N$ and token budget $D$ under a fixed training compute budget $C = \kappa_c N D$ \citep{kaplan2020scaling,hoffmann2022chinchilla}, we impose standard smoothness and slice-convexity conditions on the joint excess risk surface.

\begin{assumption}[Regularity of the Iso-Compute Risk Surface]
\label{assump:compute}
In logarithmic resource coordinates $u \Let \ln N \in \R$ and $v \Let \ln D \in \R$, the compute budget $C = \kappa_c N D$ ($\kappa_c > 0$) defines the affine iso-compute submanifold
\begin{equation}
  \mc{M}_t \;\Let\; \left\{ (u, v) \in \R^2 : u + v = t \right\}, \qquad t \;\Let\; \ln\!\left(\frac{C}{\kappa_c}\right) \in \R.
\end{equation}
The excess risk surface $\mc{E} : \R^2 \to \R_{>0}$ is three times continuously differentiable ($\mc{E} \in C^3(\R^2)$), strictly decreasing in each coordinate ($\mc{E}_u < 0, \mc{E}_v < 0$), and strictly convex along every iso-compute slice $\mc{M}_t$, meaning that the second directional derivative along $(1, -1)$ satisfies
\begin{equation}
  \mc{H}_{\mathrm{iso}}(u, v) \;\Let\; \mc{E}_{uu}(u, v) - 2\mc{E}_{uv}(u, v) + \mc{E}_{vv}(u, v) \;>\; 0 \qquad \forall (u, v) \in \R^2,
  \label{eq:iso-compute-convexity}
\end{equation}
and admits a unique interior minimizer $(u\opt(t), v\opt(t)) \in \mc{M}_t$ for each log-compute budget $t \in \R$.
\end{assumption}

\section{Finite-Capacity Approximation and Identifiability}
\label{sec:spectral}

We begin our analysis of the excess risk decomposition~\eqref{eq:unified-risk-decomposition} with the finite-capacity approximation error $\mc{E}_{\mathrm{spec}}(N)$ in $\mc{H}$ under \Cref{assump:spectral}. In empirical scaling studies, practitioners commonly fit a single power law $E_\infty + A_0 N^{-\alpha_\infty}$ to small-scale pilot runs and extrapolate across several orders of magnitude \citep{hestness2017scaling,rosenfeld2019predictability,kaplan2020scaling,henighan2020autoregressive}. On finite model grids, however, the local log-log slope $\alpha_{\mathrm{eff}}(N) \Let -\frac{\dd \ln \mc{E}_{\mathrm{spec}}(N)}{\dd \ln N}$ departs systematically from $\alpha_\infty$, either undershooting or overshooting the asymptotic rate before converging at large $N$ \citep{weyl1912eigenvalues,spigler2020learning,bordelon2020spectrum,caballero2023broken}.

To characterize these finite-scale departures directly from the operator spectrum, we apply Mellin--Barnes contour integration to the spectral risk series $\mc{E}_{\mathrm{spec}}(N) = \sum_{m=1}^{\infty} c_m \psi_q(m/M)$. Shifting the integration contour into the left complex half-plane $\Re(s) < 0$ reveals that pre-asymptotic capacity scaling is governed by a residue competition between two distinct families of singularities:
\begin{itemize}
  \item \textbf{Spectral--geometric zeta poles} at $s = -(r-1)$ and $s = -(r+\delta-1)$, generated by the analytic continuation of $\zeta(r+s)$ and $\zeta(r+\delta+s)$ encoding the target source decay and Weyl manifold correction; and
  \item \textbf{Regularization filter poles} at $s = -mq$ for integers $m \ge 2$, generated by the Mellin kernel $\Psi_q(s)$ of the rational resolvent filter $\psi_q(u) = (1+u^{-q})^{-2}$ (whereas the apparent pole at $s = -q$ is removable).
\end{itemize}
We establish the resulting three-term residue expansion and its pre-asymptotic slope profile in \Cref{subsec:mellin-barnes}, analyze pole confluence and early-stopped gradient flow in \Cref{subsec:gf-holomorphic}, and characterize multi-power ill-conditioning and difference identification in \Cref{subsec:identifiability}.

\subsection{Three-Term Risk Expansion and Pre-Asymptotic Slope}
\label{subsec:mellin-barnes}
\label{subsec:slope-collapse}

Under the capacity scaling $M = N^{\kappa}$ (\Cref{assump:spectral}), each simple pole $s = -p$ in the left half-plane $\Re(s) < 0$ contributes a power-law mode $M^{-p} = N^{-\kappa p}$ to the approximation risk $\mc{E}_{\mathrm{spec}}(N)$. The three rightmost singularities therefore decompose pre-asymptotic capacity scaling into three distinct exponents:
\begin{itemize}
  \item the \emph{asymptotic tail exponent} $\alpha_\infty \Let \kappa(r - 1)$, governed by the primary source pole $s = -(r - 1)$ as $N \to \infty$;
  \item the \emph{geometric correction gap} $\Delta_g \Let \kappa\delta$, which shifts the subleading Weyl manifold pole $s = -(r + \delta - 1)$ to the faster decay rate $\alpha_\infty + \Delta_g$; and
  \item the \emph{regularization filter gap} $\Delta_f \Let \kappa(2q - r + 1)$, which shifts the first active resolvent pole $s = -2q$ to the finite-resolution decay rate $\alpha_\infty + \Delta_f$.
\end{itemize}
Whether the finite-scale correction is dominated by manifold geometry ($\Delta_g < \Delta_f$) or by the regularization filter ($\Delta_f < \Delta_g$) is governed by the dimensionless \emph{spectral gap ratio} $\varrho$:
\begin{equation}
  \alpha_\infty \Let \kappa(r - 1), \qquad \Delta_g \Let \kappa\delta, \qquad \Delta_f \Let \kappa(2q - r + 1), \qquad
  \varrho \Let \frac{\Delta_f}{\Delta_g} = \frac{2q - r + 1}{\delta}.
  \label{eq:gap-definitions}
\end{equation}
To evaluate the contour integral of $\mc{E}_{\mathrm{spec}}(N)$, we begin by determining the meromorphic continuation of the rational filter kernel $\Psi_q(s)$; notably, its apparent singularity at $s = -q$ cancels against a simple zero of the linear numerator, shifting the first active left-half-plane pole to $s = -2q$.

\begin{lemma}[Mellin Kernel of the Rational Resolvent Filter]
\label{lem:mellin-kernel}
For $q > 0$ and $\Re(s) \in (-2q, 0)$, the Mellin transform $\Psi_q(s) \Let \int_0^{\infty} \psi_q(u) u^{s-1}\dd u$ of the rational filter $\psi_q(u) = (1 + u^{-q})^{-2}$ converges absolutely and extends meromorphically to $\C$ via
\begin{equation}
  \Psi_q(s) \;=\; \frac{1}{q}\Gamma\!\left(2 + \frac{s}{q}\right)\Gamma\!\left(-\frac{s}{q}\right)
  \;=\; -\frac{s + q}{q^2}\,\frac{\pi}{\sin(\pi s / q)}.
  \label{eq:psi-q-mellin}
\end{equation}
The linear factor $(s + q)$ cancels the simple zero of $\sin(\pi s / q)$ at $s = -q$, rendering $s = -q$ a removable singularity with $\Psi_q(-q) = \frac{1}{q} > 0$. Consequently, $\Psi_q(s) > 0$ for all real $s \in (-2q, 0)$, and the poles of $\Psi_q(s)$ in the open left half-plane $\Re(s) < 0$ are simple poles located at $s_m = -mq$ for integers $m \ge 2$, with residues
\begin{equation}
  \operatorname*{Res}_{s = -mq} \Psi_q(s) \;=\; (-1)^{m-2}(m - 1) \qquad \forall m \in \N_{\ge 2}.
  \label{eq:psi-q-residues}
\end{equation}
\end{lemma}
\noindent\ProofRef{lem:mellin-kernel}

Combining the meromorphic filter kernel $\Psi_q(s)$ of \Cref{lem:mellin-kernel} with the Weyl Dirichlet series $\sum_{m=1}^{\infty} c_m m^{-s} = c\{\zeta(r+s) + b\,\zeta(r+\delta+s)\}$ represents $\mc{E}_{\mathrm{spec}}(N)$ as a vertical contour integral of the Mellin integrand $\Phi(s; M)$ whose left-half-plane pole abscissae form the set $\mc{P}$ \citep{flajolet1995mellin,paris2001asymptotics}:
\begin{align}
  \Phi(s; M) &\;\Let\; c M^s \Psi_q(s)\bigl\{\zeta(r+s) + b\,\zeta(r+\delta+s)\bigr\}, \nonumber\\
  \mc{P} &\;\Let\; \{r - 1,\, r + \delta - 1\} \cup \{mq : m \in \N_{\ge 2}\}.
  \label{eq:mellin-integrand-pole-set}
\end{align}
Shifting the integration contour leftward to $\Re(s) = -\rho$ for any abscissa $\rho \in (0, \infty) \setminus \mc{P}$ isolates the finite-capacity residue expansion of $\mc{E}_{\mathrm{spec}}(N)$ up to the contour remainder $R_{\rho}(M) \Let \frac{1}{2\pi i}\int_{-\rho - i\infty}^{-\rho + i\infty}\Phi(s; M)\dd s$.

\begin{theorem}[Finite-Capacity Mellin--Barnes Expansion]
\label{thm:master-mellin}
Suppose \Cref{assump:spectral} holds with cutoff scale $M = N^{\kappa}$ and non-overlapping spectral and filter poles, $\{r-1, r+\delta-1\} \cap \{mq : m \in \N_{\ge 2}\} = \emptyset$. Then for any shifted abscissa $\rho \in (0, \infty) \setminus \mc{P}$, the finite-capacity spectral approximation risk $\mc{E}_{\mathrm{spec}}(N)$ admits the residue expansion
\begin{align}
  \mc{E}_{\mathrm{spec}}(N)
  &\;=\; A_0 M^{-(r-1)}\Ind{r-1 < \rho} + A_g M^{-(r+\delta-1)}\Ind{r+\delta-1 < \rho} \nonumber\\
  &\qquad + \sum_{\substack{m \ge 2 \\ mq < \rho}} A_{f,m} M^{-mq} + R_{\rho}(M),
  \label{eq:general-mellin-expansion}
\end{align}
where the residues and contour remainder (with finite constant $C_{\rho} > 0$ from \Cref{lem:app-mellin-strip}) satisfy
\begin{gather}
  A_0 \;\Let\; c\,\Psi_q(1-r), \qquad A_g \;\Let\; c b\,\Psi_q(1-r-\delta), \qquad
  \abs{R_{\rho}(M)} \;\le\; C_{\rho} M^{-\rho}, \\
  A_{f,m} \;\Let\; c(-1)^{m-2}(m-1)\left\{\zeta(r - mq) + b\,\zeta(r + \delta - mq)\right\} \quad (m \ge 2).
  \label{eq:mellin-remainder-bound}
\end{gather}
In particular, in the non-degenerate three-term regime where the spectral and filter exponents obey the ordering in~\eqref{eq:three-term-expansion}, the expansion specializes to
\begin{gather}
  0 < r - 1 < r + \delta - 1 < 2q < \rho < 3q, \qquad
  -2 < r - 2q < r + \delta - 2q < 1, \nonumber\\
  \mc{E}_{\mathrm{spec}}(N) \;=\; A_0 N^{-\alpha_\infty} + A_g N^{-(\alpha_\infty + \Delta_g)} + A_f N^{-(\alpha_\infty + \Delta_f)} + R_{\rho}(N^{\kappa}), \nonumber\\
  A_f \;\Let\; A_{f,2} \;=\; c\left\{\zeta(r-2q) + b\,\zeta(r+\delta-2q)\right\} \;<\; 0,
  \label{eq:three-term-expansion}
\end{gather}
where $\varrho > 1$, $A_0 > 0$, $A_g \ge 0$ (with $A_g > 0$ whenever $b > 0$), and $A_f < 0$ for every $b \ge 0$ because $\zeta(z) < 0$ across the open interval $z \in (-2, 1)$ (strictly above the first trivial zero $\zeta(-2) = 0$).
\end{theorem}
\begin{proof}[Proof Sketch of \Cref{thm:master-mellin}]
For $\sigma \in (\max\{-2q, 1-r\}, 0)$, Mellin inversion of $\psi_q(m/M)$ and Fubini--Tonelli summation over $m \ge 1$ express $\mc{E}_{\mathrm{spec}}(N)$ as a vertical contour integral of $\Phi(s; M)$. Shifting the contour leftward from $\Re(s) = \sigma$ to $\Re(s) = -\rho$ via Cauchy's residue theorem (\Cref{lem:app-mellin-strip}) and evaluating the residue signs via Riemann's functional equation \citep{titchmarsh1986riemann} yields
\begin{align*}
  \mc{E}_{\mathrm{spec}}(N)
  &\;=\; \frac{1}{2\pi i}\int_{\sigma - i\infty}^{\sigma + i\infty} \Phi(s; M)\dd s
  \;=\; \sum_{s_\star \in (-\rho, \sigma)} \operatorname*{Res}_{s = s_\star}\Phi(s; M)
  + R_\rho(M), \\
  \Psi_q(s) > 0 \;\text{ on } (-2q, 0) &\;\Longrightarrow\; A_0 > 0,\; A_g \ge 0, \qquad
  \zeta(z) < 0 \;\text{ on } (-2, 1) \;\Longrightarrow\; A_f < 0,
\end{align*}
collecting the simple poles of $\zeta(r+s)$ and $\zeta(r+\delta+s)$ at $s \in \{1-r, 1-r-\delta\}$ and the filter poles of $\Psi_q(s)$ at $s = -mq$ ($m \ge 2$) from \Cref{lem:mellin-kernel}. \ProofRef{thm:master-mellin}
\end{proof}

\begin{remark}[Series Truncation and Low-Mode Perturbations]
\label{rem:gevrey-truncation}
Two structural properties govern the higher-order behavior and low-index sensitivity of the Mellin--Barnes expansion~\eqref{eq:general-mellin-expansion}:
\begin{enumerate}[label=(\roman*), font=\normalfont]
  \item \emph{Finite-order truncation versus factorial growth as $\rho \to \infty$.} For any fixed abscissa $\rho > 0$,~\eqref{eq:general-mellin-expansion} is a finite sum with contour remainder $\abs{R_\rho(M)} \le C_\rho M^{-\rho}$. Sending $\rho \to \infty$ over all filter poles $s = -mq$ ($m \ge 2$ with $r - mq < 0$) produces two distinct regimes governed by Euler's reflection formula and the trivial zeros of $\zeta(z)$ at $-2\N_{\ge 1}$ \citep{boyd1999devil,paris2001asymptotics}:
  \begin{align}
    r - mq \in -2\N_{\ge 1},\; b = 0
    &\Longrightarrow
    \zeta(r - mq) = 0 \quad (\forall\text{ large } m \ge 2), \nonumber\\
    \operatorname{dist}(r - mq, -2\N_{\ge 1}) \ge \delta_0
    &\Longrightarrow
    \abs{\zeta(r - mq)} \ge 2(2\pi)^{r - mq - 1}\Gamma(1 - r + mq)\sin(\pi\delta_0/2),
  \end{align}
  for any fixed separation $\delta_0 > 0$. Thus, in the resonant case $r - mq \in -2\N_{\ge 1}$ ($b = 0$), the filter series terminates at finite algebraic order; in the non-resonant case, the filter terms grow factorially in $qm$ (Gevrey-$1$ in $qm$), whose Stirling envelope $\Gamma(1 - r + mq)(2\pi M)^{-mq}$ attains its minimum near the optimal truncation scale $qm = 2\pi M$.
  \item \emph{Invariance of geometric poles under finite low-mode perturbations.} When the true mode energies $c_j$ depart from the power-law profile~\eqref{eq:weyl-mode-energy} at finitely many low indices $j \in \{1, \dots, m_0\}$, the finite Dirichlet polynomial $\sum_{j=1}^{m_0} \delta c_j\,j^{-s}$ is entire on $\C$ and introduces no poles at $s \in \{1-r, 1-r-\delta\}$:
  \begin{gather}
    \delta c_j \;\Let\; c_j - c\,j^{-r}\!\left(1 + b j^{-\delta}\right) \quad (1 \le j \le m_0), \nonumber\\
    \Delta A_0 \;=\; \Delta A_g \;=\; 0, \qquad
    \Delta A_{f,m} \;=\; (-1)^{m-2}(m-1)\sum_{j=1}^{m_0} \delta c_j\,j^{mq} \quad (m \ge 2).
    \label{eq:low-mode-residue-shift}
  \end{gather}
  Consequently, the spectral--geometric residues $(A_0, A_g)$ depend solely on the high-mode tail $j \to \infty$, whereas the filter coefficients $A_{f,m}$ absorb finite low-mode energy perturbations.
\end{enumerate}
\end{remark}

\Cref{thm:master-mellin} explains why pre-asymptotic scaling curves exhibit non-monotonic local slopes when $\varrho > 1$: the positive subleading geometric residue $A_g N^{-(\alpha_\infty + \Delta_g)}$ ($A_g > 0$) and the negative leading filter residue $A_f N^{-(\alpha_\infty + \Delta_f)}$ ($A_f < 0$) pull the local slope in opposite directions across scales. To isolate this competition on the three-term truncation $\mc{E}_0(N)$, let $\alpha_0(N) \Let -\frac{\dd \ln \mc{E}_0(N)}{\dd \ln N}$ denote its local scaling exponent on the positive-risk domain $\mc{E}_0(N) > 0$ (which includes all $N > \abs{f}^{1/\Delta_f}$) and define the relative amplitudes $(g, f)$ and rescaled slope deviation $T_3(N)$ on $(0, \infty)$ by
\begin{gather}
  \mc{E}_0(N) \;\Let\; A_0 N^{-\alpha_\infty}\bigl(1 + g N^{-\Delta_g} + f N^{-\Delta_f}\bigr), \qquad
  g \;\Let\; \frac{A_g}{A_0} \;>\; 0, \qquad
  f \;\Let\; \frac{A_f}{A_0} \;<\; 0, \nonumber\\
  T_3(N) \;\Let\; \Delta_g g N^{-\Delta_g} - \Delta_f \abs{f} N^{-\Delta_f}
  \;=\; \left(\alpha_0(N) - \alpha_\infty\right)\frac{\mc{E}_0(N)}{A_0 N^{-\alpha_\infty}} \quad (\mc{E}_0(N) > 0).
  \label{eq:T3-functional}
\end{gather}
Normalizing $T_3(N)$ by its unique extremum collapses all amplitude dependencies $(A_0, A_g, A_f)$ onto a single-parameter profile $F_{\varrho}(x)$ governed solely by $\varrho$.

\begin{corollary}[Normalized Pre-Asymptotic Slope Profile]
\label{cor:slope-collapse}
In the regime $\varrho > 1$ with $g > 0$ and $f < 0$, the rescaled slope deviation $T_3(N)$ in~\eqref{eq:T3-functional} attains a unique global maximum on $(0, \infty)$ at the characteristic peak scale $N_{\mathrm{peak}}\opt$ with maximum height $H_{\mathrm{peak}}$:
\begin{align}
  N_{\mathrm{peak}}\opt &\;\Let\; \left(\frac{\Delta_f^2 \abs{f}}{\Delta_g^2 g}\right)^{\frac{1}{\Delta_f - \Delta_g}}, \nonumber\\
  H_{\mathrm{peak}} &\;\Let\; T_3(N_{\mathrm{peak}}\opt) \;=\; \Delta_g g (N_{\mathrm{peak}}\opt)^{-\Delta_g}\!\left(1 - \frac{1}{\varrho}\right) \;>\; 0.
  \label{eq:N-peak-law}
\end{align}
Under the dimensionless coordinate $x \Let (N / N_{\mathrm{peak}}\opt)^{\Delta_g}$ on $(0, \infty)$, the normalized slope deviation satisfies
\begin{equation}
  \frac{T_3(N)}{H_{\mathrm{peak}}} \;=\; F_\varrho(x) \;\Let\; \frac{\varrho\,x^{-1} - x^{-\varrho}}{\varrho - 1}
  \qquad \forall x \in (0, \infty).
  \label{eq:master-slope-curve}
\end{equation}
\end{corollary}
\noindent\ProofRef{cor:slope-collapse}

\subsection{Pole Confluence and Early-Stopped Gradient Flow}
\label{subsec:gf-holomorphic}

The non-degenerate expansion~\eqref{eq:three-term-expansion} requires the spectral--geometric poles and the regularization filter poles to remain separated. When a spectral pole $s = -p$ approaches a filter pole $s = -mq$ at separation $h = p - mq \to 0$ (such as $\varrho \to 0$ or $\varrho \to 1$ at $m = 2$), we isolate the colliding residue pair via the single-mode integrand $\Phi_p(s; M)$ and the holomorphic regular parts $\Psi_{q,m}^{\mathrm{reg}}(s)$ and $\zeta_{\mathrm{reg}}(s + p)$ obtained after subtracting the simple poles at $s = -mq$ and $s = -p$ (\Cref{lem:app-laurent-collision}):
\begin{gather}
  p \in \{r - 1,\, r + \delta - 1\}, \qquad
  c_p \in \{c,\, cb\}, \qquad
  h \;\Let\; p - mq \quad (m \ge 2), \nonumber\\
  \Phi_p(s; M) \;\Let\; c_p M^s \Psi_q(s)\zeta(p + 1 + s), \qquad
  \Psi_{q,m}^{\mathrm{reg}}(-mq) \;=\; -\frac{(-1)^{m-2}}{q}, \qquad
  \zeta_{\mathrm{reg}}(0) \;=\; 0.
\end{gather}
Although the two individual residues diverge with opposite signs as $h \to 0$, their sum forms a divided difference in $M^{-s}$ that converges smoothly to a logarithmic correction \citep{flajolet1995mellin}.

\begin{proposition}[Divided-Difference Representation and Pole Confluence]
\label{prop:pole-collision}
For any non-zero pole separation $h \neq 0$, the combined residue pair $\mc{C}_{p,m}(M; h)$ of $\Phi_p(s; M)$ at $s = -p$ and $s = -mq$ obeys the divided-difference identity
\begin{align}
  \mc{C}_{p,m}(M; h)
  &\;\Let\; \operatorname*{Res}_{s=-p}\Phi_p(s; M) + \operatorname*{Res}_{s=-mq}\Phi_p(s; M) \nonumber\\
  &\;=\; c_p(-1)^{m-2}(m-1)\frac{M^{-p} - M^{-mq}}{-h}
  + c_p M^{-p}\Psi_{q,m}^{\mathrm{reg}}(-p) \nonumber\\
  &\qquad + c_p(-1)^{m-2}(m-1)M^{-mq}\left(\gamma_E + \zeta_{\mathrm{reg}}(h)\right),
  \label{eq:divided-difference-uniform}
\end{align}
where $\gamma_E$ is the Euler--Mascheroni constant. In the collision limit $h \to 0$ ($p \to mq$), $\mc{C}_{p,m}(M; h)$ converges smoothly to the logarithmic confluence law
\begin{equation}
  \lim_{p \to mq} \mc{C}_{p,m}(M; p - mq)
  \;=\; c_p(-1)^{m-2}(m-1) M^{-mq}\left(\ln M + \gamma_E - \frac{1}{(m-1)q}\right).
  \label{eq:pole-collision-log}
\end{equation}
\end{proposition}
\noindent\ProofRef{prop:pole-collision}

A complementary boundary regime arises when explicit Tikhonov regularization is replaced by early-stopped iterative optimization \citep{yao2007early,ali2019continuous,bordelon2024dynamical,li2025functional}. Under continuous-time gradient flow $\psi_{\mathrm{GF}}(u) = \exp(-2u^{-q})$, the associated Mellin kernel is holomorphic throughout the open left half-plane $\Re(s) < 0$, eliminating all regularization filter poles so that pre-asymptotic scaling is governed exclusively by the spectral--geometric zeta poles.

\begin{proposition}[Gradient-Flow Mellin Kernel]
\label{prop:gf-holomorphic}
Replace the rational filter $\psi_q(u)$ in \Cref{assump:spectral} with the continuous-time gradient-flow residual filter $\psi_{\mathrm{GF}}(u) \Let \exp(-2 u^{-q})$ for $u = m / M$ and $q > 0$. Then the associated Mellin transform
\begin{equation}
  \Psi_{\mathrm{GF}}(s) \;\Let\; \int_0^{\infty} e^{-2 u^{-q}} u^{s-1}\dd u
  \;=\; \frac{2^{s/q}}{q}\,\Gamma\!\left(-\frac{s}{q}\right)
  \label{eq:gf-mellin}
\end{equation}
is holomorphic throughout the open left half-plane $\Re(s) < 0$. Consequently, shifting the Mellin--Barnes contour into $\Re(s) < 0$ encounters no filter poles, and the pre-asymptotic risk expansion contains only the spectral--geometric zeta poles at $s = -(r-1)$ and $s = -(r+\delta-1)$ up to the contour remainder.
\end{proposition}
\noindent\ProofRef{prop:gf-holomorphic}

\subsection{Multi-Power Ill-Conditioning and Difference Identification}
\label{subsec:identifiability}

\Cref{thm:master-mellin} also clarifies an empirical difficulty in scaling-law extrapolation: although the finite-capacity risk contains three power-law terms $A_0 N^{-\alpha_\infty} + A_g N^{-(\alpha_\infty + \Delta_g)} + A_f N^{-(\alpha_\infty + \Delta_f)}$ above the irreducible floor $E_\infty$, fitting a multi-exponential sum directly to a narrow pilot grid is severely ill-conditioned \citep{lanczos1956applied,gautschi1962inverses,varah1985fitting,mlodozeniec2026ambiguous}. On an ordered log-capacity pilot grid $\{u_j\}_{j=1}^J$ with $u_j \Let \ln(N_j / N_1)$, we compare the unconstrained 7-parameter three-power model $f_7(u; \theta_7)$ and the 5-parameter gap-locked model $f_5(u; \theta_5)$ (with fixed spectral gaps $\Delta_g, \Delta_f$ from~\eqref{eq:gap-definitions}), along with their $J \times p$ Jacobian matrices $\bm{J}_p(\theta_p)$:
\begin{gather}
  0 \;\le\; u_1 < u_2 < \dots < u_J, \qquad
  \bm{J}_p(\theta_p) \;\Let\; \left[\frac{\partial f_p(u_j; \theta_p)}{\partial \theta_p}\right]_{j=1}^J \in \R^{J \times p}, \nonumber\\
  \theta_7 \;\Let\; (E_\infty, A_1, A_2, A_3, \alpha_1, \alpha_2, \alpha_3) \in \R^7, \qquad
  \theta_5 \;\Let\; (E_\infty, A_0, A_g, A_f, \alpha_\infty) \in \R^5, \nonumber\\
  f_7(u; \theta_7) \;\Let\; E_\infty + \sum_{k=1}^3 A_k e^{-\alpha_k u}, \nonumber\\
  f_5(u; \theta_5) \;\Let\; E_\infty + A_0 e^{-\alpha_\infty u} + A_g e^{-(\alpha_\infty + \Delta_g)u} + A_f e^{-(\alpha_\infty + \Delta_f)u}.
\end{gather}
While $\bm{J}_7(\theta_7)$ suffers algebraic rank collapse along the single-power locus, even the gap-locked Jacobian $\bm{J}_5(\theta_5)$ suffers a quadratic singular-value decay in the dimensionless pilot span $\Delta_g u_J$ due to second-order Taylor cancellation across its three exponential columns.

\begin{proposition}[Multi-Power Jacobian Rank Collapse and Ill-Conditioning]
\label{prop:identifiability}
On any strictly increasing pilot grid $\{u_j\}_{j=1}^J$ in $[0, \infty)$ with $J \ge 7$, the unconstrained 7-parameter Jacobian $\bm{J}_7(\theta_7) \in \R^{J \times 7}$ is rank-deficient both at exponent confluence and on the single-power submanifold:
\begin{align}
  \alpha_1 = \alpha_2 = \alpha_3 = \alpha_\infty > 0
  &\quad\Longrightarrow\quad
  \operatorname{rank}\!\left(\bm{J}_7(\theta_7)\right) \;\le\; 3 \;<\; 7, \nonumber\\
  A_2 = A_3 = 0
  &\quad\Longrightarrow\quad
  \operatorname{rank}\!\left(\bm{J}_7(\theta_7)\right) \;\le\; 5 \;<\; 7.
  \label{eq:rank-collapse-7}
\end{align}
Furthermore, for the 5-parameter gap-locked model $f_5(u; \theta_5)$ with $\varrho > 1$ and any pilot span $\Delta_g u_J \in (0, 1)$, the smallest singular value and 2-norm condition number of $\bm{J}_5(\theta_5) \in \R^{J \times 5}$ obey the amplitude-independent bounds
\begin{gather}
  \sigma_{\min}\!\left(\bm{J}_5(\theta_5)\right) \;\le\; C_5\,(\Delta_g u_J)^2, \qquad
  \kappa_2\!\left(\bm{J}_5(\theta_5)\right) \;\ge\; \frac{\sqrt{J}}{C_5}\,(\Delta_g u_J)^{-2}, \nonumber\\
  C_5 \;\Let\; \frac{\sqrt{J}\,\varrho(\varrho + 1)}{2\sqrt{(\varrho - 1)^2 + \varrho^2 + 1}} \;>\; 0.
  \label{eq:J5-conditioning-blowup}
\end{gather}
\end{proposition}
\noindent\ProofRef{prop:identifiability}

To mitigate the $(\Delta_g u_J)^{-2}$ conditioning blowup of $\bm{J}_5$ on narrow pilot grids where $\Delta_g u_J < 1$, one must either widen the pilot log-span $u_J = \ln(N_J / N_1)$ across scales \citep{meng2026sldbench} or compress the pre-asymptotic expansion into a three-parameter shifted power law with local effective exponent $\alpha_{\mathrm{eff}} > 0$. Even for the three-parameter Jacobian $\bm{J}_3$ below, fitting $(E_\infty, A_0, \alpha_{\mathrm{eff}})$ jointly in raw coordinates retains a second-order $u_J^2$ cancellation between the constant column $\mathbf{1}$ and the linear Taylor terms of $e^{-\alpha_{\mathrm{eff}} u}$. Applying first-order difference operators across a geometric subgrid eliminates the nuisance offset $E_\infty$ so that the unweighted two-parameter difference Jacobian exhibits linear $u_J$ rather than quadratic $u_J^2$ column collinearity, yielding a closed-form algebraic initializer for $(\alpha_{\mathrm{eff}}, A_0, E_\infty)$ via classical Aitken $\Delta^2$ identification \citep{aitken1926bernoulli}.

\begin{corollary}[Aitken Difference Identification of Shifted Power Laws]
\label{cor:aitken-identifiability}
For $\theta_3 = (E_\infty, A_0, \alpha_{\mathrm{eff}}) \in \R \times \R_{>0} \times \R_{>0}$, consider the 3-parameter effective reduction and its $J \times 3$ Jacobian matrix on any strictly increasing pilot grid $\{u_j\}_{j=1}^J$ in $[0, \infty)$ with $J \ge 3$:
\begin{equation}
  f_3(u; \theta_3) \;\Let\; E_\infty + A_0 e^{-\alpha_{\mathrm{eff}} u}, \qquad
  \bm{J}_3(\theta_3) \;=\; \Bigl[\,\mathbf{1},\;\; e^{-\alpha_{\mathrm{eff}} u},\;\; -A_0 u e^{-\alpha_{\mathrm{eff}} u}\,\Bigr] \in \R^{J \times 3}.
\end{equation}
Then $\bm{J}_3(\theta_3)$ has full column rank $\operatorname{rank}(\bm{J}_3(\theta_3)) = 3$ on any grid with $J \ge 3$ distinct points. Moreover, on any geometric subgrid $u_j = (j-1)\Delta u$, the first-order differences $\Delta y_j \Let f_3(u_{j+1}; \theta_3) - f_3(u_j; \theta_3)$ eliminate $E_\infty$ and uniquely recover $(\alpha_{\mathrm{eff}}, A_0, E_\infty)$ via the Aitken identities
\begin{equation}
  \alpha_{\mathrm{eff}} = -\frac{1}{\Delta u}\ln\!\left(\frac{\Delta y_{j+1}}{\Delta y_j}\right), \quad
  A_0 = \frac{-\Delta y_j}{(1 - e^{-\alpha_{\mathrm{eff}}\Delta u})e^{-\alpha_{\mathrm{eff}} u_j}}, \quad
  E_\infty = y_j - A_0 e^{-\alpha_{\mathrm{eff}} u_j}.
  \label{eq:aitken-locking}
\end{equation}
\end{corollary}
\noindent\ProofRef{cor:aitken-identifiability}

\begin{remark}[Bias--Variance Trade-Off and Octave-Grid Step Size]
\label{rem:aitken-snr}
Consider noisy pilot evaluations $\hat{y}_j = y_j + \varepsilon_j$ with independent Gaussian noise $\varepsilon_j \in \mc{N}(0, \sigma_\varepsilon^2)$ and first-difference operator $\bm{D} \in \R^{(J-1)\times J}$. Generalized least-squares whitening of $\bm{D}\hat{y}$ preserves the exact Fisher information of variable-projection elimination of $E_\infty$, whereas for the unweighted three-point ratio~\eqref{eq:aitken-locking}, conditioning on the bounded-noise event $\mc{A}_j$ in the high-SNR regime $\abs{\Delta y_{j+1}} \ge c_{\mathrm{snr}}\sigma_\varepsilon$ yields the conditional variance expansion (with standard normal CDF $\Phi$):
\begin{gather}
  \bm{D}^{\top}(\bm{D}\bm{D}^{\top})^{-1}\bm{D} = \bm{I}_J - \frac{1}{J}\mathbf{1}\mathbf{1}^{\top}, \quad
  \Delta\hat{y}_j = \Delta y_j + (\varepsilon_{j+1} - \varepsilon_j), \quad
  \Cov(\Delta\hat{y}_j, \Delta\hat{y}_{j+1}) = -\sigma_\varepsilon^2, \nonumber\\
  \mc{A}_j \;\Let\; \Bigl\{\max_{k \in \{j, j+1\}}\abs{\Delta\hat{y}_k - \Delta y_k} \;\le\; \tfrac{1}{2}\abs{\Delta y_{j+1}}\Bigr\}, \qquad
  \PP(\mc{A}_j^c) \;\le\; 4\Phi\!\left(-\frac{\abs{\Delta y_{j+1}}}{2\sqrt{2}\,\sigma_\varepsilon}\right), \nonumber\\
  \Var_{\mc{A}_j}(\hat{\alpha}_{\mathrm{eff}})
  \;=\; \frac{2\sigma_\varepsilon^2\bigl(1 + e^{\alpha_{\mathrm{eff}}\Delta u} + e^{2\alpha_{\mathrm{eff}}\Delta u}\bigr)}{(\Delta u)^2(\Delta y_j)^2}(1 + R_{\mathrm{snr}})
  \;=\; \mc{O}\!\left(\frac{\sigma_\varepsilon^2}{A_0^2\alpha_{\mathrm{eff}}^2 e^{-2\alpha_{\mathrm{eff}}u_j}(\Delta u)^4}\right),
\end{gather}
where the linearization remainder satisfies $\abs{R_{\mathrm{snr}}} \le C_{\mathrm{snr}}\sigma_\varepsilon/\abs{\Delta y_{j+1}}$. Because $\abs{\Delta y_j} = \mc{O}(\Delta u)$ at small step sizes, the leading variance factor scales as $(\Delta u)^{-4}$ within the high-SNR regime; choosing a sufficiently wide geometric spacing (such as octave doubling $\Delta u = \ln 2$) keeps the per-step signal drop above the noise floor and supplies a stable closed-form initializer for full-grid nonlinear least squares.
\end{remark}

\section{Multi-Domain Data Mixing and Sequential Allocation}
\label{sec:mixing}

The second component of the unified risk decomposition~\eqref{eq:unified-risk-decomposition} is the multi-domain sample estimation error $\mc{E}_{\mathrm{mix}}(w; D)$ under \Cref{assump:mixing}, which governs how a finite token budget $D > 0$ is allocated across $K \ge 2$ heterogeneous data domains via mixture weights $w = (w_1, \dots, w_K) \in \Delta_{K-1}^{\circ}$.

In pretraining practice, domain mixtures are often chosen via fixed heuristics, small-scale proxy regression \citep{xie2023doremi,ye2025datamixing,ge2024bimix,liu2025regmix}, or empirical and skill-competition mixture laws \citep{shukor2025scaling,dai2026explaining}, even though corpus deduplication, sample pruning, and multi-epoch repetition substantially alter domain-level sample efficiency \citep{hernandez2021transfer,lee2021deduplication,hernandez2022repetition,sorscher2022pruning}. When individual domains obey distinct sample-scaling exponents $\beta_1, \dots, \beta_K > 0$, however, any static mixture $w$ becomes suboptimal as $D$ grows: fast-learning domains ($\beta_k$ large) saturate early, whereas slow-learning bottleneck domains ($\beta_k$ small) dominate the residual excess risk at large scale. In \Cref{subsec:water-filling}, we show that KKT duality reduces optimal simplex mixing to a one-dimensional water-filling equation and derive explicit finite-budget bottleneck bounds; in \Cref{subsec:online-curriculum}, we derive the horizon-free instantaneous sampling distribution; and in \Cref{subsec:simplex-hessian-metric}, we show that the simplex Hessian metric is canonically dual to this sequential law.

\subsection{Simplex Water-Filling and Finite-Budget Bounds}
\label{subsec:water-filling}

Under \Cref{assump:mixing}, let $c_k \Let \pi_k B_k > 0$ denote the effective weight of domain $k \in \{1, \dots, K\}$, so that the separable simplex excess risk is $\mc{E}_{\mathrm{mix}}(w; D) = \sum_{k=1}^K c_k (w_k D)^{-\beta_k}$. Because each domain risk term is strictly convex on $(0, 1)$ and diverges as $w_k \downarrow 0^+$, interior coercivity and KKT duality \citep{boyd2004convex} reduce the $(K-1)$-dimensional simplex optimization problem to a one-dimensional monotone root-finding equation.

\begin{proposition}[KKT Water-Filling Characterization]
\label{prop:water-filling}
Under \Cref{assump:mixing} with $c_k = \pi_k B_k > 0$ and $\beta_k > 0$ for each domain $k \in \{1, \dots, K\}$, the objective $w \mapsto \mc{E}_{\mathrm{mix}}(w; D)$ is strictly convex on $\Delta_{K-1}^{\circ}$ with strictly positive diagonal Hessian entries and diverges at the boundary:
\begin{equation}
  \frac{\partial^2 \mc{E}_{\mathrm{mix}}(w; D)}{\partial w_k^2}
  \;=\; c_k \beta_k(\beta_k + 1) D^{-\beta_k} w_k^{-(\beta_k + 2)} \;>\; 0,
  \qquad
  \lim_{w_k \downarrow 0^+} \mc{E}_{\mathrm{mix}}(w; D) \;=\; +\infty.
\end{equation}
Consequently, for every finite token budget $D > 0$, the objective $\mc{E}_{\mathrm{mix}}(\cdot; D)$ admits a unique interior minimizer $w\opt(D) \Let \argmin_{w \in \Delta_{K-1}} \mc{E}_{\mathrm{mix}}(w; D) \in \Delta_{K-1}^{\circ}$, given in terms of the unique positive root $\nu = \nu(D) > 0$ of the strictly decreasing scalar water-filling equation
\begin{equation}
  \sum_{k=1}^K \left(\frac{c_k \beta_k D^{-\beta_k}}{\nu}\right)^{\frac{1}{\beta_k + 1}} \;=\; 1
  \;\;\Longleftrightarrow\;\;
  w_k\opt(D) \;=\; \left(\frac{c_k \beta_k D^{-\beta_k}}{\nu(D)}\right)^{\frac{1}{\beta_k + 1}} \;\; \forall k \in \{1, \dots, K\}.
  \label{eq:water-filling-kkt}
\end{equation}
\end{proposition}
\noindent\ProofRef{prop:water-filling}

Eliminating the dual multiplier $\nu(D)$ between any domain $k \in \{1, \dots, K\}$ ($\beta_k \neq \beta_1$) and reference domain $1$ in \Cref{prop:water-filling} motivates defining the crossover budget $D_k\opt$, the normalized budget $\tilde{D}_k$, and the powered share ratio $\Upsilon_k(D)$:
\begin{equation}
  \CrossoverBudgetDef, \qquad
  \NormalizedBudgetDef, \qquad
  \Upsilon_k(D) \;\Let\; \left(\frac{w_k\opt(D)^{\beta_k + 1}}{w_1\opt(D)^{\beta_1 + 1}}\right)^{\frac{1}{\beta_k - \beta_1}}.
  \label{eq:crossover-budget-def}
\end{equation}
In these dimensionless coordinates, the KKT stationarity conditions collapse onto an exact inverse-budget power law across all token scales.

\begin{corollary}[Invariant Powered Share-Ratio Law]
\label{cor:powered-share-ratio}
For every finite token budget $D > 0$ and every domain $k \in \{1, \dots, K\}$ with $\beta_k \neq \beta_1$, the optimal mixture weights from \Cref{prop:water-filling} obey the exact identity
\begin{equation}
  \frac{w_k\opt(D)^{\beta_k + 1}}{w_1\opt(D)^{\beta_1 + 1}} \;=\; \frac{c_k \beta_k}{c_1 \beta_1}\,D^{-(\beta_k - \beta_1)} \;=\; \tilde{D}_k^{-(\beta_k - \beta_1)}
  \quad \Longleftrightarrow \quad
  \PoweredShareRatioDef.
  \label{eq:powered-ratio-law}
\end{equation}
\end{corollary}
\noindent\ProofRef{cor:powered-share-ratio}

To quantify non-asymptotic departures from the large-budget bottleneck limit when one or more domains share the slowest decay exponent, partition the domain index set into the slowest-decaying bottleneck tie set $S$ and its fast-decaying complement $S^c$:
\begin{equation}
  b_0 \;\Let\; \min_{1 \le k \le K} \beta_k \;>\; 0, \quad
  S \;\Let\; \left\{ s \in \{1, \dots, K\} : \beta_s = b_0 \right\}, \quad
  S^c \;\Let\; \{1, \dots, K\} \setminus S.
  \label{eq:slowest-set-def}
\end{equation}
On the bottleneck face $S$, define the normalization constant $H$ and the asymptotic equilibrium shares $\theta_s$ ($s \in S$), and for the fast complement $S^c$ define the exponent ratios $r_k \in (0, 1)$, the individual and total slack variables $z_k(D)$ and $Z(D)$, and the total fast-domain probability mass $\varepsilon(D)$ (with $\varepsilon(D) = Z(D) = 0$ when $S^c = \emptyset$):
\begin{gather}
  H \;\Let\; \sum_{s \in S} \left(b_0 c_s\right)^{\frac{1}{b_0 + 1}}, \qquad
  \theta_s \;\Let\; \frac{\left(b_0 c_s\right)^{\frac{1}{b_0 + 1}}}{H} \quad (s \in S), \qquad
  \varepsilon(D) \;\Let\; \sum_{k \in S^c} w_k\opt(D), \label{eq:tie-aware-H-theta} \\
  r_k \Let \frac{b_0 + 1}{\beta_k + 1} \in (0, 1), \quad
  z_k(D) \Let \left(\beta_k c_k\right)^{\frac{1}{\beta_k + 1}} H^{-r_k} D^{-(1 - r_k)}, \quad
  Z(D) \Let \sum_{k \in S^c} z_k(D). \label{eq:tie-aware-Z-def}
\end{gather}
Expressing the KKT multiplier $\nu(D)$ in terms of the slow-domain mass $1 - \varepsilon(D)$ reduces the multi-domain equilibrium to a scalar fixed-point equation for $\varepsilon(D)$.

\begin{theorem}[Finite-Budget Bottleneck Approximation and Tie-Aware Bounds]
\label{thm:master-mixing}
Under \Cref{assump:mixing}, for every finite token budget $D > 0$, the KKT minimizer $w\opt(D) \in \Delta_{K-1}^{\circ}$ from \Cref{prop:water-filling} satisfies $w_s\opt(D) = \theta_s (1 - \varepsilon(D))$ for all $s \in S$, while the fast-domain probability mass $\varepsilon(D) = \sum_{k \in S^c} w_k\opt(D)$ obeys the fixed-point identity and two-sided sandwich bound
\begin{gather}
  \varepsilon(D) \;=\; \sum_{k \in S^c} z_k(D)\bigl(1 - \varepsilon(D)\bigr)^{r_k}, \nonumber\\
  \frac{Z(D)}{1 + Z(D)} \;\le\; \varepsilon(D) \;\le\; \min\bigl\{Z(D), 1\bigr\} \qquad \forall D > 0.
  \label{eq:eps-sandwich}
\end{gather}
Furthermore, the closed-form plug-in approximation $\tilde{w}(D) \in \R^K$ (which lies inside $\Delta_{K-1}^{\circ}$ whenever $Z(D) < 1$) and its $L_1$ approximation error satisfy
\begin{gather}
  \tilde{w}_k(D) \;\Let\; z_k(D) \;\; (k \in S^c), \qquad
  \tilde{w}_s(D) \;\Let\; \theta_s\bigl(1 - Z(D)\bigr) \;\; (s \in S), \nonumber\\
  \norm{\tilde{w}(D) - w\opt(D)}_1 \;=\; 2\left(Z(D) - \varepsilon(D)\right) \;\le\; \frac{2 Z(D)^2}{1 + Z(D)} \;\le\; 2 Z(D)^2.
  \label{eq:plugin-L1-bound}
\end{gather}
Finally, relative to the leading bottleneck risk $\mc{E}_{\mathrm{lead}}(D) \Let \frac{H^{b_0 + 1}}{b_0} D^{-b_0}$, the optimal value function $\mc{E}_{\mathrm{mix}}\opt(D) \Let \mc{E}_{\mathrm{mix}}(w\opt(D); D)$ satisfies for all $\varepsilon(D) \le 1/2$:
\begin{align}
  \frac{\mc{E}_{\mathrm{mix}}\opt(D)}{\mc{E}_{\mathrm{lead}}(D)}
  &\;=\; \left(1 - \varepsilon(D)\right)^{-(b_0 + 1)}\left[ 1 + \sum_{k \in S^c} \left(\frac{b_0}{\beta_k} - 1\right) w_k\opt(D) \right] \nonumber\\
  &\;=\; 1 + \sum_{k \in S^c} \frac{b_0(\beta_k + 1)}{\beta_k}\,w_k\opt(D) + R_{\mathrm{mix}}(D),
  \label{eq:mix-value-expansion}
\end{align}
where $0 \le R_{\mathrm{mix}}(D) \le C_{\mathrm{mix}}(b_0)\varepsilon(D)^2$ with $C_{\mathrm{mix}}(b_0) \Let (b_0 + 1)(b_0 + 2)2^{b_0 + 2}$.
\end{theorem}
\begin{proof}[Proof Sketch of \Cref{thm:master-mixing}]
Summing the KKT stationarity relation~\eqref{eq:water-filling-kkt} over the slowest-exponent tie set $S$ expresses the dual multiplier $\nu(D)$ in terms of the slow-domain mass $1 - \varepsilon(D)$; substituting $\nu(D)$ into~\eqref{eq:water-filling-kkt} for each $k \in S^c$ with $r_k = \frac{b_0+1}{\beta_k+1} \in (0, 1)$ and applying Bernoulli's inequality yields
\begin{align*}
  \nu(D)^{-1} &\;=\; H^{-(b_0+1)} D^{b_0}\bigl(1 - \varepsilon(D)\bigr)^{b_0+1}
  \quad\Longrightarrow\quad
  w_k\opt(D) \;=\; z_k(D)\bigl(1 - \varepsilon(D)\bigr)^{r_k} \quad \forall k \in S^c, \\
  1 - \varepsilon(D) &\;\le\; \bigl(1 - \varepsilon(D)\bigr)^{r_k} \;\le\; 1
  \quad\Longrightarrow\quad
  Z(D)\bigl(1 - \varepsilon(D)\bigr) \;\le\; \varepsilon(D) \;\le\; Z(D),
\end{align*}
which directly establishes the sandwich inequality~\eqref{eq:eps-sandwich}, the $L_1$ error bound~\eqref{eq:plugin-L1-bound}, and (via \Cref{lem:app-simplex-sandwich}) the value expansion~\eqref{eq:mix-value-expansion}. \ProofRef{thm:master-mixing}
\end{proof}

\begin{remark}[Box-Constrained Water-Filling and Pseudocount Shift]
\label{rem:clipped-pseudocount}
Two practical extensions of \Cref{prop:water-filling} preserve its one-dimensional monotone root-finding structure:
\begin{enumerate}
  \item[(i)] \emph{Box-constrained corpus pools.} When unique tokens in high-utility domains are capped by finite corpus pools $D_{k,\max} > 0$ with $D < \sum_{k=1}^K D_{k,\max}$ \citep{muennighoff2023dataconstrained}, imposing the upper box constraints $w_k \le \bar{w}_k(D)$ yields the clipped water-filling allocation
  \begin{equation}
    \bar{w}_k(D) \;\Let\; \frac{D_{k,\max}}{D}, \qquad
    w_k\opt(D) \;=\; \min\!\left\{\bar{w}_k(D),\; \left(\frac{c_k\beta_k D^{-\beta_k}}{\nu}\right)^{\frac{1}{\beta_k+1}}\right\},
    \label{eq:clipped-water-filling}
  \end{equation}
  where the dual multiplier $\nu > 0$ is uniquely determined by the simplex normalization $\sum_{k=1}^K w_k\opt(D) = 1$.
  \item[(ii)] \emph{Positive pseudocount shift ($w_0 > 0$).} At any fixed target budget $D > 0$, when shared vocabulary and syntax supply an effective cross-domain share offset $w_0 > 0$ at zero explicit weight, replacing $w_k$ by the affine shifted share $\tilde{w}_k$ yields the sparse boundary water-filling law
  \begin{equation}
    \tilde{w}_k \;\Let\; w_k + w_0, \qquad
    w_k\opt(D) \;=\; \max\!\left\{0,\; \left(\frac{c_k\beta_k D^{-\beta_k}}{\nu}\right)^{\frac{1}{\beta_k+1}} - w_0\right\},
    \label{eq:pseudocount-water-filling}
  \end{equation}
  explaining why finite $w_0 > 0$ induces boundary sparsity ($w_k\opt(D) = 0$) on low-priority domains (whereas sequential monotonicity across scales in \Cref{prop:curriculum-rn} requires either unshifted shares $w_0 = 0$ or a fixed additive token prior $n_{k,0} > 0$).
\end{enumerate}
\end{remark}

\subsection{Sequential Allocation and Instantaneous Sampling}
\label{subsec:online-curriculum}

\Cref{prop:water-filling,thm:master-mixing} characterize the optimal cumulative mixture $w\opt(D)$ at any total budget $D > 0$. When training proceeds sequentially from $t = 0$ to $t = D$ via online curriculum sampling or mid-training domain annealing \citep{xie2023doremi,ge2024bimix,bi2024deepseek}, one seeks a sequential sampling policy $q\opt(D) \in \Delta_{K-1}^{\circ}$ whose cumulative domain token counts $n_k^\star(D) = D w_k^\star(D)$ match the optimal cumulative allocation at every intermediate checkpoint $D > 0$ without fixing the final stopping horizon in advance. Differentiating $n\opt(D)$ with respect to $D$ via implicit differentiation of~\eqref{eq:water-filling-kkt} shows that every component $n_k\opt(D)$ is strictly increasing, so the trajectory $\{w\opt(D)\}_{D > 0}$ is realized sequentially by an instantaneous sampling distribution $q\opt(D)$ that tilts $w\opt(D)$ toward slower-decaying domains by the factor $(\beta_k + 1)^{-1}$.

\begin{proposition}[Sequential Allocation and Instantaneous Sampling]
\label{prop:curriculum-rn}
Let $w\opt(D) \in \Delta_{K-1}^{\circ}$ be the optimal cumulative mixture from \Cref{prop:water-filling}, and let $n_k\opt(D) = D w_k\opt(D)$ denote the optimal cumulative token allocation to domain $k$ at budget $D > 0$. Then for every $D > 0$ and every domain $k \in \{1, \dots, K\}$, the marginal token allocation rate $q_k\opt(D)$ is strictly positive and defines a probability distribution $q\opt(D) \in \Delta_{K-1}^{\circ}$ satisfying
\begin{gather}
  q_k\opt(D) \;\Let\; \frac{\dd n_k\opt(D)}{\dd D}
  \;=\; \frac{\dfrac{w_k\opt(D)}{\beta_k + 1}}{\displaystyle\sum_{j=1}^K \dfrac{w_j\opt(D)}{\beta_j + 1}}
  \qquad \forall k \in \{1, \dots, K\}, \nonumber\\
  \int_0^D q_k\opt(t)\dd t \;=\; D w_k\opt(D), \qquad
  \mc{E}_{\mathrm{mix}}\!\left(\frac{1}{D}\int_0^D q\opt(t)\dd t; D\right) - \mc{E}_{\mathrm{mix}}\opt(D) \;=\; 0.
  \label{eq:instantaneous-curriculum}
\end{gather}
Equivalently, viewing $w\opt(D)$ and $q\opt(D)$ as probability measures $\PP_D\opt$ and $\QQ_D\opt$ on $\{1, \dots, K\}$, the Radon--Nikodym derivative of $\QQ_D\opt$ with respect to $\PP_D\opt$ and the relative tilt between any two domains $\beta_j < \beta_k$ are given by
\begin{equation}
  \frac{\dd \QQ_D\opt}{\dd \PP_D\opt}(k) \;=\; \frac{q_k\opt(D)}{w_k\opt(D)} \;=\; \frac{(\beta_k + 1)^{-1}}{\E_{\PP_D\opt}\!\left[(\beta_I + 1)^{-1}\right]}, \qquad
  \frac{q_j\opt(D) / w_j\opt(D)}{q_k\opt(D) / w_k\opt(D)} \;=\; \frac{\beta_k + 1}{\beta_j + 1} \;>\; 1,
  \label{eq:radon-nikodym-tilt}
\end{equation}
so the sequential allocation policy upweights slower-learning bottleneck domains at the margin.
\end{proposition}
\noindent\ProofRef{prop:curriculum-rn}

\subsection{Simplex Hessian Metric and Cross-Scale Transfer}
\label{subsec:simplex-hessian-metric}

Evaluating the diagonal Hessian matrix $\bm{D}_H(D)$ of $\mc{E}_{\mathrm{mix}}(w; D)$ at the KKT solution $w\opt(D)$ induces a weighted Riemannian metric $g_{w\opt(D)}$ on the simplex tangent space $T_{w\opt(D)}\Delta_{K-1}$ \citep{amari2016information,martens2020natural}:
\begin{gather}
  \bm{D}_H(D) \;\Let\; \nabla^2 \mc{E}_{\mathrm{mix}}\bigl(w\opt(D); D\bigr), \qquad
  T_{w\opt(D)}\Delta_{K-1} \;\Let\; \left\{\delta w \in \R^K : \sum_{k=1}^K \delta w_k = 0\right\}, \nonumber\\
  g_{w\opt(D)}(\delta w, \delta\tilde{w}) \;\Let\; \ip{\delta w}{\bm{D}_H(D)\,\delta\tilde{w}}_{\R^K}.
\end{gather}
Substituting the KKT stationarity condition~\eqref{eq:water-filling-kkt} into $\bm{D}_H(D)$ eliminates the power-law factors $c_k D^{-\beta_k}$ and shows that $\bm{D}_H(D)^{-1}$ is canonically proportional to the instantaneous sampling distribution $q\opt(D)$ from \Cref{prop:curriculum-rn}.

\begin{proposition}[Simplex Hessian Metric and Sequential Duality]
\label{prop:simplex-hessian}
At the KKT minimizer $w\opt(D) \in \Delta_{K-1}^{\circ}$ with dual multiplier $\nu(D) > 0$, the induced Riemannian metric $g_{w\opt(D)}$ on $T_{w\opt(D)}\Delta_{K-1}$, the diagonal entries and minimum eigenvalue of $\bm{D}_H(D)$, and its inverse $\bm{D}_H(D)^{-1}$ satisfy
\begin{align}
  g_{w\opt(D)}(\delta w, \delta\tilde{w})
  &\;=\; \nu(D)\sum_{k=1}^K (\beta_k + 1)\,\frac{\delta w_k\,\delta\tilde{w}_k}{w_k\opt(D)}, \nonumber\\
  \bigl(\bm{D}_H(D)\bigr)_{kk} &\;=\; \nu(D)\frac{\beta_k+1}{w_k\opt(D)} \;\ge\; \lambda_{\min}\!\bigl(\bm{D}_H(D)\bigr) \;\ge\; \nu(D)(b_0 + 1) \;>\; 0, \nonumber\\
  \bm{D}_H(D)^{-1} &\;=\; \frac{\E_{\PP_D\opt}\!\left[(\beta_I + 1)^{-1}\right]}{\nu(D)}\,\diag\!\left(q\opt(D)\right),
  \label{eq:simplex-hessian-metric}
\end{align}
where $q\opt(D)$ is the sequential allocation distribution in~\eqref{eq:instantaneous-curriculum}. Moreover, for any perturbed mixture $w = w\opt(D) + \delta w \in \Delta_{K-1}^{\circ}$ with componentwise lower envelope $\underline{w}_k \Let \min\{w_k\opt(D), w_k\}$, the excess risk obeys
\begin{gather}
  \mc{E}_{\mathrm{mix}}\!\left(w\opt(D) + \delta w; D\right) - \mc{E}_{\mathrm{mix}}\opt(D)
  \;=\; \frac{1}{2}g_{w\opt(D)}(\delta w, \delta w) + R_3(\delta w), \nonumber\\
  \abs{R_3(\delta w)} \;\le\; \frac{1}{6}\,\nu(D)\sum_{k=1}^K (\beta_k + 1)(\beta_k + 2)\,\frac{w_k\opt(D)^{\beta_k + 1}}{\underline{w}_k^{\beta_k + 3}}\abs{\delta w_k}^3.
  \label{eq:simplex-R3-bound}
\end{gather}
Consequently, on the relative trust neighborhood $\frac{1}{2}w_k\opt(D) \le \hat{w}_k \le 2w_k\opt(D)$ ($\forall k$), any surrogate mixture $\hat{w} = w\opt(D) + \delta w \in \Delta_{K-1}^{\circ}$ with tangent gradient error at most $\varepsilon_{\mathrm{mix}}$ in the inverse metric $\bm{D}_H(D)^{-1}$ obeys the quadratic mixture regret bound
\begin{gather}
  \norm{\nabla\mc{E}_{\mathrm{mix}}(\hat{w}; D) + \nu(D)\mathbf{1}}_{\bm{D}_H(D)^{-1}} \;\le\; \varepsilon_{\mathrm{mix}}
  \quad\Longrightarrow\quad
  \mc{E}_{\mathrm{mix}}(\hat{w}; D) - \mc{E}_{\mathrm{mix}}\opt(D) \;\le\; \frac{\overline{\kappa}_w}{2}\,\varepsilon_{\mathrm{mix}}^2, \nonumber\\
  \overline{\kappa}_w \;\Let\; \max_{1 \le k \le K} 2^{\beta_k + 2}.
\end{gather}
\end{proposition}
\noindent\ProofRef{prop:simplex-hessian}

When shared representations induce cross-domain transfer or interference via a quadratic coupling matrix $\bm{\Gamma} \succeq 0$ as in \Cref{assump:mixing} \citep{hernandez2021transfer,wu2026datamixtures}, the Hessian--sampling duality~\eqref{eq:simplex-hessian-metric} yields a closed-form first-order expansion for the coupled simplex minimizer $w_{\bm{\Gamma}}\opt(D)$ at any fixed finite budget $D > 0$ in the weak-coupling regime where $\norm{\bm{\Gamma}}_{\mathrm{op}} / \nu(D)$ is small (whereas if $\bm{\Gamma} \neq \bm{0}$ is held fixed as $D \to \infty$, $\nu(D)$ decays as $D^{-b_0}$ and the quadratic coupling eventually dominates).

\begin{corollary}[First-Order Perturbation Under Cross-Domain Coupling]
\label{thm:fisher-transfer}
For each fixed finite budget $D > 0$ and any symmetric positive semidefinite coupling matrix $\bm{\Gamma} \succeq 0$ in a neighborhood of $\bm{\Gamma} = \bm{0}$, the coupled simplex minimizer $w_{\bm{\Gamma}}\opt(D)$ obeys the first-order perturbation expansion (with a finite constant $C_{\mathrm{coup}}(D) > 0$ independent of $\bm{\Gamma}$)
\begin{gather}
  w_{\bm{\Gamma}}\opt(D)
  \;\Let\; \argmin_{w \in \Delta_{K-1}} \left\{\mc{E}_{\mathrm{mix}}(w; D) + \frac{1}{2}\ip{w}{\bm{\Gamma} w}_{\R^K}\right\}
  \;=\; w\opt(D) + \Delta w_{\mathrm{nat}}(D) + R_{\bm{\Gamma}}(D), \nonumber\\
  \Delta w_{\mathrm{nat}}(D)
  \;\Let\; -\left(\bm{I}_K - \frac{\bm{D}_H(D)^{-1}\mathbf{1}\mathbf{1}^{\top}}{\mathbf{1}^{\top}\bm{D}_H(D)^{-1}\mathbf{1}}\right)\bm{D}_H(D)^{-1}\bm{\Gamma} w\opt(D), \quad
  \norm{R_{\bm{\Gamma}}(D)}_2 \le C_{\mathrm{coup}}(D)\norm{\bm{\Gamma}}_{\mathrm{op}}^2,
  \label{eq:natural-grad-shift}
\end{gather}
whose $k$-th component simplifies via~\eqref{eq:simplex-hessian-metric} to the centered covariance identity under $\QQ_D\opt$:
\begin{align}
  \Delta w_{\mathrm{nat},k}(D)
  &\;=\; -\frac{1}{\nu(D)}\,\frac{w_k\opt(D)}{\beta_k + 1}\left[ \left(\bm{\Gamma} w\opt(D)\right)_k - \frac{\displaystyle\sum_{j=1}^K \frac{w_j\opt(D)}{\beta_j + 1}\left(\bm{\Gamma} w\opt(D)\right)_j}{\displaystyle\sum_{j=1}^K \frac{w_j\opt(D)}{\beta_j + 1}} \right] \nonumber\\
  &\;=\; -\frac{\E_{\PP_D\opt}\!\left[(\beta_I + 1)^{-1}\right]}{\nu(D)}\,q_k\opt(D)\Bigl[\bigl(\bm{\Gamma} w\opt(D)\bigr)_k - \E_{\QQ_D\opt}\!\bigl[\bigl(\bm{\Gamma} w\opt(D)\bigr)_I\bigr]\Bigr].
  \label{eq:natural-grad-component}
\end{align}
\end{corollary}
\noindent\ProofRef{thm:fisher-transfer}

\section{Compute-Optimal Allocation and Exponent Drift}
\label{sec:compute}

We now couple the finite-capacity approximation error of \Cref{sec:spectral} and the multi-domain sample estimation error of \Cref{sec:mixing} across the two-dimensional log-resource plane $(u, v) = (\ln N, \ln D) \in \R^2$, foliated under \Cref{assump:compute} by the iso-compute submanifolds $\mc{M}_t = \{(u, v) \in \R^2 : u + v = t\}$ with $t = \ln(C / \kappa_c)$.

A recurring question in neural scaling is the discrepancy between the parameter-heavy compute allocation exponent reported by \citet{kaplan2020scaling} ($\eta = 0.73$) and the balanced allocation exponent reported by \citet{hoffmann2022chinchilla} ($\eta = 0.50$), together with subsequent shifts observed at multi-trillion-token budgets \citep{bi2024deepseek,besiroglu2024chinchilla,pearce2024reconciling,porian2024resolving}. While small-scale warm-up offsets and uncalibrated learning-rate schedules distort pilot trajectories, a basic analytical question remains: \emph{when is the compute-optimal allocation exponent $\eta(t) \Let \frac{\dd u\opt(t)}{\dd t}$ invariant across budgets $t$, and how do the multi-term spectral and domain expansions of \Cref{sec:spectral,sec:mixing} cause $\eta(t)$ to drift across scales?}

We first show that monotone loss transformations preserve a constant allocation exponent whereas discrete-time gradient descent with a frozen step size breaks this invariance (\Cref{subsec:invariance-counterexample}). We then derive the iso-compute curvature differential equation and exponent-variance identity for multi-power risk expansions (\Cref{subsec:gibbs-drift}), and conclude with the hyperparameter envelope and quadratic allocation regret bounds (\Cref{subsec:envelope-regret}).

\subsection{Scale Invariance Under Monotone Loss Transformations}
\label{subsec:invariance-counterexample}

It is often assumed that any cross-derivative coupling $\mc{E}_{uv} \neq 0$ between log-capacity $u = \ln N$ and log-data $v = \ln D$ causes the compute-optimal exponent $\eta(t)$ to drift with $t$. In fact, cross-derivative coupling alone does not induce path drift: applying any strictly increasing differentiable transformation $F : \R_{>0} \to \R_{>0}$ ($F'(z) > 0$) to a single power-law sum $z(u, v)$---such as nonlinear downstream evaluation metrics that distort cardinal loss curves \citep{wei2022emergent,schaeffer2023mirage}---defines the non-separable risk
\begin{gather}
  \mc{E}(N, D) \;\Let\; F\bigl(z(u, v)\bigr), \qquad
  z(u, v) \;\Let\; c_1 N^{-a} + c_2 D^{-b} \;=\; c_1 e^{-au} + c_2 e^{-bv}, \nonumber\\
  \mc{E}_{uv}(u, v) \;=\; a b c_1 c_2 e^{-au - bv} F''\bigl(z(u, v)\bigr),
  \label{eq:single-aggregator-model}
\end{gather}
where $c_1, c_2, a, b > 0$. Although $\mc{E}_{uv} \neq 0$ whenever $F''(z) \neq 0$, the positive scalar factor $F'(z) > 0$ cancels identically in the marginal rate of substitution along $\mc{M}_t$, preserving a constant allocation exponent across all compute budgets.

\begin{proposition}[Scale Invariance Under Monotone Loss Transformations]
\label{thm:invariance-counterexample}
For the transformed risk $\mc{E}(N, D) = F(z(u, v))$ in~\eqref{eq:single-aggregator-model} with $F'(z) > 0$, under the monomial compute constraint $C = \kappa_c N^m D^n$ ($m, n > 0$), the unique compute-optimal trajectory $(N\opt(C), D\opt(C)) \Let \argmin_{\kappa_c N^m D^n = C} \mc{E}(N, D)$ obeys the scale-invariant power laws
\begin{equation}
  N\opt(C) \;=\; A_N C^{\eta_N}, \quad D\opt(C) \;=\; A_D C^{\eta_D}, \quad
  \eta_N \;=\; \frac{b}{mb + na}, \quad \eta_D \;=\; \frac{a}{mb + na},
  \label{eq:invariant-exponents}
\end{equation}
for every $C > 0$, independent of $F$. In particular, for $m = n = 1$ ($C = \kappa_c N D$), the log-parameter allocation rate $\eta(t) = \frac{b}{a+b}$ is constant ($\eta'(t) \equiv 0$) for all $t \in \R$.
\end{proposition}
\noindent\ProofRef{thm:invariance-counterexample}

\begin{remark}[Step-Size Sensitivity Under Loss Rescaling in Discrete Gradient Descent]
\label{rem:gd-rescaling}
While \Cref{thm:invariance-counterexample} establishes invariance under monotone loss wrappers at the static risk-minimizing envelope, discrete-time optimization with a frozen step size breaks this invariance even under a positive linear rescaling of a single quadratic mode \citep{paquette2024phases,li2025functional,ramani2026optimizer}. Running $T \in \N_{\ge 1}$ steps of gradient descent with a fixed learning rate $\xi_0 = \alpha > 0$ from $\tilde{\theta}_0 = \theta_0 \neq 0$ (for $\alpha\lambda \notin \{1, \frac{2}{k+1}\}$) yields
\begin{gather}
  \ell(\theta) \;\Let\; \frac{1}{2}\lambda\theta^2 \;\; (\lambda > 0), \qquad
  \tilde{\ell}(\theta) \;\Let\; k\,\ell(\theta) \;\; (k > 0,\; k \neq 1), \nonumber\\
  \theta_T \;=\; (1 - \alpha\lambda)^T\theta_0, \qquad
  \tilde{\theta}_T \;=\; (1 - \alpha k\lambda)^T\theta_0, \qquad
  \frac{\ell(\tilde{\theta}_T)}{\ell(\theta_T)} \;=\; \left(\frac{1 - \alpha k\lambda}{1 - \alpha\lambda}\right)^{2T} \;\neq\; 1.
  \label{eq:gd-rescaling-ratio}
\end{gather}
In particular, whenever $\alpha\lambda \in (0, 2)$ while $\alpha k\lambda > 2$, the ratio~\eqref{eq:gd-rescaling-ratio} diverges exponentially as $T \to \infty$ unless the step size tracks the calibrated schedule $\tilde{\alpha} = \alpha / k$ (foreshadowing the Schur-complement hyperparameter envelope in \Cref{thm:master-envelope}).
\end{remark}

\subsection{Iso-Compute Curvature and Exponent Variance}
\label{subsec:gibbs-drift}

Because a single power-law pair preserves a constant allocation exponent (\Cref{thm:invariance-counterexample}), scale drift of $\eta(t)$ requires competing decay modes across capacity or data---precisely the multi-term Mellin expansion along log-capacity $u$ (\Cref{thm:master-mellin}) and the multi-domain expansions along log-data $v$ (\Cref{thm:master-mixing}). Under \Cref{assump:compute}, each iso-compute line $\mc{M}_t$ admits a unique interior minimizer $(u\opt(t), v\opt(t))$ characterized by marginal-risk stationarity and complementary tangent allocation rates:
\begin{gather}
  \bigl(u\opt(t), v\opt(t)\bigr) \;\Let\; \argmin_{(u, v) \in \mc{M}_t} \mc{E}(u, v), \qquad
  \mc{E}_u\bigl(u\opt(t), v\opt(t)\bigr) \;=\; \mc{E}_v\bigl(u\opt(t), v\opt(t)\bigr), \nonumber\\
  \eta(t) \;\Let\; \frac{\dd u\opt(t)}{\dd t}, \qquad
  1 - \eta(t) \;=\; \frac{\dd v\opt(t)}{\dd t}.
  \label{eq:optimal-path-stationarity}
\end{gather}
Differentiating the stationarity condition in~\eqref{eq:optimal-path-stationarity} along $u\opt(t) + v\opt(t) = t$ yields the iso-compute curvature differential equation for $\eta(t)$.

\begin{proposition}[Iso-Compute Curvature Differential Equation]
\label{prop:curvature-ode}
Under \Cref{assump:compute}, along the compute-optimal trajectory $(u\opt(t), v\opt(t)) \in \mc{M}_t$, the allocation exponent $\eta(t)$ and the tangent directional derivative operator $\mc{D}$ are given by
\begin{equation}
  \eta(t) \;=\; \frac{\mc{E}_{vv} - \mc{E}_{uv}}{\mc{E}_{uu} - 2\mc{E}_{uv} + \mc{E}_{vv}}\Bigg|_{(u\opt(t), v\opt(t))}, \qquad
  \mc{D} \;\Let\; \eta(t)\partial_u + \bigl(1 - \eta(t)\bigr)\partial_v.
  \label{eq:eta-general-hessian}
\end{equation}
Whenever $\eta(t) \in (0, 1)$, its scale derivative obeys the curvature drift ODE
\begin{align}
  \eta'(t) &\;=\; \eta(t)\bigl(1 - \eta(t)\bigr)\mc{K}(t), \nonumber\\
  \mc{K}(t) &\;\Let\; \frac{\dd}{\dd t}\ln\!\left(\frac{\eta(t)}{1 - \eta(t)}\right)
  \;=\; \left[\frac{\mc{D}(\mc{E}_{vv} - \mc{E}_{uv})}{\mc{E}_{vv} - \mc{E}_{uv}} - \frac{\mc{D}(\mc{E}_{uu} - \mc{E}_{uv})}{\mc{E}_{uu} - \mc{E}_{uv}}\right]_{(u\opt(t), v\opt(t))}.
  \label{eq:eta-prime-general}
\end{align}
\end{proposition}
\noindent\ProofRef{prop:curvature-ode}

For a separable risk surface $\mc{E}(u, v) = U(u) + V(v)$ with strictly convex marginals, the log-odds drift rate $\mc{K}(t)$ in~\eqref{eq:eta-prime-general} reduces to the difference between the third-to-second derivative ratios of $U$ and $V$ (valid for arbitrary residue signs, including $A_f < 0$). When $U$ and $V$ are positive-weight multi-power potentials, normalizing their marginal derivatives defines Gibbs probability measures $\PP_t$ on $\{1, \dots, I\}$ and $\QQ_t$ on $\{1, \dots, K\}$ along $(u\opt(t), v\opt(t))$:
\begin{gather}
  U(u) \;=\; \sum_{i=1}^I A_i e^{-\alpha_i u}, \qquad
  V(v) \;=\; \sum_{j=1}^K B_j e^{-\beta_j v} \qquad
  (A_i, B_j, \alpha_i, \beta_j > 0), \nonumber\\
  \PP_t(i) \;\Let\; \frac{A_i \alpha_i e^{-\alpha_i u\opt(t)}}{\sum_{m=1}^I A_m \alpha_m e^{-\alpha_m u\opt(t)}}, \qquad
  \QQ_t(j) \;\Let\; \frac{B_j \beta_j e^{-\beta_j v\opt(t)}}{\sum_{\ell=1}^K B_{\ell} \beta_{\ell} e^{-\beta_{\ell} v\opt(t)}}.
  \label{eq:gibbs-measures-def}
\end{gather}
Under $\PP_t$ and $\QQ_t$, the log-odds drift rate $\mc{K}(t)$ equals the difference between the normalized Gibbs exponent variances of the capacity and data spectra.

\begin{theorem}[Exponent-Variance Identity for Compute-Optimal Drift]
\label{thm:master-drift}
Under \Cref{assump:compute}, for any separable risk surface $\mc{E}(u, v) = U(u) + V(v)$ with $U''(u) > 0$ and $V''(v) > 0$, the allocation exponent $\eta(t)$ and log-odds drift rate $\mc{K}(t)$ in \Cref{prop:curvature-ode} satisfy
\begin{align}
  \eta(t) &\;=\; \frac{V''(v\opt(t))}{U''(u\opt(t)) + V''(v\opt(t))} \;\in\; (0, 1), \nonumber\\
  \mc{K}(t) &\;=\; \eta(t)\!\left(-\frac{U'''(u\opt(t))}{U''(u\opt(t))}\right) - \bigl(1 - \eta(t)\bigr)\!\left(-\frac{V'''(v\opt(t))}{V''(v\opt(t))}\right).
  \label{eq:K-third-derivative}
\end{align}
Furthermore, for the fixed-weight multi-power potentials and marginal-gradient Gibbs measures $(\PP_t, \QQ_t)$ in~\eqref{eq:gibbs-measures-def}, the effective local exponents and $\mc{K}(t)$ obey the Gibbs moment and variance-to-mean identities
\begin{gather}
  \alpha_{\mathrm{eff}}(t) \;\Let\; \frac{U''(u\opt(t))}{-U'(u\opt(t))} \;=\; \E_{\PP_t}[\alpha], \qquad
  \beta_{\mathrm{eff}}(t) \;\Let\; \frac{V''(v\opt(t))}{-V'(v\opt(t))} \;=\; \E_{\QQ_t}[\beta], \nonumber\\
  \eta(t) \;=\; \frac{\E_{\QQ_t}[\beta]}{\E_{\PP_t}[\alpha] + \E_{\QQ_t}[\beta]}, \qquad
  \mc{K}(t) \;=\; \frac{\eta(t)\,\Var_{\PP_t}(\alpha)}{\E_{\PP_t}[\alpha]} - \frac{\bigl(1 - \eta(t)\bigr)\,\Var_{\QQ_t}(\beta)}{\E_{\QQ_t}[\beta]}.
  \label{eq:gibbs-variance-identity}
\end{gather}
\end{theorem}
\begin{proof}[Proof Sketch of \Cref{thm:master-drift}]
Expressing the second- and third-derivative ratios of $U$ and $V$ as cumulants of the marginal-gradient Gibbs measures $\PP_t$ and $\QQ_t$ (\Cref{lem:app-gibbs-cumulants}) yields the cumulant identities and first-moment balance along $(u\opt(t), v\opt(t))$:
\begin{align*}
  -\frac{U'''(u\opt(t))}{U''(u\opt(t))} &\;=\; \E_{\PP_t}[\alpha] + \frac{\Var_{\PP_t}(\alpha)}{\E_{\PP_t}[\alpha]}, \qquad
  -\frac{V'''(v\opt(t))}{V''(v\opt(t))} \;=\; \E_{\QQ_t}[\beta] + \frac{\Var_{\QQ_t}(\beta)}{\E_{\QQ_t}[\beta]}, \\
  \eta(t)\E_{\PP_t}[\alpha] &\;-\; \bigl(1 - \eta(t)\bigr)\E_{\QQ_t}[\beta] \;=\; 0.
\end{align*}
Substituting this cumulant decomposition into~\eqref{eq:K-third-derivative} cancels the first-moment terms identically and isolates~\eqref{eq:gibbs-variance-identity}. \ProofRef{thm:master-drift}
\end{proof}

When the data potential represents the KKT water-filling envelope $V(v) \Let \mc{E}_{\mathrm{mix}}\opt(e^v)$ optimized over domain weights at each token budget $D = e^v$ (\Cref{prop:water-filling,thm:master-mixing}), differentiating the envelope multiplier $-V'(v) = \nu(e^v)$ via \Cref{prop:curriculum-rn} connects Axis~II directly to Axis~III.

\begin{proposition}[Water-Filling Envelope Curvature and Inverse-Harmonic Drift]
\label{prop:water-filling-drift}
Let $V(v) \Let \mc{E}_{\mathrm{mix}}\opt(e^v)$ denote the optimal multi-domain water-filling envelope of \Cref{prop:water-filling} at log-token budget $v = \ln D$, with optimal domain share law $\PP_{e^v}\opt(k) \Let w_k\opt(e^v)$. Defining the inverse-exponent weight $\rho_k$ and its first moment $m(v)$, the effective local data exponent $\beta_{\mathrm{eff}}\opt(v)$ and third-to-second derivative ratio obey the exact variance-decay identities
\begin{gather}
  \rho_k \;\Let\; (\beta_k + 1)^{-1}, \qquad
  m(v) \;\Let\; \E_{\PP_{e^v}\opt}[\rho_I] \;\in\; (0, 1), \qquad
  \beta_{\mathrm{eff}}\opt(v) \;\Let\; \frac{V''(v)}{-V'(v)} \;=\; \frac{1}{m(v)} - 1, \nonumber\\
  \frac{\dd\beta_{\mathrm{eff}}\opt(v)}{\dd v} \;=\; -\frac{\Var_{\PP_{e^v}\opt}(\rho_I)}{m(v)^3} \;\le\; 0, \qquad
  -\frac{V'''(v)}{V''(v)} \;=\; \beta_{\mathrm{eff}}\opt(v) + \frac{\Var_{\PP_{e^v}\opt}(\rho_I)}{m(v)^2\bigl(1 - m(v)\bigr)}.
  \label{eq:water-filling-drift-identity}
\end{gather}
Consequently, when $V(v) = \mc{E}_{\mathrm{mix}}\opt(e^v)$ is coupled with a single-power capacity potential $U(u) = A_0 e^{-\alpha_0 u}$ along $u + v = t$, the compute-optimal allocation exponent $\eta(t) = \frac{\beta_{\mathrm{eff}}\opt(v\opt(t))}{\alpha_0 + \beta_{\mathrm{eff}}\opt(v\opt(t))}$ obeys the log-odds drift equation
\begin{equation}
  \mc{K}(t) \;=\; -\bigl(1 - \eta(t)\bigr)\frac{\Var_{\PP_{e^{v\opt(t)}}\opt}(\rho_I)}{m\bigl(v\opt(t)\bigr)^2\bigl(1 - m(v\opt(t))\bigr)} \;\le\; 0.
  \label{eq:water-filling-K-drift}
\end{equation}
\end{proposition}
\noindent\ProofRef{prop:water-filling-drift}

In a two-stratum crossover where a two-mode capacity potential $U(u)$ ($0 < \alpha_1 < \alpha_2$) couples with a single-mode data potential $V(v)$, the faster-decaying mode $\alpha_2$ saturates first as $t$ increases, driving $\eta(t)$ upward between asymptotic plateaus $\eta_2 < \eta_1$ across its midpoint scale $t_\star$. Centering log-compute at $\tau = t - t_\star$, the normalized transition coordinate $s(t)$ obeys an exact variable-rate log-odds law in the intrinsic coordinate $z_{\mathrm{int}}(t)$ alongside a constant-rate logistic reference profile $s_{\mathrm{logistic}}(\tau)$:
\begin{gather}
  U(u) \;=\; A_1 e^{-\alpha_1 u} + A_2 e^{-\alpha_2 u}, \qquad
  V(v) \;=\; B e^{-\beta v}, \qquad
  \eta_i \;\Let\; \frac{\beta}{\alpha_i + \beta} \quad (i \in \{1, 2\}), \nonumber\\
  \tau \;\Let\; t - t_\star, \qquad
  \eta(t_\star) \;=\; \bar{\eta} \;\Let\; \frac{\eta_1 + \eta_2}{2}, \qquad
  s(t) \;\Let\; \frac{\eta(t) - \eta_2}{\eta_1 - \eta_2} \;\in\; (0, 1), \nonumber\\
  \lim_{t \to -\infty}\eta(t) \;=\; \eta_2, \qquad
  \lim_{t \to +\infty}\eta(t) \;=\; \eta_1, \qquad
  z_{\mathrm{int}}(t) \;\Let\; (\alpha_2 - \alpha_1)\bigl(u\opt(t) - u\opt(t_\star)\bigr), \nonumber\\
  \frac{\dd}{\dd t}\ln\!\left(\frac{s(t)}{1 - s(t)}\right) = (\alpha_2 - \alpha_1)\eta(t) = \frac{\dd z_{\mathrm{int}}(t)}{\dd t} > 0, \quad
  \frac{\dd}{\dd\tau}\ln\!\left(\frac{s_{\mathrm{logistic}}(\tau)}{1 - s_{\mathrm{logistic}}(\tau)}\right) = \gamma > 0.
  \label{eq:two-stratum-boundary}
\end{gather}

\begin{corollary}[Two-Stratum Logistic Transition Law]
\label{cor:logistic-transition}
Under the two-stratum setup~\eqref{eq:two-stratum-boundary}, the constant-rate reference coordinate $s_{\mathrm{logistic}}(\tau) \in (0, 1)$ with $s_{\mathrm{logistic}}(0) = \frac{1}{2}$ satisfies $\frac{\dd s_{\mathrm{logistic}}}{\dd\tau} = \gamma s_{\mathrm{logistic}}(1 - s_{\mathrm{logistic}})$, yielding the closed-form logistic profile and peak drift speed
\begin{gather}
  \eta_{\mathrm{logistic}}(\tau) \;\Let\; \eta_2 + \frac{\eta_1 - \eta_2}{1 + e^{-\gamma\tau}}, \qquad
  s_{\mathrm{logistic}}(\tau) \;=\; \sigma(\gamma\tau) \;\Let\; \frac{1}{1 + e^{-\gamma\tau}}, \nonumber\\
  \max_{\tau \in \R}\abs{\eta_{\mathrm{logistic}}'(\tau)} \;=\; \frac{\gamma}{4}\abs{\eta_1 - \eta_2}.
  \label{eq:logistic-drift-law}
\end{gather}
Furthermore, for the exact two-mode system in~\eqref{eq:two-stratum-boundary}, the normalized state satisfies $s(t) = \sigma(z_{\mathrm{int}}(t))$ identically in the intrinsic coordinate $z_{\mathrm{int}}(t)$, and with midpoint rate $\gamma \Let (\alpha_2 - \alpha_1)\bar{\eta} > 0$ obeys the non-asymptotic uniform approximation bound in $t$:
\begin{equation}
  \left\lvert \eta(t) - \eta_{\mathrm{logistic}}(t - t_\star) \right\rvert
  \;\le\; \frac{1}{8}(\alpha_2 - \alpha_1)\abs{\eta_1 - \eta_2}^2\abs{t - t_\star}
  \qquad \forall t \in \R.
  \label{eq:logistic-error-bound}
\end{equation}
\end{corollary}
\noindent\ProofRef{cor:logistic-transition}

\begin{remark}[Sign of Exponent Drift and Three-Regime Taxonomy]
\label{rem:drift-direction}
Combining~\eqref{eq:K-third-derivative}--\eqref{eq:water-filling-K-drift} with \Cref{thm:master-mellin,thm:master-mixing,thm:master-envelope} separates three structural drivers of exponent drift $\eta'(t)$ across scales:
\begin{enumerate}
  \item[(i)] \emph{Multi-domain data heterogeneity ($\Var_{\PP_t}(\alpha) = 0$).} Both under a fixed multi-domain mixture ($\Var_{\QQ_t}(\beta) > 0$) and under KKT water-filling ($\Var_{\PP_{e^v}\opt}(\rho_I) > 0$ in \Cref{prop:water-filling-drift}), fast-decaying data modes saturate as $v\opt(t)$ grows, driving the effective data exponent and $\eta(t)$ strictly downward:
  \begin{equation}
    \frac{\dd \beta_{\mathrm{eff}}}{\dd v} = -\Var_{\QQ_t}(\beta) < 0, \quad
    \mc{K}(t) = -\bigl(1-\eta(t)\bigr)\frac{\Var_{\QQ_t}(\beta)}{\E_{\QQ_t}[\beta]} < 0
    \;\Longrightarrow\;
    \eta'(t) < 0,
  \end{equation}
  which shifts compute allocation toward data $D$ before relaxing to $\frac{b_0}{\alpha_\infty + b_0}$ as $\varepsilon(D) \to 0$.
  \item[(ii)] \emph{Positive subleading capacity modes ($\Var_{\PP_t}(\alpha) > 0$, $\Var_{\QQ_t}(\beta) = 0$).} Conversely, when positive subleading spectral--geometric modes decay faster than the asymptotic tail (\Cref{thm:master-mellin,prop:gf-holomorphic}), the effective capacity exponent $\alpha_{\mathrm{eff}}(t) = \E_{\PP_t}[\alpha]$ decreases with $u\opt(t)$ and drives $\eta(t)$ upward:
  \begin{equation}
    \frac{\dd \alpha_{\mathrm{eff}}}{\dd u} = -\Var_{\PP_t}(\alpha) < 0, \quad
    \mc{K}(t) = +\eta(t)\frac{\Var_{\PP_t}(\alpha)}{\E_{\PP_t}[\alpha]} > 0
    \;\Longrightarrow\;
    \eta'(t) > 0.
  \end{equation}
  \item[(iii)] \emph{Signed filter poles and uncalibrated hyperparameter curvature.} In~\eqref{eq:K-third-derivative}, adding a negative rational filter mode $A_f e^{-\alpha_f u}$ ($A_f = -\abs{A_f} < 0$, $\alpha_f \Let \alpha_\infty + \Delta_f$ in \Cref{thm:master-mellin}) to a positive-weight spectral potential $U_+(u)$ shifts the capacity log-curvature ratio according to
  \begin{equation}
    -\frac{U'''(u)}{U''(u)} - \left(-\frac{U_+'''(u)}{U_+''(u)}\right)
    \;=\; -\frac{\abs{A_f}\alpha_f^2 e^{-\alpha_f u}}{U''(u)}\left(\alpha_f - \left(-\frac{U_+'''(u)}{U_+''(u)}\right)\right),
  \end{equation}
  contributing with sign opposite to $\alpha_f + U_+'''(u)/U_+''(u)$. Complementarily, as shown next in \Cref{thm:master-envelope}, omitting learning-rate schedule calibration in pilot sweeps adds a positive semidefinite Schur-complement curvature penalty that biases empirical estimates $\widehat{\eta}$ upward whenever hyperparameter sensitivity along token horizon $v$ dominates that along capacity $u$, before calibrated fits relax toward $\eta = 0.50$ \citep{kaplan2020scaling,hoffmann2022chinchilla,porian2024resolving}.
\end{enumerate}
\end{remark}

\subsection{Hyperparameter Envelope and Allocation Regret}
\label{subsec:envelope-regret}

In pretraining sweeps, the joint excess risk $\bar{\mc{E}}(\bm{r}, \xi) \in C^3(\R^2 \times \R^p)$ depends on both the log-resource vector $\bm{r}$ and auxiliary optimization hyperparameters $\xi \in \R^p$ (such as peak learning rate and batch size), which require across-scale calibration to avoid distorting pilot scaling estimates \citep{yang2021tuning,pearce2024reconciling,porian2024resolving,li2025functional,ramani2026optimizer}. When $\xi \mapsto \bar{\mc{E}}(\bm{r}, \xi)$ admits a unique interior minimizer $\xi\opt(\bm{r})$ with positive definite hyperparameter Hessian, minimizing out $\xi$ defines the calibrated envelope risk surface $\mc{E}(\bm{r})$ and its partitioned Hessian blocks at $(\bm{r}, \xi\opt(\bm{r}))$:
\begin{gather}
  \bm{r} \;\Let\; (u, v)^{\top} \;\in\; \R^2, \qquad
  \xi\opt(\bm{r}) \;\Let\; \argmin_{\xi \in \R^p} \bar{\mc{E}}(\bm{r}, \xi), \nonumber\\
  \mc{E}(\bm{r}) \;\Let\; \bar{\mc{E}}\bigl(\bm{r}, \xi\opt(\bm{r})\bigr), \qquad
  \bm{H}_{\xi\xi}(\bm{r}) \;\Let\; \nabla_{\xi\xi}^2 \bar{\mc{E}}\bigl(\bm{r}, \xi\opt(\bm{r})\bigr) \;\succ\; 0, \nonumber\\
  \bm{H}_{rr}(\bm{r}) \;\Let\; \nabla_{rr}^2 \bar{\mc{E}}\bigl(\bm{r}, \xi\opt(\bm{r})\bigr) \in \R^{2 \times 2}, \qquad
  \bm{H}_{\xi r}(\bm{r}) \;\Let\; \nabla_{\xi r}^2 \bar{\mc{E}}\bigl(\bm{r}, \xi\opt(\bm{r})\bigr) \in \R^{p \times 2}.
  \label{eq:envelope-hessian-blocks}
\end{gather}
By implicit differentiation and block-matrix elimination \citep{boyd2004convex,zhang2005schur}, the Hessian of the calibrated envelope $\mc{E}(\bm{r})$ is reduced relative to the frozen-hyperparameter curvature $\bm{H}_{rr}(\bm{r})$ by a positive semidefinite Schur complement.

\begin{theorem}[Schur-Complement Hyperparameter Envelope]
\label{thm:master-envelope}
For any joint risk surface $\bar{\mc{E}} \in C^3(\R^2 \times \R^p)$ with partitioned Hessian blocks~\eqref{eq:envelope-hessian-blocks} satisfying $\bm{H}_{\xi\xi}(\bm{r}) \succ 0$, the optimal hyperparameter Jacobian $\bm{J}_{\xi}(\bm{r})$ and the gradient and Hessian of the envelope risk $\mc{E}(\bm{r})$ at $(\bm{r}, \xi\opt(\bm{r}))$ are given by
\begin{gather}
  \bm{J}_{\xi}(\bm{r}) \;\Let\; \frac{\partial \xi\opt(\bm{r})}{\partial \bm{r}^{\top}} \;=\; -\bm{H}_{\xi\xi}(\bm{r})^{-1}\bm{H}_{\xi r}(\bm{r}), \qquad
  \nabla_{\bm{r}} \mc{E}(\bm{r}) \;=\; \nabla_{\bm{r}} \bar{\mc{E}}(\bm{r}, \xi\opt(\bm{r})), \nonumber\\
  \nabla_{rr}^2 \mc{E}(\bm{r}) \;=\; \bm{H}_{rr}(\bm{r}) - \bm{\Delta}_S(\bm{r}), \qquad
  \bm{\Delta}_S(\bm{r}) \;\Let\; \bm{H}_{r\xi}(\bm{r})\bm{H}_{\xi\xi}(\bm{r})^{-1}\bm{H}_{\xi r}(\bm{r}) \;\succeq\; 0.
  \label{eq:schur-envelope}
\end{gather}
Consequently, if a pilot sweep freezes $\xi \equiv \xi_0$ at an anchor optimum $\xi_0 = \xi\opt(\bm{r}_0)$ for $\bm{r}_0 \in \R^2$, the hyperparameter drift $\Delta\xi(\bm{r}) \Let \xi_0 - \xi\opt(\bm{r})$ induces the frozen-hyperparameter excess risk, anchor curvature penalty, and apparent allocation-slope shift:
\begin{gather}
  \bar{\mc{E}}(\bm{r}, \xi_0) - \mc{E}(\bm{r})
  \;=\; \frac{1}{2}\Delta\xi(\bm{r})^{\top}\bm{H}_{\xi\xi}(\bm{r})\Delta\xi(\bm{r}) + R_{\mathrm{env}}(\bm{r}, \xi_0), \quad
  \abs{R_{\mathrm{env}}(\bm{r}, \xi_0)} \;\le\; \frac{1}{6}C_{\xi}\norm{\Delta\xi(\bm{r})}_2^3, \nonumber\\
  \nabla_{rr}^2\bigl[\bar{\mc{E}}(\bm{r}, \xi_0) - \mc{E}(\bm{r})\bigr]\Big|_{\bm{r} = \bm{r}_0}
  \;=\; \bm{J}_{\xi}(\bm{r}_0)^{\top}\bm{H}_{\xi\xi}(\bm{r}_0)\bm{J}_{\xi}(\bm{r}_0)
  \;=\; \bm{\Delta}_S(\bm{r}_0)
  \;\Let\; \begin{psmallmatrix}\Delta_{uu} & \Delta_{uv}\\\Delta_{uv} & \Delta_{vv}\end{psmallmatrix} \;\succeq\; 0, \nonumber\\
  \eta_{\mathrm{frozen}}(\bm{r}_0) - \eta(\bm{r}_0)
  \;=\; \frac{\bigl(1 - \eta(\bm{r}_0)\bigr)(\Delta_{vv} - \Delta_{uv}) - \eta(\bm{r}_0)(\Delta_{uu} - \Delta_{uv})}{\mc{H}_{\mathrm{iso}}(\bm{r}_0) + \Delta_{uu} - 2\Delta_{uv} + \Delta_{vv}}.
\end{gather}
\end{theorem}
\begin{proof}[Proof Sketch of \Cref{thm:master-envelope}]
Differentiating $\nabla_{\xi}\bar{\mc{E}}(\bm{r}, \xi\opt(\bm{r})) = 0$ in $\bm{r}$ and completing the square on the joint second-order differential around $(\bm{r}, \xi\opt(\bm{r}))$ gives
\begin{align*}
  \bm{H}_{\xi r}(\bm{r}) + \bm{H}_{\xi\xi}(\bm{r})\bm{J}_{\xi}(\bm{r}) &\;=\; \bm{0}
  \quad\Longleftrightarrow\quad
  \bm{J}_{\xi}(\bm{r}) \;=\; -\bm{H}_{\xi\xi}(\bm{r})^{-1}\bm{H}_{\xi r}(\bm{r}), \\
  \dd^2\bar{\mc{E}}(\bm{r}, \xi\opt(\bm{r}))
  &\;=\; \dd \bm{r}^{\top}\!\left(\bm{H}_{rr}(\bm{r}) - \bm{\Delta}_S(\bm{r})\right)\dd \bm{r}
  + \norm{\bm{H}_{\xi\xi}(\bm{r})^{1/2}\!\left(\dd\xi - \bm{J}_{\xi}(\bm{r})\dd \bm{r}\right)}_2^2,
\end{align*}
whose restriction to the envelope tangent space $\dd\xi = \bm{J}_{\xi}(\bm{r})\dd \bm{r}$ isolates~\eqref{eq:schur-envelope}. \ProofRef{thm:master-envelope}
\end{proof}

Restricting $\mc{E}(u, v)$ to a fixed iso-compute slice $\mc{M}_t$ around $(u\opt(t), v\opt(t))$ defines the one-dimensional slice risk $\phi_t(\delta u)$ in terms of the log-allocation shift $\delta u$ \citep{hoffmann2022chinchilla,sardana2024inference,besiroglu2024chinchilla}, along with its extreme directional curvatures on $[-\Delta, \Delta]$ ($\Delta > 0$):
\begin{gather}
  \delta u \;\Let\; u - u\opt(t) \;=\; -\bigl(v - v\opt(t)\bigr), \qquad
  \phi_t(\delta u) \;\Let\; \mc{E}\bigl(u\opt(t) + \delta u, v\opt(t) - \delta u\bigr), \nonumber\\
  \underline{\mc{H}}_{t,\Delta} \;\Let\; \min_{\abs{s} \le \Delta} \mc{H}_{\mathrm{iso}}\bigl(u\opt(t)+s, v\opt(t)-s\bigr) \;>\; 0, \qquad
  \overline{\mc{H}}_{t,\Delta} \;\Let\; \max_{\abs{s} \le \Delta} \mc{H}_{\mathrm{iso}}\bigl(u\opt(t)+s, v\opt(t)-s\bigr).
\end{gather}

\begin{corollary}[Quadratic Allocation Regret]
\label{cor:allocation-regret}
For any log-compute budget $t \in \R$ and any shift interval $[-\Delta, \Delta]$ ($\Delta > 0$), the slice excess risk $\phi_t(\delta u)$ satisfies
\begin{gather}
  \phi_t'(0) \;=\; 0, \qquad
  \phi_t''(\delta u) \;=\; \mc{H}_{\mathrm{iso}}\bigl(u\opt(t) + \delta u, v\opt(t) - \delta u\bigr) \;>\; 0, \nonumber\\
  \frac{1}{2}\,\underline{\mc{H}}_{t,\Delta}\,(\delta u)^2
  \;\le\; \phi_t(\delta u) - \phi_t(0)
  \;\le\; \frac{1}{2}\,\overline{\mc{H}}_{t,\Delta}\,(\delta u)^2
  \qquad \forall \delta u \in [-\Delta, \Delta],
  \label{eq:quadratic-regret-sandwich}
\end{gather}
and for the separable power-law risk $\mc{E}(u, v) = c_1 e^{-au} + c_2 e^{-bv}$ with optimal value $\mc{E}\opt(t) \Let \phi_t(0)$, the relative curvature at the optimum is scale-invariant:
\begin{equation}
  a c_1 e^{-au\opt(t)} \;=\; b c_2 e^{-bv\opt(t)} \;=\; \frac{ab}{a + b}\,\mc{E}\opt(t), \qquad
  \frac{\phi_t''(0)}{\mc{E}\opt(t)} \;=\; a b \;>\; 0.
\end{equation}
\end{corollary}
\noindent\ProofRef{cor:allocation-regret}

Finally, consider a composite surrogate slice objective $\widehat{\phi}_t : [-\Delta, \Delta] \to \R$ formed by combining capacity, mixing, and envelope-plus-coupling surrogates along $\mc{M}_t$, with total slice marginal-derivative error budget $\varepsilon_t$ (or function-value error budget $\varepsilon_{0,t}$):
\begin{equation}
  \widehat{\phi}_t(s) \;\Let\; \widehat{U}\bigl(u\opt(t)+s\bigr) + \widehat{V}\bigl(v\opt(t)-s\bigr) + \widehat{R}_{\mathrm{env},t}(s), \qquad
  \varepsilon_t \;\Let\; \varepsilon_{\mathrm{spec}}(t) + \varepsilon_{\mathrm{mix}}(t) + \varepsilon_{\mathrm{env}}(t).
  \label{eq:surrogate-slice-def}
\end{equation}
Combining~\eqref{eq:surrogate-slice-def} with the slice strong-convexity sandwich~\eqref{eq:quadratic-regret-sandwich} yields deterministic stability bounds on the resulting compute-allocation shift and decision regret.

\begin{proposition}[Surrogate-to-Decision Allocation Regret Stability]
\label{prop:surrogate-decision-regret}
Let $\widehat{\phi}_t : [-\Delta, \Delta] \to \R$ be a surrogate slice objective on $[-\Delta, \Delta]$ ($\Delta > 0$):
\begin{enumerate}
  \item[(i)] \emph{First-order derivative control.} For any derivative error budget $\varepsilon_t \ge 0$ and first-order optimization tolerance $\tau_t \ge 0$, any candidate shift $\delta\hat{u}_t \in [-\Delta, \Delta]$ obeys
  \begin{align}
    &\sup_{\abs{s} \le \Delta}\abs{\widehat{\phi}_t'(s) - \phi_t'(s)} \;\le\; \varepsilon_t
    \quad\text{and}\quad
    \abs{\widehat{\phi}_t'(\delta\hat{u}_t)} \;\le\; \tau_t \nonumber\\
    &\quad\Longrightarrow\quad
    \abs{\delta\hat{u}_t} \;\le\; \frac{\varepsilon_t + \tau_t}{\underline{\mc{H}}_{t,\Delta}}, \qquad
    \phi_t(\delta\hat{u}_t) - \phi_t(0)
    \;\le\; \frac{(\varepsilon_t + \tau_t)^2}{2\,\underline{\mc{H}}_{t,\Delta}}
    \;\le\; \frac{\overline{\mc{H}}_{t,\Delta}}{2\,\underline{\mc{H}}_{t,\Delta}^2}(\varepsilon_t + \tau_t)^2.
    \label{eq:end-to-end-regret-bound}
  \end{align}
  \item[(ii)] \emph{Zeroth-order function-value control.} For any function-value error budget $\varepsilon_{0,t} \ge 0$ and zeroth-order optimization tolerance $\tau_{0,t} \ge 0$, any candidate shift $\delta\hat{u}_t \in [-\Delta, \Delta]$ obeys
  \begin{align}
    &\sup_{\abs{s} \le \Delta}\abs{\widehat{\phi}_t(s) - \phi_t(s)} \;\le\; \frac{1}{2}\varepsilon_{0,t}
    \quad\text{and}\quad
    \widehat{\phi}_t(\delta\hat{u}_t) \;\le\; \min_{\abs{s} \le \Delta}\widehat{\phi}_t(s) + \tau_{0,t} \nonumber\\
    &\quad\Longrightarrow\quad
    \abs{\delta\hat{u}_t} \;\le\; \sqrt{\frac{2(\varepsilon_{0,t} + \tau_{0,t})}{\underline{\mc{H}}_{t,\Delta}}}, \qquad
    \phi_t(\delta\hat{u}_t) - \phi_t(0) \;\le\; \varepsilon_{0,t} + \tau_{0,t}.
    \label{eq:end-to-end-zeroth-order-bound}
  \end{align}
\end{enumerate}
\end{proposition}
\noindent\ProofRef{prop:surrogate-decision-regret}

\section{Numerical Verification and Empirical Evaluation}
\label{sec:experiments}

We evaluate the theoretical predictions of \Cref{sec:spectral,sec:mixing,sec:compute} in two stages. First, in \Cref{subsec:exp-pillar1,subsec:exp-pillar2,subsec:exp-pillar3}, we test our analytical expansions against high-precision numerical truth on controlled spectral resolvents, multi-domain simplex mixtures, and iso-compute risk surfaces. Second, in \Cref{subsec:exp-external}, we examine out-of-sample extrapolation, pilot conditioning, and allocation regret on three public foundation-model benchmarks: Pythia \citep{biderman2023pythia}, Chinchilla \citep{hoffmann2022chinchilla}, and RegMix \citep{liu2025regmix}.

\subsection{Finite-Capacity Approximation and Pole Confluence}
\label{subsec:exp-pillar1}

We first test the finite-capacity Mellin--Barnes expansion (\Cref{thm:master-mellin}) and its boundary reductions (\Cref{prop:pole-collision,prop:gf-holomorphic}) across 45 spectral configurations spanning model capacities $N \in [10^2, 10^8]$. These comprise a 36-cell factorial grid over filter exponent $q \in [0.8, 2.0]$, source decay $r \in [1.4, 2.2]$, and manifold dimension $d \in \{2, 4, 8\}$ ($\delta = 1/d$), together with 9 pole-collision ($\varrho \in \{0, 1\}$) and gradient-flow ($\psi_{\mathrm{GF}}(u) = e^{-2u^{-q}}$) slices, benchmarked against Euler--Maclaurin series evaluations accurate to $10^{-11}$.

\begin{figure}[t!]
  \centering
  \includegraphics[width=0.98\textwidth]{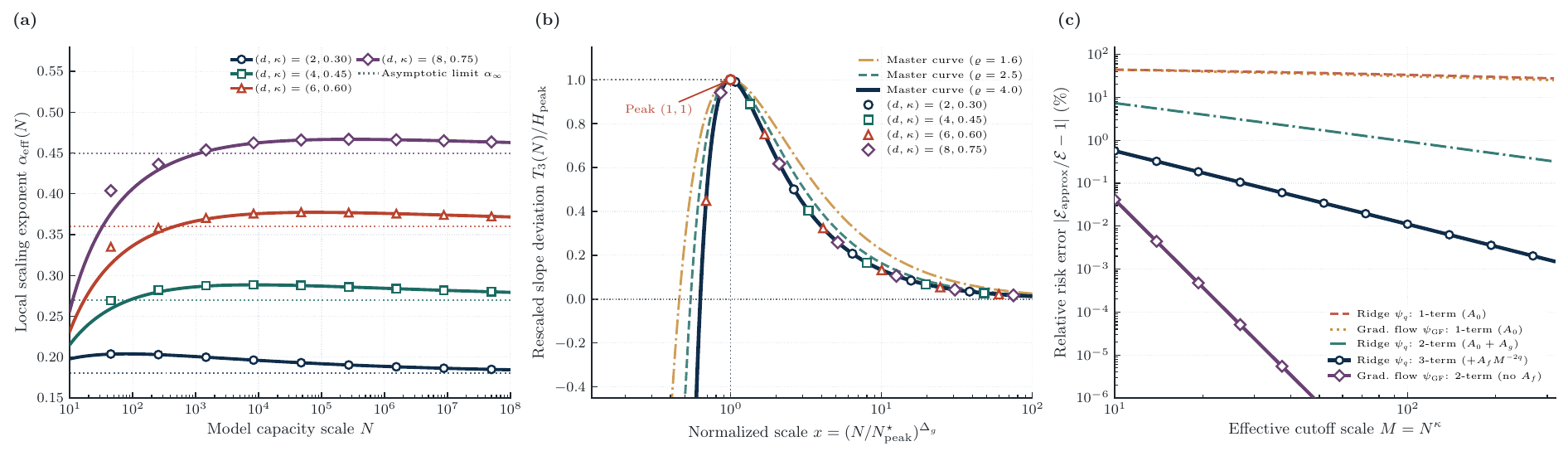}
  \caption{Pre-asymptotic spectral scaling trajectories and filter holomorphicity (\Cref{thm:master-mellin,cor:slope-collapse,prop:gf-holomorphic}). (a) Local exponent $\alpha_{\mathrm{eff}}(N)$ (curves: 3-term expansion; markers: numerical series) for four $(d, \kappa)$ systems at $\varrho = 4.0$, converging to $\alpha_\infty = \kappa(r-1)$. (b) Normalized slope deviation $T_3(N)/H_{\mathrm{peak}}$ versus $x = (N / N_{\mathrm{peak}}\opt)^{\Delta_g}$, collapsing onto the master curve $F_\varrho(x)$~\eqref{eq:master-slope-curve}. (c) Relative risk error ($\%$) versus $M = N^\kappa$ under rational filtering $\psi_q$ (1--3 terms) and gradient flow $\psi_{\mathrm{GF}}$ (1--2 terms).}
  \label{fig:pillar1-mellin}
\end{figure}

\begin{figure}[t!]
  \centering
  \includegraphics[width=0.98\textwidth]{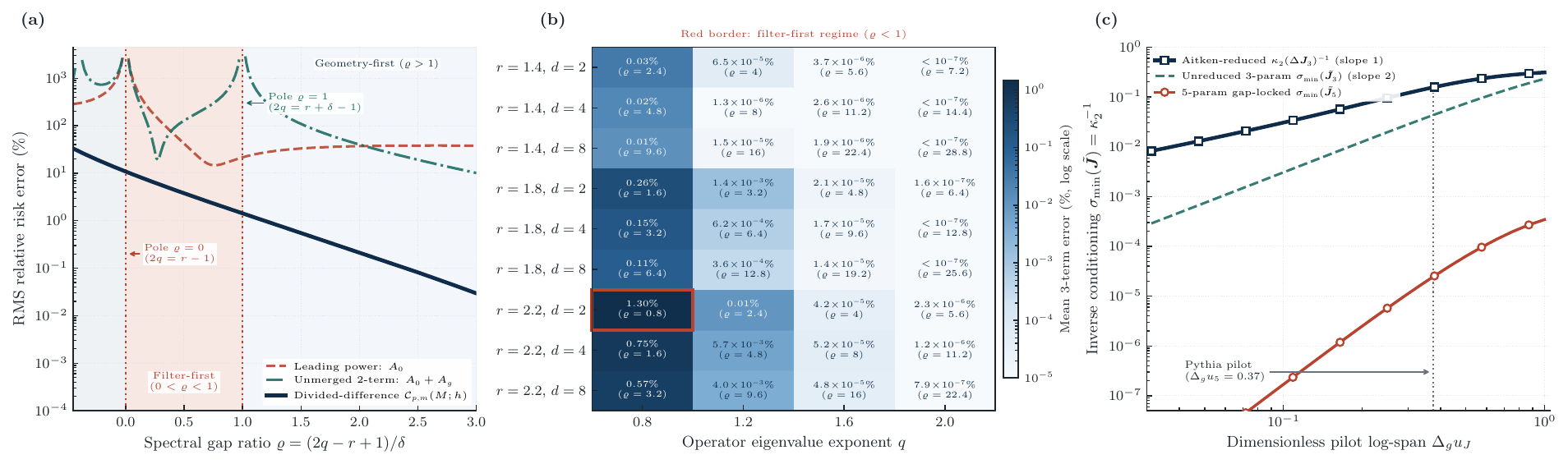}
  \caption{Pole-collision regularity, 36-cell factorial accuracy, and pilot conditioning (\Cref{thm:master-mellin,prop:pole-collision,prop:identifiability}). (a) RMS relative error ($\%$) across gap ratio $\varrho$, comparing the leading term $A_0$, the unmerged sum $A_0 + A_g$, and the divided-difference form $\mc{C}_{p,m}(M; h)$~\eqref{eq:divided-difference-uniform} across $\varrho \in \{0, 1\}$. (b) Mean 3-term relative error ($\%$) across the 36-cell $(r, d, q)$ grid (red border: $\varrho < 1$). (c) Smallest singular value $\sigma_{\min}(\tilde{\bm{J}}) = \kappa_2^{-1}$ versus pilot log-span $\Delta_g u_J$ for Aitken-reduced $\Delta\bm{J}_3$ (slope $1$), unreduced $\tilde{\bm{J}}_3$ (slope $2$), and 5-parameter $\tilde{\bm{J}}_5$.}
  \label{fig:pillar1-spectrum-grid}
\end{figure}

As shown in \Cref{fig:pillar1-mellin}(a,b), although the local scaling exponents $\alpha_{\mathrm{eff}}(N)$ peak at widely separated capacities across different manifold dimensions $d$ and cutoff exponents $\kappa$, normalizing the local slope deviation by its analytical peak scale $N_{\mathrm{peak}}\opt$ and peak height $H_{\mathrm{peak}}$ (\Cref{cor:slope-collapse}) collapses all four systems onto the single-parameter master curve $F_\varrho(x)$ to machine precision ($9.62 \times 10^{-16}$ maximum deviation). \Cref{fig:pillar1-mellin}(c) further illustrates why iterative early stopping and explicit Tikhonov regularization exhibit distinct finite-capacity corrections (\Cref{prop:gf-holomorphic}): because the gradient-flow Mellin kernel is holomorphic in the left half-plane, the two spectral--geometric zeta poles alone approximate the risk within $8.03 \times 10^{-5}\%$ already at $M = 31.62$ ($N = 10^3$), whereas under rational ridge filtering $\psi_q$, omitting the negative filter residue $A_f M^{-2q}$ leaves a $2.64\%$ error that drops by thirty-fold (to $0.0798\%$) once the filter pole is included.

When the filter pole approaches either spectral--geometric pole ($\varrho \to 0$ or $\varrho \to 1$), individual Mellin residues diverge with opposite signs and cause the unmerged power-law sum to blow up (\Cref{fig:pillar1-spectrum-grid}(a)). Combining the colliding poles into the divided-difference representation $\mc{C}_{p,m}(M; h)$ (\Cref{prop:pole-collision}) restores smooth convergence across both logarithmic resonances, reducing median relative error across the 36-cell factorial grid from $26.93\%$ (single power law) to $0.00005\%$ (\Cref{fig:pillar1-spectrum-grid}(b); $\MetricPillarIMedianError$ across all 45 tasks). Despite this analytical accuracy, \Cref{fig:pillar1-spectrum-grid}(c) shows why fitting multi-power sums directly on narrow pilot grids is ill-posed (\Cref{prop:identifiability}): as the dimensionless pilot log-span $\Delta_g u_J$ shrinks, both the column-normalized five-parameter Jacobian $\tilde{\bm{J}}_5$ and the unreduced three-parameter Jacobian $\tilde{\bm{J}}_3$ suffer at least quadratic singular-value decay, whereas eliminating $E_\infty$ via the first-order Aitken difference Jacobian $\Delta\bm{J}_3$ (\Cref{cor:aitken-identifiability}) restores linear $u_J$ conditioning (slope $1$).

\subsection{Multi-Domain Mixing and Sequential Allocation}
\label{subsec:exp-pillar2}

We next evaluate the KKT water-filling law (\Cref{prop:water-filling}) and finite-budget bottleneck bounds (\Cref{thm:master-mixing}) on multi-domain mixtures over $\Delta_{K-1}$, covering both distinct domain exponents ($\beta = (0.18, 0.42, 0.85)$ for $K = 3$ and $\beta = (0.18, 0.35, 0.55, 0.85)$ for $K = 4$) and tied slowest bottleneck exponents ($\beta = (0.22, 0.22, 0.65)$).

\begin{figure}[t!]
  \centering
  \includegraphics[width=0.98\textwidth]{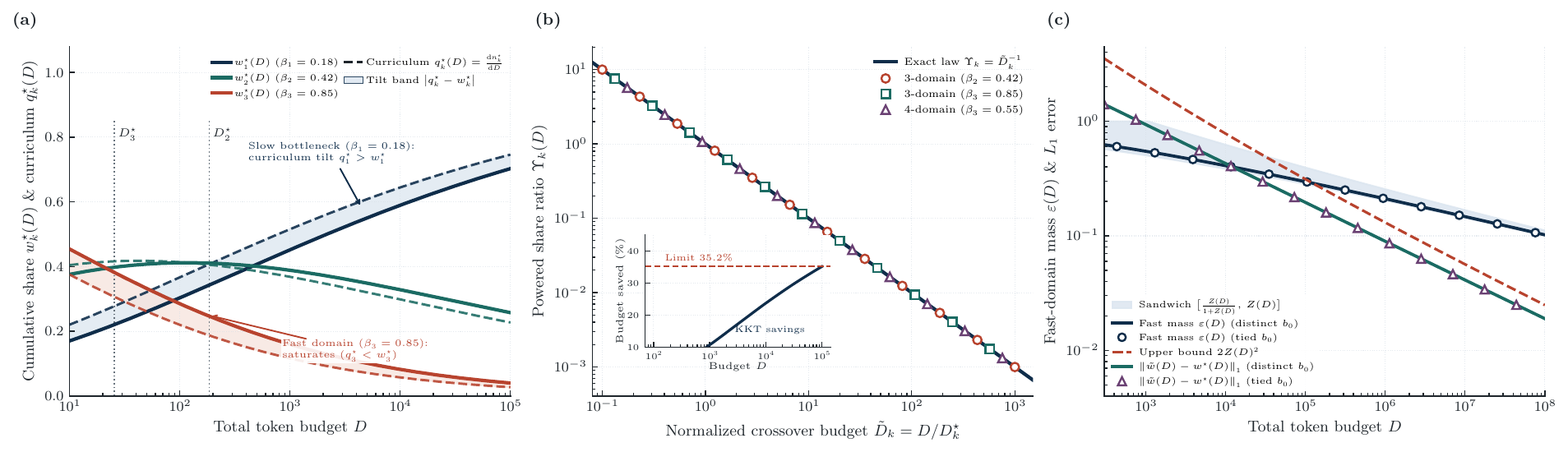}
  \caption{Simplex data mixing and finite-budget bottleneck bounds (\Cref{prop:water-filling,cor:powered-share-ratio,prop:curriculum-rn,thm:master-mixing}). (a) Cumulative shares $w_k\opt(D)$ (solid), sequential rates $q_k\opt(D)$ (dashed), and Radon--Nikodym tilt bands (shaded) across $D \in [10^1, 10^5]$ for $\beta = (0.18, 0.42, 0.85)$. (b) Powered share ratio $\Upsilon_k(D)$ versus $\tilde{D}_k = D/D_k\opt$, collapsing onto $\tilde{D}_k^{-1}$ (inset: $35.2\%$ token budget saved over uniform sampling). (c) Fast-domain mass $\varepsilon(D)$ within its sandwich band~\eqref{eq:eps-sandwich} and plug-in error $\norm{\tilde{w}(D) - w\opt(D)}_1 \le 2Z(D)^2$~\eqref{eq:plugin-L1-bound} for distinct and tied bottleneck exponents.}
  \label{fig:pillar2-mixing}
\end{figure}

As the total token budget $D$ grows, the water-filling solution shifts cumulative weight $w_k\opt(D)$ away from fast-saturating domains toward the slowest bottleneck domain (\Cref{fig:pillar2-mixing}(a)). Because sequential training must anticipate this shift at the margin, the instantaneous allocation policy $q_k\opt(D) = \frac{\dd[D w_k\opt(D)]}{\dd D}$ systematically leads the cumulative mixture via the Radon--Nikodym tilt factor $(\beta_k + 1)^{-1}$ (\Cref{prop:curriculum-rn,prop:simplex-hessian}), saving $\MetricPillarIIBudgetSaved$ of the total token budget compared to uniform sampling (\Cref{fig:pillar2-mixing}(b), inset). Normalizing the token budget by each domain's crossover scale $D_k\opt$ collapses the powered share ratio $\Upsilon_k(D)$ across all three- and four-domain systems onto the exact inverse law $\tilde{D}_k^{-1}$ (\Cref{cor:powered-share-ratio}, \Cref{fig:pillar2-mixing}(b)). In the large-budget regime (\Cref{fig:pillar2-mixing}(c)), whether the slowest exponent $b_0$ is unique or shared across multiple bottleneck domains, the residual fast-domain probability mass $\varepsilon(D)$ remains trapped inside the analytical sandwich band~\eqref{eq:eps-sandwich} and the closed-form plug-in mixture $\tilde{w}(D)$ converges to $w\opt(D)$ at the quadratic rate $2Z(D)^2$.

\subsection{Compute-Optimal Drift and Allocation Regret}
\label{subsec:exp-pillar3}

We next examine how multi-term capacity and data expansions govern the compute-optimal allocation trajectory along log-resource slices $\mc{M}_t$ (\Cref{thm:invariance-counterexample,thm:master-drift,thm:master-envelope}).

\begin{figure}[t!]
  \centering
  \includegraphics[width=0.98\textwidth]{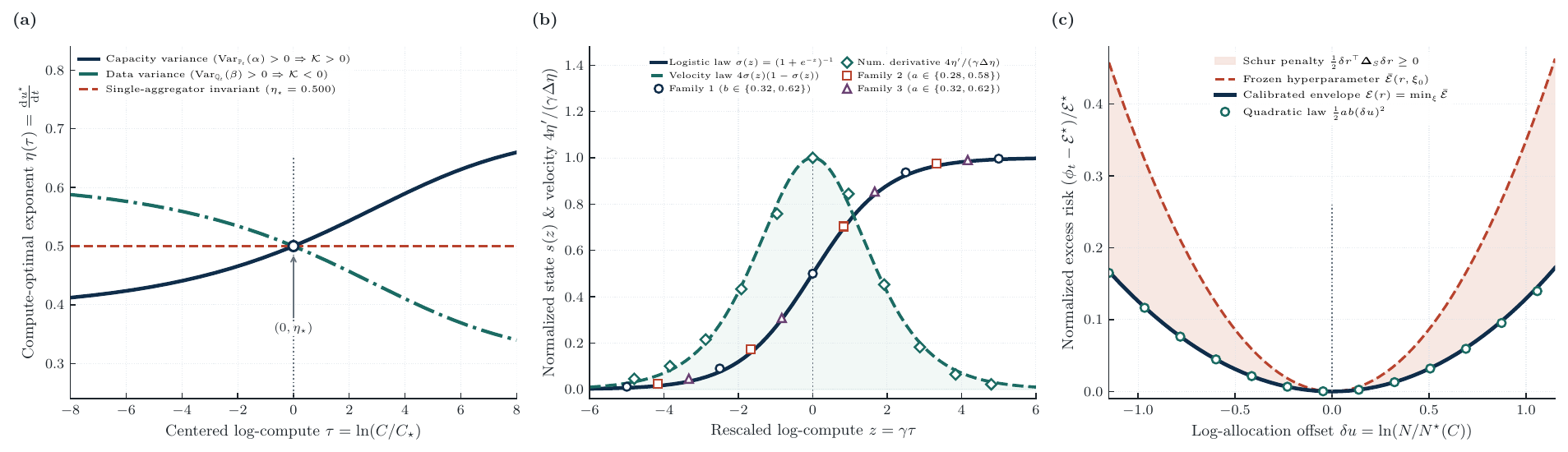}
  \caption{Iso-compute exponent drift, logistic transition, and Schur-complement regret (\Cref{thm:invariance-counterexample,thm:master-drift,cor:logistic-transition,thm:master-envelope,cor:allocation-regret}). (a) Optimal allocation exponent $\eta(\tau)$ across centered log-compute $\tau = \ln(C/C_\star)$ for capacity-dominated drift ($\mc{K}(t) > 0$), data-dominated drift ($\mc{K}(t) < 0$), and a monotone wrapper ($\eta_\star = 0.500$). (b) Normalized state $s(z)$ and velocity $4\eta'/(\gamma\Delta\eta)$ across three two-stratum families versus $z = \gamma\tau$, tracking $\sigma(z)$ and $4\sigma(z)(1-\sigma(z))$. (c) Relative slice regret $(\phi_t - \mc{E}\opt)/\mc{E}\opt$ versus $\delta u = \ln(N/N\opt(C))$ for the calibrated envelope $\mc{E}(\bm{r})$, quadratic law $\frac{1}{2}ab(\delta u)^2$, and frozen-schedule slice $\bar{\mc{E}}(\bm{r},\xi_0)$.}
  \label{fig:pillar3-coupling}
\end{figure}

\Cref{fig:pillar3-coupling}(a) demonstrates that cross-derivative coupling alone does not cause the compute-optimal exponent $\eta(t)$ to drift: wrapping a single power-law sum in a nonlinear monotone function preserves a flat allocation exponent $\eta(\tau) \equiv 0.500$ across all scales (\Cref{thm:invariance-counterexample}). In contrast, genuine multi-stratum heterogeneity induces S-shaped exponent drift whose direction is fixed by the Gibbs variance-to-mean rate $\mc{K}(t)$ (\Cref{thm:master-drift,rem:drift-direction}), shifting upward when capacity-side spectral variance dominates ($\mc{K}(t) > 0$) and downward when data-side domain variance dominates ($\mc{K}(t) < 0$). In the intrinsic transition coordinate $z_{\mathrm{int}}(t)$, the normalized trajectory $s(t)$ matches $\sigma(z_{\mathrm{int}})$ to machine precision ($\MetricPillarIIICollapseSpread$), while across the constant-rate rescaled log-compute coordinate $z = \gamma\tau \in [-5, 5]$ in \Cref{fig:pillar3-coupling}(b), all three two-stratum families stay numerically within $1.62 \times 10^{-2}$ of $\sigma(z)$ in state (inside the analytical bound~\eqref{eq:logistic-error-bound} of \Cref{cor:logistic-transition}) and within $4.83 \times 10^{-2}$ of $4\sigma(z)(1-\sigma(z))$ in normalized velocity ($1.44 \times 10^{-2}$ and $4.35 \times 10^{-2}$ on the plotted markers). Along a fixed compute budget (\Cref{fig:pillar3-coupling}(c)), the calibrated risk envelope closely follows the scale-invariant quadratic regret law $\frac{1}{2}ab(\delta u)^2$ (\Cref{cor:allocation-regret}), whereas freezing auxiliary hyperparameters $\xi \equiv \xi_0$ steepens the slice parabola by the positive semidefinite Schur-complement penalty $\bm{\Delta}_S(\bm{r}) \succeq 0$ (\Cref{thm:master-envelope}).

\subsection{Evaluation on Public Pretraining Benchmarks}
\label{subsec:exp-external}

We now examine how these pre-asymptotic scaling and identifiability principles inform out-of-sample extrapolation and candidate selection on three public foundation-model benchmarks: Pythia \citep{biderman2023pythia}, Chinchilla \citep{hoffmann2022chinchilla}, and RegMix \citep{liu2025regmix}.

\begin{figure}[t!]
  \centering
  \includegraphics[width=0.98\textwidth]{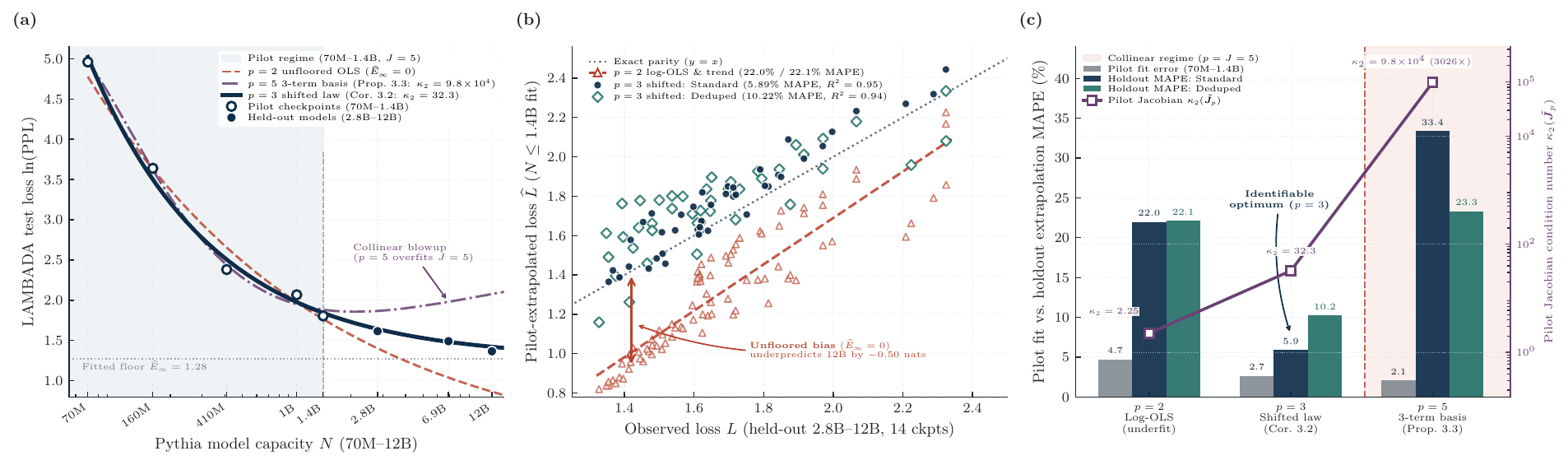}
  \caption{Out-of-sample extrapolation and pilot conditioning on the Pythia suite ($70\text{M}\text{--}12\text{B}$; \Cref{prop:identifiability,cor:aitken-identifiability}). (a) LAMBADA test loss $\ln(\mathrm{PPL})$ on Pythia-Standard fitted on the $J=5$ pilot sizes ($70\text{M}\text{--}1.4\text{B}$) and extrapolated to $2.8\text{B}\text{--}12\text{B}$ for orders $p \in \{2, 3, 5\}$. (b) Held-out parity across 42 post-warmup (of 48 total) large-model checkpoint evaluations per corpus, comparing $p=2$ log-OLS (triangles and dashed trend) against $p=3$ shifted power-law predictions ($5.89\%$ and $10.22\%$ MAPE). (c) Pilot and holdout MAPE across $p \in \{2, 3, 5\}$ versus column-normalized pilot condition number $\kappa_2(\tilde{\bm{J}}_p)$.}
  \label{fig:pythia-extrapolation}
\end{figure}

\paragraph{Out-of-Sample Extrapolation and Conditioning on Pythia.}
On EleutherAI's Pythia suite across the Standard and Deduped Pile corpora (8 model sizes from 70M to 12B across 16 logged checkpoints \citep{gao2020pile,lee2021deduplication,biderman2023pythia}), we fit scaling laws strictly on the 5 smallest pilot models ($70\text{M}\text{--}1.4\text{B}$, spanning $1.30$ decades) and extrapolate zero-shot to the 3 largest held-out models ($2.8\text{B}, 6.9\text{B}, 12\text{B}$; evaluated at the final checkpoint in Panel~(a,c) and across all 14 post-warmup checkpoints with $D \ge 27\text{B}$ tokens in Panel~(b)). Because an unfloored power law ($p = 2$) ignores the irreducible entropy floor, it systematically underpredicts large-model loss by $-0.50\text{ nats}$ at $12\text{B}$; in contrast, the three-parameter shifted power law ($p = 3$, fitted via multi-start log-loss nonlinear least squares on the $J = 5$ pilot grid as in \Cref{cor:aitken-identifiability,rem:aitken-snr}) tracks the curved trajectory across scale and cuts held-out LAMBADA \citep{paperno2016lambada} prediction error from $21.98\%$ to $\MetricPillarIPythiaStdMAPE$ MAPE on Standard and from $22.13\%$ to $\MetricPillarIPythiaDedupMAPE$ on Deduped (\Cref{fig:pythia-extrapolation}(a,b)).

A natural question is why fitting higher-order spectral expansions directly on the $J = 5$ pilot sizes fails, and how a practitioner can select model order $p = 3$ from pilot diagnostics alone (\Cref{prop:identifiability,cor:aitken-identifiability}). As shown in \Cref{fig:pythia-extrapolation}(c), whereas in-sample pilot fit error drops monotonically as parameters are added across $p \in \{2, 3, 5\}$ ($4.7\% \to 2.7\% \to 2.1\%$), out-of-sample holdout error traces a sharp U-curve governed by the unit-$\ell_2$ column-normalized pilot Jacobian condition number $\kappa_2(\tilde{\bm{J}}_p)$ on Standard. At $p = 3$, the pilot Jacobian is well-conditioned ($\kappa_2(\tilde{\bm{J}}_3) = \MetricPillarICondThree$, versus $\kappa_2(\tilde{\bm{J}}_2) = 2.25$) and achieves the unique extrapolation minimum on both corpora. Expanding around the $p = 3$ pilot exponent $\hat{\alpha}_0$ to the fixed-exponent five-column linearized basis $[1, N^{-\hat{\alpha}_0}, N^{-(\hat{\alpha}_0 + \Delta_g)}, N^{-(\hat{\alpha}_0 + \Delta_f)}, -(\ln N)N^{-\hat{\alpha}_0}]$ ($\Delta_g = 0.125, \Delta_f = 0.50$) over the narrow log-span $\Delta_g u_5 = 0.37$ induces a $3{,}026\times$ conditioning blowup ($\kappa_2(\tilde{\bm{J}}_5) = \MetricPillarICondFive$, unnormalized $1.67 \times 10^5$): its fifth singular value drops to $5.97 \times 10^{-6}\sigma_1$ (truncated to rank $4$ at threshold $\mathrm{rcond} = 10^{-4}$), and the remaining near-collinear directions amplify pilot noise into holdout divergence ($33.4\%$ MAPE on Standard, $23.3\%$ on Deduped). For the unconstrained three-power sum ($p = 7$), the parameter count exceeds the pilot grid size and, by \Cref{prop:identifiability}, remains algebraically rank-deficient ($\operatorname{rank}(\bm{J}_7) \le \MetricPillarIMaxRankSeven < 7$) around the single-power locus regardless of grid density.

\begin{figure}[t!]
  \centering
  \includegraphics[width=0.98\textwidth]{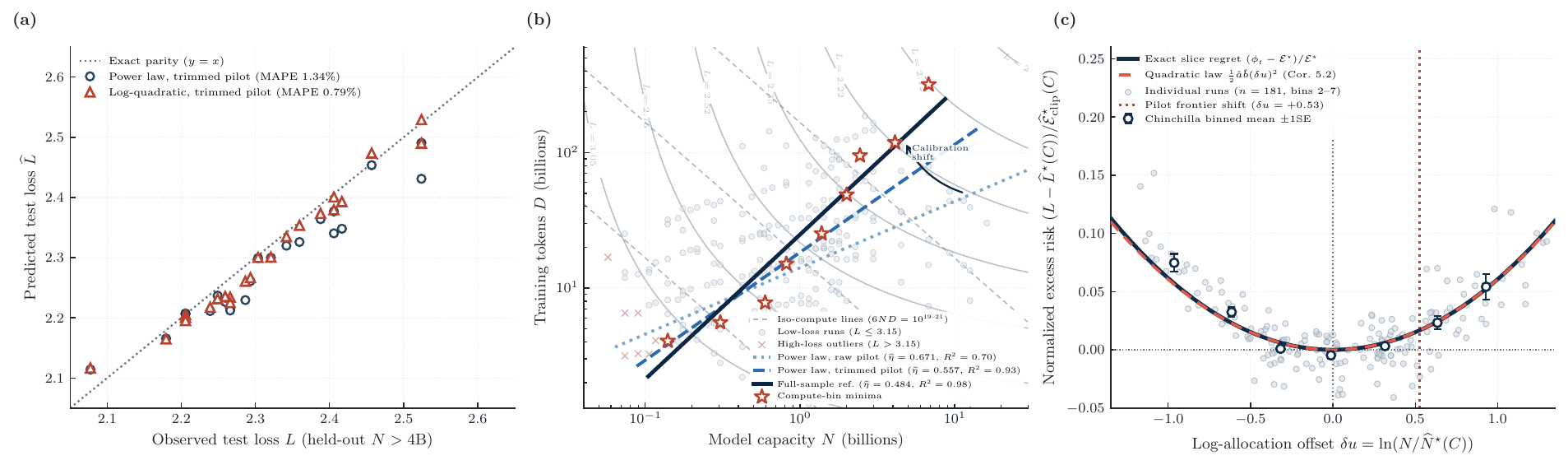}
  \caption{Iso-loss surface estimation, frontier calibration, and allocation regret on Chinchilla (245 runs; \Cref{thm:master-drift,thm:master-envelope,cor:allocation-regret}). (a) Held-out parity on large models ($N > 4\text{B}, L \le 3.15$) fitted on pilot runs ($N \le 4\text{B}$), comparing the trimmed power law ($1.34\%$ MAPE) and log-quadratic surface ($0.79\%$ MAPE). (b) Fitted $(N, D)$ iso-loss contours and allocation frontiers for the raw pilot ($\widehat{\eta} = 0.6712$), trimmed pilot ($\widehat{\eta} = 0.5569$), and full-sample reference ($\widehat{\eta} = 0.4839$). (c) Normalized slice regret versus $\delta u = \ln(N/\widehat{N}\opt(C))$ across 181 low-loss runs in bins $2\text{--}7$, compared against the quadratic law $\frac{1}{2}\hat{a}\hat{b}(\delta u)^2$.}
  \label{fig:chinchilla-regret}
\end{figure}

\begin{table}[t!]
  \centering
  \caption{Frontier estimation and out-of-sample extrapolation on the 245-run Chinchilla benchmark (221 pilot runs with $N \le 4\text{B}$; 24 held-out runs with $N > 4\text{B}$, 23 satisfying $L \le 3.15$; \Cref{thm:master-drift,thm:master-envelope,cor:allocation-regret,prop:surrogate-decision-regret}). Bold denotes the best pilot-only estimator. For $\widehat{\eta}$, $90\%$ CI reports the $5\%\text{--}95\%$ digitization-jitter interval across 40 replicates, and $R^2$ measures the 2D log-frontier fit across the 9 bin centers. High-$C$ Regret (nats) is the mean excess loss across the top four compute bins (bins $5\text{--}8$, 81 runs) under 1D frontier matching at bin-minimum compute $C_b$ (\emph{Frontier}; trimmed power-law regret remains $0.0393\text{ nats}$ under proximity and near-tie rules) and 2D predicted-loss minimization $\argmin_i \widehat{L}(N_i, D_i)$ (\emph{Surface}).}
  \label{tab:chinchilla-calibration}
  \scriptsize
  \setlength{\tabcolsep}{0pt}
  \renewcommand{\arraystretch}{1.28}
  \begin{tabular*}{\textwidth}{@{\extracolsep{\fill}}llccccccc@{}}
    \toprule
    & & \multicolumn{3}{c}{\textbf{Fitted $\widehat{\eta}$}} & \multicolumn{2}{c}{\textbf{Holdout MAPE}} & \multicolumn{2}{c@{}}{\textbf{High-$C$ Regret}} \\
    \cmidrule(lr){3-5} \cmidrule(lr){6-7} \cmidrule(l){8-9}
    \textbf{Estimator} & \textbf{Fit Sample} & Est. & $90\%$ CI & $R^2$ & $L\le 3.15$ & All 24 & Frontier & Surface \\
    \midrule
    Power law (raw) & $N \le 4\text{B}$ (221) & $\MetricPillarIIIChinchillaFullEta$ & $[0.6639, 0.6778]$ & $0.70$ & $1.92\%$ & $1.96\%$ & $0.0909$ & $0.0289$ \\
    Power law (trimmed) & $N \le 4\text{B}, L \le 3.15$ (205) & $\MetricPillarIIIChinchillaFiltEta$ & $[0.5468, 0.5655]$ & $0.93$ & $1.34\%$ & $2.34\%$ & $\mathbf{0.0393}$ & $0.0139$ \\
    \textbf{Log-quadratic} & \textbf{$N \le 4\text{B}, L \le 3.15$ (205)} & $\mathbf{0.5288}$ & $\mathbf{[0.5216, 0.5347]}$ & $\mathbf{0.96}$ & $\mathbf{0.79\%}$ & $\mathbf{1.86\%}$ & $0.0709$ & $\mathbf{0.0068}$ \\
    \midrule
    Full-sample reference & All $N, L \le 3.15$ (228) & $0.4839$ & $[0.4755, 0.4880]$ & $0.98$ & --- & --- & $0.0349$ & $0.0000$ \\
    \bottomrule
  \end{tabular*}
\end{table}

\paragraph{Iso-Compute Frontier Estimation on Chinchilla.}
On the 245 training runs of \citet{hoffmann2022chinchilla} digitized by \citet{besiroglu2024chinchilla}, we fit iso-loss surfaces on the 221 small-scale pilot runs ($N \le 4\text{B}$) and evaluate out-of-sample extrapolation on the 24 held-out large models ($N > 4\text{B}$) alongside retrospective candidate-pool regret across 9 equal-width log-compute bins (\Cref{fig:chinchilla-regret,tab:chinchilla-calibration}). Because the digitized summary records only $(N, D, L)$ triples without per-run learning-rate schedules, we apply the threshold $L \le 3.15$ as an empirical filter for high-loss short-horizon runs.

Fitting a separable power law to all 221 pilot runs yields $\widehat{\eta} = \MetricPillarIIIChinchillaFullEta$, consistent with the upward slope shift in \Cref{thm:master-envelope} and \Cref{rem:drift-direction}(iii) when high-loss runs are retained. Trimming $L > 3.15$ on the pilot sample (205 runs) lowers the estimated exponent to $\widehat{\eta} = \MetricPillarIIIChinchillaFiltEta$ (shifting the high-compute log-capacity frontier by $\delta u = +0.53$ in \Cref{fig:chinchilla-regret}(c)) and reduces held-out MAPE on the 23 low-loss large models from $1.92\%$ to $1.34\%$. To capture local surface curvature on the same 205-run pilot sample, we fit a second-order log-loss surface in $(u, v) = (\ln N, \ln D)$:
\begin{align}
  \ln\widehat{L}(u, v)
  &\;=\; c_0 + c_1 u + c_2 v + c_3 u^2 + c_4 v^2 + c_5 u v \nonumber\\
  \quad\Longrightarrow\quad
  \hat{u}\opt(t)
  &\;=\; \frac{c_2 - c_1}{2(c_3 + c_4 - c_5)} + \widehat{\eta}_{\mathrm{quad}}\,t,
  \qquad
  \widehat{\eta}_{\mathrm{quad}} \;=\; \frac{2c_4 - c_5}{2(c_3 + c_4 - c_5)},
  \label{eq:chinchilla-quad-frontier}
\end{align}
along each iso-compute line $u + v = t = \ln(C/6)$, where positive slice curvature $2(c_3 + c_4 - c_5) = 0.0439 > 0$ ensures a unique slice minimum and constant affine slope $\widehat{\eta}_{\mathrm{quad}} = 0.5288$, moving closer to the full-sample reference ($\widehat{\eta} = 0.4839$) and cutting held-out MAPE to $0.79\%$ on the 23 low-loss runs and $1.86\%$ across all 24 runs (\Cref{fig:chinchilla-regret}(a)). Across the top four compute bins (bins $5\text{--}8$), trimming high-loss pilot runs cuts 1D frontier-matching regret by $\MetricPillarIIIChinchillaHighRegretRed$ under the table-order protocol (from $0.0909$ to $0.0393\text{ nats}$, remaining at $0.0393\text{ nats}$ across alternative tie-breaking rules in \Cref{tab:chinchilla-calibration}), whereas minimizing predicted loss $\widehat{L}(N_i, D_i)$ directly over the 2D candidate coordinates (\emph{Surface}) leverages the full second-order surface to cut mean regret by $52.1\%$ for the trimmed power law ($0.0139\text{ nats}$) and by $76.4\%$ for the log-quadratic surface ($0.0068\text{ nats}$). In \Cref{fig:chinchilla-regret}(c), the 181 low-loss residuals in bins $2\text{--}7$ closely follow the analytical slice regret curve and quadratic parabola $\frac{1}{2}\hat{a}\hat{b}(\delta u)^2$ of \Cref{cor:allocation-regret}.

\begin{figure}[t!]
  \centering
  \includegraphics[width=0.98\textwidth]{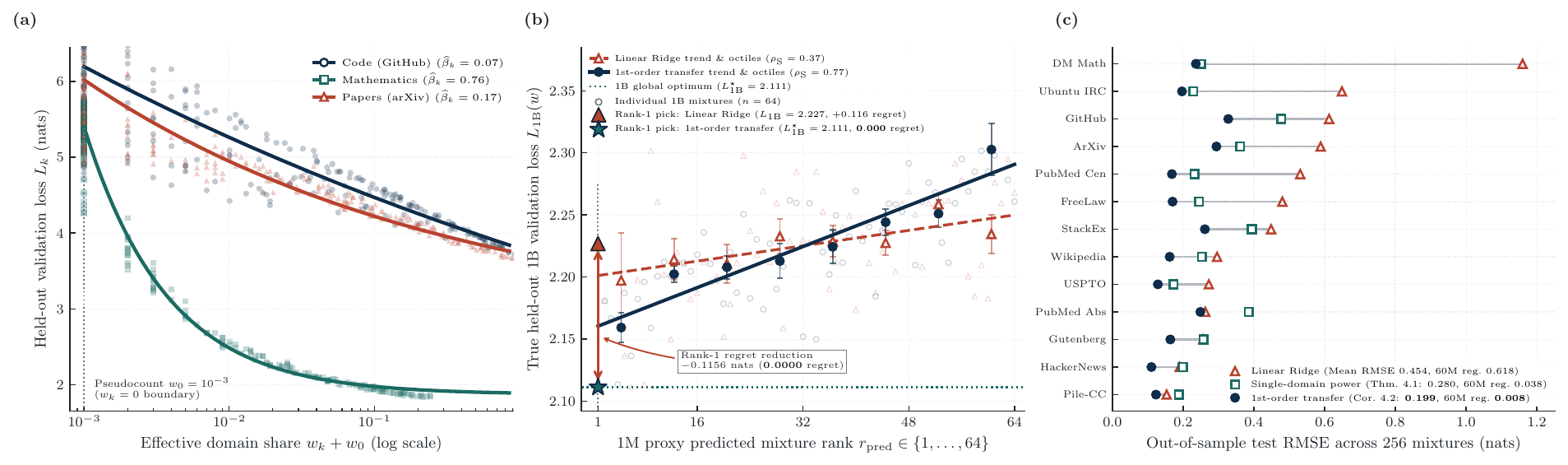}
  \caption{Multi-domain mixing and zero-shot $1\text{M}\to 1\text{B}$ transfer on the RegMix benchmark (\Cref{thm:master-mixing,rem:clipped-pseudocount,thm:fisher-transfer}). (a) Held-out loss $L_k$ versus shifted share $w_k + w_0$ ($w_0 = 10^{-3}$) across 256 test mixtures for Code, Mathematics, and arXiv, fitted on 512 pilot mixtures at 1M. (b) Held-out 1B loss $L_{1\mathrm{B}}(w)$ across 64 candidate mixtures versus 1M predicted rank $r_{\mathrm{pred}}$, contrasting linear Ridge ($\rho_{\mathrm{S}} = 0.37$) and 1st-order log-transfer ($\rho_{\mathrm{S}} = 0.77$). (c) Out-of-sample test RMSE across all 13 Pile validation domains on 256 held-out 1M mixtures.}
  \label{fig:regmix-transfer}
\end{figure}

\begin{table}[t!]
  \centering
  \caption{Domain prediction RMSE (on 256 held-out 1M mixtures across 13 Pile domains) and zero-shot $1\text{M} \to 60\text{M} \to 1\text{B}$ transfer on the 17-domain RegMix benchmark (fitted on 512 pilot mixtures at 1M; \Cref{thm:master-mixing,rem:clipped-pseudocount,prop:simplex-hessian,thm:fisher-transfer}).}
  \label{tab:regmix-transfer}
  \scriptsize
  \setlength{\tabcolsep}{0pt}
  \renewcommand{\arraystretch}{1.28}
  \begin{tabular*}{\textwidth}{@{\extracolsep{\fill}}lccccccccc@{}}
    \toprule
    & \multicolumn{3}{c}{\textbf{1M Test RMSE}} & \multicolumn{3}{c}{\textbf{Spearman $\rho_{\mathrm{S}}$ ($\uparrow$)}} & \multicolumn{3}{c@{}}{\textbf{Selection Regret (nats, $\downarrow$)}} \\
    \cmidrule(lr){2-4} \cmidrule(lr){5-7} \cmidrule(l){8-10}
    \textbf{Predictor Family (1M Fit)} & Mean & Agg. & $\Delta\%$ & $1\text{M}$ & $60\text{M}$ & $1\text{B}$ & $1\text{M}$ & $60\text{M}$ & $1\text{B}$ \\
    \midrule
    17-domain linear Ridge (baseline) & $\MetricPillarIIRegMixLinearRMSE$ & $0.2277$ & --- & $0.62$ & $0.56$ & $0.37$ & $0.7086$ & $0.6179$ & $0.1156$ \\
    Single-domain power (\Cref{thm:master-mixing}) & $\MetricPillarIIRegMixPowerRMSE$ & $0.1308$ & $-38.3\%$ & $0.92$ & $0.84$ & $0.74$ & $0.0733$ & $0.0377$ & $\mathbf{0.0000}$ \\
    Monotone PCHIP spline (non-param.) & $\MetricPillarIIRegMixSplineRMSE$ & $0.1329$ & $-39.0\%$ & $0.92$ & $0.84$ & $0.74$ & $\mathbf{0.0000}$ & $\mathbf{0.0080}$ & $\mathbf{0.0000}$ \\
    \textbf{$+$ 1st-order transfer (\Cref{thm:fisher-transfer})} & $\MetricPillarIIRegMixTransferRMSE$ & $0.0990$ & $-56.1\%$ & $0.95$ & $0.93$ & $\mathbf{0.77}$ & $\mathbf{0.0000}$ & $\mathbf{0.0080}$ & $\mathbf{0.0000}$ \\
    \textbf{$+$ Diag.\ log-quad.\ ($\ln w, \ln^2 w$)} & $\mathbf{\MetricPillarIIRegMixNonlinearRMSE}$ & $\mathbf{0.0908}$ & $\mathbf{-59.4\%}$ & $\mathbf{0.96}$ & $\mathbf{0.94}$ & $0.76$ & $0.1131$ & $0.0450$ & $\mathbf{0.0000}$ \\
    \bottomrule
  \end{tabular*}
\end{table}

\paragraph{Zero-Shot Cross-Scale Data Mixing on RegMix.}
Finally, on the RegMix benchmark \citep{liu2025regmix}, we fit domain-loss predictors across the 17 training domains on 512 pilot mixtures at 1M scale and evaluate out-of-sample mixture prediction across the 13 tracked Pile validation domains \citep{gao2020pile} on 256 held-out mixtures at 1M, together with zero-shot cross-scale ranking and selection transfer to 256 mixtures at 60M and 64 mixtures at 1B at fixed per-scale token budgets (\Cref{fig:regmix-transfer,tab:regmix-transfer}). Replacing linear Ridge with separable single-domain power laws $E_k + A_k(w_k + w_0)^{-\beta_k}$—where the pseudocount offset $w_0 = 10^{-3}$ (\Cref{rem:clipped-pseudocount}~\eqref{eq:pseudocount-water-filling}) regularizes zero-share boundary evaluations $w_k = 0$—cuts 1M domain-mean RMSE by $38.3\%$ (from $\MetricPillarIIRegMixLinearRMSE$ to $\MetricPillarIIRegMixPowerRMSE\text{ nats}$, matching the non-parametric monotone PCHIP spline at $\MetricPillarIIRegMixSplineRMSE\text{ nats}$), doubles zero-shot 1B Spearman rank correlation from $\rho_{\mathrm{S}} = 0.37$ to $0.74$, and eliminates the $+0.1156\text{ nats}$ 1B selection regret of linear Ridge.

Adding cross-domain coupling via ridge regression on log-shifted mixture coordinates $\ln(w + w_0)$ over the single-domain power-law residuals (motivated by \Cref{thm:fisher-transfer}) or fitting a diagonal log-quadratic expansion $[\ln(w + w_0), \ln^2(w + w_0)]$ further reduces 1M domain-mean RMSE by $56.1\%$ and $59.4\%$ (to $\MetricPillarIIRegMixTransferRMSE$ and $\MetricPillarIIRegMixNonlinearRMSE\text{ nats}$) and lifts rank correlation to $\rho_{\mathrm{S}} \ge 0.93$ at 1M--60M and $0.76\text{--}0.77$ at 1B (\Cref{tab:regmix-transfer}). While all four nonlinear predictors identify the top candidate in the 64-mixture 1B evaluation pool (\Cref{fig:regmix-transfer}(b)), the 1st-order residual log-transfer model achieves both the highest 1B rank correlation ($\rho_{\mathrm{S}} = 0.77$) and the lowest 60M selection regret ($0.0080\text{ nats}$, a $78.8\%$ reduction over single-domain power laws, whereas the diagonal log-quadratic fit partially overfits top-1 selection at 1M and 60M) while consistently reducing RMSE across all 13 validation domains (\Cref{fig:regmix-transfer}(c)).

\section{Discussion}
\label{sec:conclusion}

Classical scaling laws model validation loss as a separable power law with fixed exponents and static domain weights \citep{zhao2026foundations}, an asymptotic reduction recovered when subleading spectral residues and exponent variances vanish (\Cref{tab:classical-reductions}). The analysis developed in this paper shows that, at finite budgets, effective scaling exponents and frontier curvatures are governed by the internal heterogeneity of the target spectrum and the training corpus (\Cref{fig:proof-dag}). When a model approximates spectral modes that decay at different rates (\Cref{sec:spectral}) or trains on data sources that saturate at different speeds (\Cref{sec:mixing}), the active bottleneck shifts with model size $N$ and token budget $D$, producing both pre-asymptotic learning-curve curvature and scale-dependent drift in the compute-optimal allocation exponent $\eta\opt(C)$ (\Cref{sec:compute}).

Beyond accounting for the empirical benchmarks in \Cref{sec:experiments}, these results shed light on several practical questions in large language model pretraining. The first concerns how small-scale pilot runs should be designed when extrapolating validation loss across orders of magnitude \citep{kaplan2020scaling,hoffmann2022chinchilla,bi2024deepseek,meng2026sldbench}. Because finite-scale learning curves depart from a pure power law, it is tempting to fit unconstrained multi-exponent sums \citep{caballero2023broken} or to train a denser grid of intermediate checkpoints over a narrow compute range. \Cref{prop:identifiability,cor:aitken-identifiability,rem:aitken-snr} explain why this strategy is poorly conditioned: on a narrow pilot log-span $u_J$, multi-power Jacobians are nearly rank-deficient due to Vandermonde collinearity \citep{varah1985fitting,mlodozeniec2026ambiguous}, and packing additional checkpoints into the same interval shrinks the step size $\Delta u$ without widening the total span or improving worst-case identifiability, while amplifying observation noise under numerical differencing. Reliable extrapolation is instead supported by widening the scale ratio $N_J / N_1$ between the smallest and largest pilot models \citep{meng2026sldbench} and using first-order Aitken differencing across a geometric octave grid to eliminate the floor parameter $E_\infty$, restoring linear $u_J$ conditioning and providing a closed-form initializer for full-grid nonlinear least squares.

A second implication concerns the scheduling of heterogeneous pretraining mixtures across web text, code, and mathematical corpora \citep{xie2023doremi,ye2025datamixing,ge2024bimix,liu2025regmix}. In practice, large-scale training runs rarely keep mixture weights static; instead, they anneal the data distribution toward code and reasoning domains during mid-training \citep{bi2024deepseek}. The water-filling and sequential sampling laws in \Cref{thm:master-mixing,prop:curriculum-rn} formalize this practice at the cumulative token-quota level. Whenever high-volume domains carry larger sample exponents $\beta_k$, their marginal contribution diminishes rapidly once the total token budget $D$ exceeds the crossover scale $D_k\opt$ (\Cref{cor:powered-share-ratio}). Maintaining a balanced marginal return across domains therefore favors down-weighting fast-saturating sources by $(\beta_k + 1)^{-1}$ and reallocating token quotas toward slower-decaying bottlenecks as total tokens $D$ grow. At the same time, the simplex Hessian inverse and cross-domain perturbation bounds in \Cref{prop:simplex-hessian,thm:fisher-transfer} quantify how weak cross-domain coupling shifts the separable water-filling optimum and bound the resulting mixture regret at finite budgets, motivating the log-shifted cross-domain transfer predictors that reduce selection regret across scales (\Cref{tab:regmix-transfer}).

A third implication distinguishes genuine shifts in the compute-optimal frontier from optimization-induced estimation bias \citep{besiroglu2024chinchilla,pearce2024reconciling,porian2024resolving,li2025functional,ramani2026optimizer}. On the one hand, corpus deduplication, quality filtering \citep{lee2021deduplication,sorscher2022pruning}, and dynamic domain weighting alter the variance of data decay rates across the training mixture, which shifts the compute-optimal allocation exponent $\eta\opt(C)$ across scales along the trajectories of \Cref{thm:master-drift,prop:water-filling-drift,cor:logistic-transition}. On the other hand, \Cref{thm:master-envelope} shows that if auxiliary optimization hyperparameters $\xi$ (such as peak learning rate, batch size, or warmup horizon) are frozen away from their scale-optimal schedule $\xi\opt(\bm{r})$, the positive semidefinite Schur-complement penalty $\bm{\Delta}_S(\bm{r}_0) \succeq 0$ steepens the observed slice curvature around the calibration anchor and biases the apparent capacity exponent $\widehat{\eta}_N$ upward whenever the weighted data-side curvature penalty $(1 - \eta\opt)(\Delta_{vv} - \Delta_{uv})$ exceeds $\eta\opt(\Delta_{uu} - \Delta_{uv})$ (\Cref{rem:drift-direction}(iii), \Cref{tab:chinchilla-calibration}). Separating intrinsic frontier drift from hyperparameter distortion therefore requires calibrating learning-rate and batch-size schedules across pilot scales prior to estimating the $(N, D)$ frontier.

Finally, the slice curvature and error-propagation bounds in \Cref{cor:allocation-regret,prop:surrogate-decision-regret} clarify why over-training smaller models for inference efficiency remains effective in deployment-constrained regimes \citep{brown2020fewshot,hernandez2022repetition,muennighoff2023dataconstrained,sardana2024inference}. Because the iso-compute excess risk surface is flat at the training-optimal allocation and its relative curvature is governed by the product of the capacity and data exponents, training a moderately smaller model ($N < N\opt(C)$) on proportionally more tokens to reduce inference memory and latency incurs only a mild second-order pretraining loss penalty within the global curvature sandwich~\eqref{eq:quadratic-regret-sandwich}. Moreover, \Cref{prop:surrogate-decision-regret} shows that marginal-derivative errors of a fitted scaling surrogate propagate quadratically into downstream compute-allocation regret while uniform function-value errors propagate linearly.

\acks{We gratefully acknowledge computational support from Shanghai Smart Logic Technology and the Tianqiong supercomputing platform. The authors declare no competing financial interests or conflicts of interest related to this work.}

\newpage
\appendix
\section{Proofs for Section~\ref{sec:preliminaries}: Problem Setup and Preliminaries}
\label{app:preliminaries}

In this appendix, we provide a hyperlinked summary of core notation and defining equations across the manuscript (\Cref{tab:notation-summary}), followed by the complete proofs of the two lemmas stated in \Cref{sec:preliminaries}: the two-sided bound for the effective spectral dimension (\Cref{lem:resolvent-trace}) and the local Fisher quadratic approximation of multi-class cross-entropy risk (\Cref{lem:bregman-fisher}).

\begin{table}[t!]
  \centering
  \caption{Summary of core mathematical notation, defining formulas, and theoretical roles across \Cref{sec:preliminaries,sec:spectral,sec:mixing,sec:compute}.}
  \label{tab:notation-summary}
  \scriptsize
  \setlength{\tabcolsep}{0pt}
  \renewcommand{\arraystretch}{1.32}
  \begin{tabular*}{\textwidth}{@{\extracolsep{\fill}}clllc@{}}
    \toprule
    \textbf{Sec.} & \textbf{Symbol} & \textbf{Definition / Formula} & \textbf{Theoretical Role} & \textbf{Ref.} \\
    \midrule
    \S\ref{sec:preliminaries} & $\Delta_{K-1},\, \Delta_{K-1}^{\circ}$ & $\{w \in \R_+^K : \sum_{k=1}^K w_k = 1\}$ & Closed and open $(K-1)$-simplices & \S\ref{subsec:prob-hilbert} \\
    \S\ref{sec:preliminaries} & $\mc{T},\, \mc{R}(\mu),\, d_{\mathrm{eff}}(\mu)$ & $\mc{R}(\mu) = (\mc{T}+\mu\mc{I})^{-1}$, $\tr(\mc{T}\mc{R}(\mu))$ & Covariance, resolvent, effective dimension & \eqref{eq:resolvent-deff-def} \\
    \S\ref{sec:preliminaries} & $\Lambda_q,\, M_{\mu,\pm}$ & $\frac{\pi/q}{\sin(\pi/q)}$, $M_{\mu,\pm} = (c_\pm/\mu)^{1/q}$ & Euler constant and regularizer cutoffs & \eqref{eq:deff-sandwich} \\
    \S\ref{sec:preliminaries} & $\mc{I}_F(z),\, \mc{T}_F,\, C_3$ & $\diag(p) - pp^\top$, $\E[\bm{\Psi}^*\mc{I}_F\bm{\Psi}]$ & Softmax Fisher matrix and operator & \eqref{eq:bregman-sandwich} \\
    \midrule
    \S\ref{sec:spectral} & $M = N^\kappa,\, c_m$ & $c_m = c m^{-r}(1 + b m^{-\delta})$ & Cutoff scale and Weyl mode energy & \eqref{eq:weyl-mode-energy} \\
    \S\ref{sec:spectral} & $\psi_q(u),\, \Psi_q(s)$ & $(1+u^{-q})^{-2}$, $-\frac{s+q}{q^2}\frac{\pi}{\sin(\pi s/q)}$ & Spectral residual filter and Mellin kernel & \eqref{eq:psi-q-mellin} \\
    \S\ref{sec:spectral} & $\alpha_\infty,\, \Delta_g,\, \Delta_f,\, \varrho$ & $\kappa(r-1)$, $\kappa\delta$, $\kappa(2q-r+1)$, $\frac{\Delta_f}{\Delta_g}$ & Tail exponent, sub-leading gaps, gap ratio & \eqref{eq:gap-definitions} \\
    \S\ref{sec:spectral} & $A_0,\, A_g,\, A_{f,m}$ & $A_0 > 0$, $A_g > 0$, $A_f < 0$ & Asymptotic, Weyl, and filter Mellin residues & \eqref{eq:three-term-expansion} \\
    \S\ref{sec:spectral} & $T_3(N),\, N_{\mathrm{peak}}\opt,\, F_\varrho$ & $F_\varrho(x) = \frac{\varrho x^{-1}-x^{-\varrho}}{\varrho-1}$ & 3-term risk, peak scale, master slope curve & \eqref{eq:master-slope-curve} \\
    \S\ref{sec:spectral} & $\bm{J}_7,\, \bm{J}_5,\, \bm{J}_3$ & $\sigma_{\min}(\bm{J}_5) \le C_5(\Delta_g u_J)^2$ & Pilot Jacobians for $p \in \{7, 5, 3\}$ models & \eqref{eq:aitken-locking} \\
    \midrule
    \S\ref{sec:mixing} & $c_k,\, b_0,\, S$ & $c_k = \pi_k B_k$, $b_0 = \min_k \beta_k$ & Domain weight, bottleneck rate, slowest set & \eqref{eq:slowest-set-def} \\
    \S\ref{sec:mixing} & $D_k\opt,\, \tilde{D}_k,\, \Upsilon_k(D)$ & $\tilde{D}_k = D/D_k\opt$, $\Upsilon_k(D) = \tilde{D}_k^{-1}$ & Crossover budget and water-filling ratio & \eqref{eq:powered-ratio-law} \\
    \S\ref{sec:mixing} & $H,\, \theta_s,\, z_k(D),\, Z(D)$ & $Z(D) = \sum_{k \in S^c} z_k(D)$ & Bottleneck sum, tie shares, fast-domain slack & \eqref{eq:tie-aware-Z-def} \\
    \S\ref{sec:mixing} & $\varepsilon(D),\, \tilde{w}(D)$ & $\varepsilon(D) = \sum_{k \in S^c} w_k\opt(D)$ & Fast-domain mass and plug-in mixture & \eqref{eq:plugin-L1-bound} \\
    \S\ref{sec:mixing} & $n_k\opt,\, q\opt(D),\, \PP_D\opt,\, \QQ_D\opt$ & $n_k\opt = D w_k\opt(D)$, $q_k\opt = \frac{\dd n_k\opt}{\dd D}$ & Cumulative token quota and curriculum & \eqref{eq:radon-nikodym-tilt} \\
    \S\ref{sec:mixing} & $g_{w\opt},\, \bm{D}_H(D),\, \Delta w_{\mathrm{nat}}$ & $g_{w\opt}(\delta w, \delta\tilde{w}) = \delta w^{\top}\bm{D}_H(D)\delta\tilde{w}$ & Simplex Hessian metric and natural step & \eqref{eq:natural-grad-component} \\
    \midrule
    \S\ref{sec:compute} & $(u, v),\, \mc{M}_t,\, \mc{H}_{\mathrm{iso}}$ & $(u, v) = (\ln N, \ln D)$, $u+v=t$ & Log-resources, iso-compute slice, curvature & \eqref{eq:iso-compute-convexity} \\
    \S\ref{sec:compute} & $\eta(t),\, \mc{K}(t),\, \sigma(\gamma\tau)$ & $\eta(t) = \frac{\dd u\opt}{\dd t}$, $\eta' = \eta(1-\eta)\mc{K}(t)$ & Compute allocation exponent and drift rate & \eqref{eq:logistic-drift-law} \\
    \S\ref{sec:compute} & $\PP_t,\, \QQ_t,\, \alpha_{\mathrm{eff}},\, \beta_{\mathrm{eff}}$ & $\mc{K}(t) = \frac{\eta\Var_{\PP_t}(\alpha)}{\alpha_{\mathrm{eff}}} - \frac{(1-\eta)\Var_{\QQ_t}(\beta)}{\beta_{\mathrm{eff}}}$ & Marginal-gradient Gibbs measures & \eqref{eq:gibbs-variance-identity} \\[2.5pt]
    \S\ref{sec:compute} & $\bm{H}_{\xi\xi},\, \bm{\Delta}_S(\bm{r}),\, \phi_t(\delta u)$ & $\bm{\Delta}_S = \bm{H}_{r\xi}\bm{H}_{\xi\xi}^{-1}\bm{H}_{\xi r} \succeq 0$ & Hyperparameter Hessian and Schur penalty & \eqref{eq:quadratic-regret-sandwich} \\
    \bottomrule
  \end{tabular*}
\end{table}

\subsection{Proof of Lemma~\ref{lem:resolvent-trace} (Effective Spectral Dimension Bounds)}
\label{app:proof:lem:resolvent-trace}

\begin{proof}[Proof of \Cref{lem:resolvent-trace} \StatementRef{lem:resolvent-trace}]
The proof of the sandwich inequality~\eqref{eq:deff-sandwich} proceeds in three steps: termwise monotonicity comparison, integral comparison on $[0, \infty)$, and exact reduction to Euler's Beta--Gamma reflection identity.

\paragraph{Step 1: Termwise comparison via spectral monotonicity.}
For any fixed regularization parameter $\mu > 0$, consider the scalar resolvent filter $\phi_\mu(\lambda)$ on $(0, \infty)$:
\begin{equation}
  \phi_\mu(\lambda) \;\Let\; \frac{\lambda}{\lambda + \mu} \;=\; 1 - \frac{\mu}{\lambda + \mu},
  \qquad
  \phi_\mu'(\lambda) \;=\; \frac{\mu}{(\lambda + \mu)^2} \;>\; 0,
  \qquad
  \frac{c_\pm m^{-q}}{\mu} \;=\; \left(\frac{m}{M_{\mu,\pm}}\right)^{-q},
\end{equation}
where $M_{\mu,\pm} \Let (c_\pm / \mu)^{1/q}$. Because $\phi_\mu$ is strictly increasing on $(0, \infty)$, applying $\phi_\mu$ to the bilateral eigenvalue hypothesis $c_- m^{-q} \le \lambda_m \le c_+ m^{-q}$ ($m \in \N_{\ge 1}$) yields for every $m \in \N_{\ge 1}$:
\begin{align}
  g_-(m) \;\Let\; \frac{1}{1 + (m / M_{\mu,-})^q}
  &\;=\; \frac{c_- m^{-q}}{c_- m^{-q} + \mu}
  \;\le\; \frac{\lambda_m}{\lambda_m + \mu} \nonumber\\
  &\;\le\; \frac{c_+ m^{-q}}{c_+ m^{-q} + \mu}
  \;=\; \frac{1}{1 + (m / M_{\mu,+})^q} \;\Let\; g_+(m).
  \label{eq:app-gpm-termwise}
\end{align}

\paragraph{Step 2: Non-asymptotic integral comparison on $[0, \infty)$.}
For each sign $\pm$, the continuous profile $g_\pm(u) = \bigl(1 + (u / M_{\mu,\pm})^q\bigr)^{-1}$ is positive and strictly decreasing on $[0, \infty)$ with boundary value $g_\pm(0) = 1$. Consequently, on every unit interval $[m - 1, m]$ ($m \in \N_{\ge 1}$), monotonicity gives the symmetric rectangle bounds
\begin{equation}
  g_\pm(m) \;\le\; \int_{m-1}^m g_\pm(u)\dd u \;\le\; g_\pm(m - 1)
  \qquad \forall m \in \N_{\ge 1}.
\end{equation}
Summing these unit-interval inequalities over $m \in \N_{\ge 1}$ and using $g_-(0) = 1$ yields the aligned integral sandwich on $[0, \infty)$:
\begin{align}
  \int_0^{\infty} g_-(u)\dd u - 1
  &\;=\; \sum_{m=1}^{\infty} \int_{m-1}^m g_-(u)\dd u - g_-(0)
  \;\le\; \sum_{m=1}^{\infty} g_-(m - 1) - g_-(0)
  \;=\; \sum_{m=1}^{\infty} g_-(m) \nonumber\\
  &\;\le\; d_{\mathrm{eff}}(\mu)
  \;=\; \sum_{m=1}^{\infty} \frac{\lambda_m}{\lambda_m + \mu}
  \;\le\; \sum_{m=1}^{\infty} g_+(m)
  \quad\text{(by~\eqref{eq:app-gpm-termwise})} \nonumber\\
  &\;\le\; \sum_{m=1}^{\infty} \int_{m-1}^m g_+(u)\dd u
  \;=\; \int_0^{\infty} g_+(u)\dd u.
  \label{eq:app-integral-chain}
\end{align}

\paragraph{Step 3: Reduction to Euler's Beta--Gamma reflection integral.}
To evaluate the improper integral of $g_\pm(u)$ on $(0, \infty)$ for $q > 1$, we apply the power substitution $t = (u / M_{\mu,\pm})^q$ followed by the rational transformation $s = t / (1 + t)$ mapping $(0, \infty)$ onto $(0, 1)$:
\begin{equation}
  u \;=\; M_{\mu,\pm} t^{\frac{1}{q}}, \qquad
  \dd u \;=\; \frac{M_{\mu,\pm}}{q} t^{\frac{1}{q} - 1}\dd t, \qquad
  t \;=\; \frac{s}{1 - s}, \qquad
  \dd t \;=\; (1 - s)^{-2}\dd s.
\end{equation}
Substituting these differentials sequentially reduces the integral to Euler's Beta and Gamma functions:
\begin{align}
  \int_0^{\infty} \frac{\dd u}{1 + (u / M_{\mu,\pm})^q}
  &\;=\; \frac{M_{\mu,\pm}}{q}\int_0^{\infty} \frac{t^{\frac{1}{q} - 1}}{1 + t}\dd t
  \quad\text{(via $t = (u / M_{\mu,\pm})^q$)} \nonumber\\
  &\;=\; \frac{M_{\mu,\pm}}{q}\int_0^1 s^{\frac{1}{q} - 1}(1 - s)^{\left(1 - \frac{1}{q}\right) - 1}\dd s
  \quad\text{(via $s = \tfrac{t}{1 + t}$)} \nonumber\\
  &\;=\; \frac{M_{\mu,\pm}}{q}\,\mathrm{B}\!\left(\frac{1}{q}, 1 - \frac{1}{q}\right)
  \;=\; \frac{M_{\mu,\pm}}{q}\Gamma\!\left(\frac{1}{q}\right)\Gamma\!\left(1 - \frac{1}{q}\right) \nonumber\\
  &\;=\; M_{\mu,\pm}\frac{\pi / q}{\sin(\pi / q)}
  \;=\; \Lambda_q M_{\mu,\pm},
  \quad\text{(by Euler's reflection formula)}
  \label{eq:app-beta-gamma-eval}
\end{align}
Substituting~\eqref{eq:app-beta-gamma-eval} into the integral sandwich chain~\eqref{eq:app-integral-chain} establishes~\eqref{eq:deff-sandwich}.
\end{proof}

\subsection{Proof of Lemma~\ref{lem:bregman-fisher} (Local Fisher Quadratic Approximation)}
\label{app:proof:lem:bregman-fisher}

\begin{proof}[Proof of \Cref{lem:bregman-fisher} \StatementRef{lem:bregman-fisher}]
We prove~\eqref{eq:bregman-sandwich} in three steps: Legendre--Fenchel Bregman duality, exact directional cumulant differentiation and trace-class verification for the Fisher-pulled-back covariance operator $\mc{T}_F$, and a sharp uniform bound on the third cumulant tensor via Popoviciu's inequality.

\paragraph{Step 1: Bregman divergence representation and integral Taylor expansion.}
For any logit vector $z \in \R^V$, the softmax probability vector $p(z) = \nabla A(z) \in \Delta_{V-1}^{\circ}$ and the Kullback--Leibler divergence between any target logit $z\opt \in \R^V$ and predicted logit $z_{\theta} \in \R^V$ satisfy
\begin{align}
  p_v(z) &\;=\; \frac{\partial A(z)}{\partial z_v} \;=\; \frac{e^{z_v}}{\sum_{j=1}^V e^{z_j}} \qquad \forall v \in \{1, \dots, V\}, \nonumber\\
  \KL{p(z\opt)}{p(z_{\theta})}
  &\;=\; \sum_{v=1}^V p_v(z\opt)\left[\left(z_v\opt - A(z\opt)\right) - \left(z_{\theta,v} - A(z_{\theta})\right)\right] \nonumber\\
  &\;=\; A(z_{\theta}) - A(z\opt) - \ip{\nabla A(z\opt)}{z_{\theta} - z\opt}_{\R^V}
  \;=\; D_A(z_{\theta}, z\opt).
  \label{eq:app-kl-bregman}
\end{align}
Fix $x \in \mc{X}$ and parameterize the line segment $z_t(x)$ between $z\opt(x)$ and $z_{\theta}(x)$ for $t \in [0, 1]$. Because $A \in C^{\infty}(\R^V)$, applying Taylor's theorem with integral remainder to $g(t) \Let A(z_t(x))$ at $t = 1$ around $t = 0$ yields
\begin{align}
  \Delta z(x) &\;\Let\; z_{\theta}(x) - z\opt(x) \;=\; \bm{\Psi}(x)(\theta - \theta\opt) \in \R^V, \qquad
  z_t(x) \;\Let\; z\opt(x) + t\,\Delta z(x), \nonumber\\
  D_A\!\left(z_{\theta}(x), z\opt(x)\right)
  &\;=\; g(1) - g(0) - g'(0)
  \;=\; \frac{1}{2}g''(0) + \frac{1}{2}\int_0^1 (1 - t)^2 g'''(t)\dd t \nonumber\\
  &\;=\; \frac{1}{2}\ip{\Delta z(x)}{\nabla^2 A(z\opt(x))\,\Delta z(x)}_{\R^V} \nonumber\\
  &\qquad + \frac{1}{2}\int_0^1 (1 - t)^2 \nabla^3 A(z_t(x))\!\left[\Delta z(x), \Delta z(x), \Delta z(x)\right]\dd t.
  \label{eq:app-bregman-taylor}
\end{align}

\paragraph{Step 2: Explicit directional cumulant derivatives of $A(z)$ and trace-class verification.}
Fix any $z \in \R^V$ and any direction vector $u = (u_1, \dots, u_V)^{\top} \in \R^V$. Let $U$ denote the discrete random variable taking value $u_v$ with probability $p_v(z)$, and define its mean $\bar{u}(z)$ and the directional derivative of $p_v(z)$ along $u$:
\begin{align}
  \bar{u}(z) &\;\Let\; \E_{p(z)}[U] \;=\; \sum_{v=1}^V p_v(z) u_v, \nonumber\\
  \frac{\partial p_v(z)}{\partial z_j} &\;=\; p_v(z)(\delta_{vj} - p_j(z)), \qquad
  D_u p_v(z) \;=\; p_v(z)\bigl(u_v - \bar{u}(z)\bigr).
\end{align}
Using this derivative identity, we compute the first three directional cumulant derivatives of the log-partition function $A(z)$ along $u$ \citep{wainwright2008graphical,bach2010selfconcordant} in turn:
\begin{align}
  D_u A(z) &\;=\; \ip{\nabla A(z)}{u}_{\R^V}
  \;=\; \sum_{v=1}^V p_v(z) u_v
  \;=\; \bar{u}(z), \nonumber\\
  D_u^2 A(z) &\;=\; \nabla^2 A(z)[u, u]
  \;=\; \sum_{v=1}^V \left(D_u p_v(z)\right) u_v
  \;=\; \sum_{v=1}^V p_v(z)\bigl(u_v - \bar{u}(z)\bigr)u_v \nonumber\\
  &\;=\; \sum_{v=1}^V p_v(z)\bigl(u_v - \bar{u}(z)\bigr)^2
  \;=\; \Var_{p(z)}(U)
  \;=\; \ip{u}{\mc{I}_F(z) u}_{\R^V}, \nonumber\\
  D_u^3 A(z) &\;=\; \nabla^3 A(z)[u, u, u] \nonumber\\
  &\;=\; \sum_{v=1}^V \left(D_u p_v(z)\right)\bigl(u_v - \bar{u}(z)\bigr)^2
  - 2\bigl(D_u \bar{u}(z)\bigr)\sum_{v=1}^V p_v(z)\bigl(u_v - \bar{u}(z)\bigr) \nonumber\\
  &\;=\; \sum_{v=1}^V p_v(z)\bigl(u_v - \bar{u}(z)\bigr)^3
  \;=\; \E_{p(z)}\!\left[\bigl(U - \bar{u}(z)\bigr)^3\right].
  \quad\text{(centered third cumulant)}
  \label{eq:app-cumulant-derivatives}
\end{align}
Next, applying the Cauchy--Schwarz inequality to $\sum_{v=1}^V p_v(z) = 1$ bounds the trace of $\mc{I}_F(z)$ and the operator trace of $\mc{T}_F$ over any orthonormal basis $\{e_m\}_{m=1}^{\infty}$ of $\mc{H}$:
\begin{align}
  \tr\bigl(\mc{I}_F(z)\bigr)
  &\;=\; 1 - \sum_{v=1}^V p_v(z)^2
  \;\le\; 1 - \frac{1}{V}\left(\sum_{v=1}^V p_v(z)\right)^2
  \;=\; 1 - \frac{1}{V} \;<\; 1, \nonumber\\
  \tr(\mc{T}_F)
  &\;=\; \E_{\mu_\pi}\!\left[\sum_{m=1}^{\infty} \ip{\bm{\Psi}(X)e_m}{\mc{I}_F(z\opt(X))\bm{\Psi}(X)e_m}_{\R^V}\right] \nonumber\\
  &\;=\; \E_{\mu_\pi}\!\left[\tr\!\left(\bm{\Psi}(X)\bm{\Psi}(X)^*\mc{I}_F(z\opt(X))\right)\right]
  \;\le\; \left(1 - \frac{1}{V}\right)\kappa_{\Psi}^2 \;<\; \infty,
  \label{eq:app-TF-trace-bound}
\end{align}
confirming that $\mc{T}_F$ is a positive, self-adjoint, trace-class operator on $\mc{H}$.

\paragraph{Step 3: Sharp uniform operator-norm bound $C_3 \le \frac{1}{\sqrt{2}} < 1$.}
Let $u \in \R^V$ be any unit vector ($\norm{u}_2 = 1$), and set $u_{\max} \Let \max_{1 \le v \le V} u_v$ and $u_{\min} \Let \min_{1 \le v \le V} u_v$. If $u_{\max} = u_{\min}$, then $U$ is constant and $D_u^3 A(z) = 0$. Otherwise, let $v_+ \neq v_-$ be indices attaining $u_{\max}$ and $u_{\min}$; then Cauchy--Schwarz on those two coordinates gives
\begin{equation}
  (u_{\max} - u_{\min})^2
  \;\le\; 2\left(u_{v_+}^2 + u_{v_-}^2\right)
  \;\le\; 2\norm{u}_2^2 \;=\; 2
  \quad\Longrightarrow\quad
  u_{\max} - u_{\min} \;\le\; \sqrt{2}.
  \label{eq:app-spread-bound}
\end{equation}
Because $U \in [u_{\min}, u_{\max}]$ almost surely under $p(z)$ and $\bar{u}(z) \in [u_{\min}, u_{\max}]$, combining the pointwise range bound with Popoviciu's variance inequality on $[u_{\min}, u_{\max}]$ \citep{boucheron2013concentration} bounds the third central moment uniformly over all $z \in \R^V$ and $\norm{u}_2 = 1$:
\begin{align}
  \abs{U - \bar{u}(z)} &\;\le\; u_{\max} - u_{\min} \;\le\; \sqrt{2}, \qquad
  \Var_{p(z)}(U) \;\le\; \frac{1}{4}(u_{\max} - u_{\min})^2 \;\le\; \frac{1}{2}, \nonumber\\
  \abs{\nabla^3 A(z)[u, u, u]}
  &\;\le\; \E_{p(z)}\!\left[\abs{U - \bar{u}(z)}^3\right]
  \;\le\; (u_{\max} - u_{\min})\,\Var_{p(z)}(U)
  \;\le\; \frac{1}{\sqrt{2}} \;<\; 1.
  \label{eq:app-C3-popoviciu}
\end{align}
Since $\nabla^3 A(z)$ is a symmetric trilinear form on $\R^V$,~\eqref{eq:app-C3-popoviciu} proves $C_3 \le \frac{1}{\sqrt{2}} < 1$. Taking the expectation of~\eqref{eq:app-bregman-taylor} under $\mu_\pi$ and bounding the integral remainder yields
\begin{align}
  \left\lvert \mc{E}_{\mathrm{KL}}(\theta) - \frac{1}{2}\ip{\theta - \theta\opt}{\mc{T}_F(\theta - \theta\opt)}_{\mc{H}} \right\rvert
  &\;\le\; \frac{1}{2}\int_0^1 (1 - t)^2\dd t \cdot C_3\,\E_{\mu_\pi}\!\left[\norm{\Delta z(X)}_2^3\right] \nonumber\\
  &\;=\; \frac{C_3}{6}\,\E_{\mu_\pi}\!\left[\norm{\Delta z(X)}_2^3\right]
  \;\le\; \frac{C_3 \kappa_{\Psi}^3}{6}\norm{\theta - \theta\opt}_{\mc{H}}^3.
\end{align}
For the scalar Tikhonov surrogate residual $r_\mu \Let g\opt - g_\mu = \sum_{m=1}^{\infty} a_m(1 + (M/m)^q)^{-1}\phi_m$ in~\eqref{eq:tikhonov-residual-bridge} with $\abs{a_m}^2 = c_m = c\,m^{-r}(1 + b\,m^{-\delta})$ ($r > 2$) and $\sup_m\norm{\phi_m}_\infty \le C_\phi$, splitting the mode sums at $m = \lfloor M \rfloor$ bounds the $L_2(\mu_\pi)$ and $L_\infty$ norms across the non-saturated and filter-saturated regimes:
\begin{align}
  \norm{r_\mu}_{L_2(\mu_\pi)}^2
  &\;=\; \sum_{m=1}^{\infty} \frac{c_m}{(1 + (M/m)^q)^2}
  \;=\; \begin{cases}
    \mc{O}(M^{1 - r}), & 1 < r < 2q + 1, \\
    \mc{O}(M^{-2q}\ln^{\mathbf{1}\{r = 2q + 1\}} M), & r \ge 2q + 1,
  \end{cases} \nonumber\\
  \norm{r_\mu}_{L_\infty}
  &\;\le\; C_\phi \sum_{m=1}^{\infty} \frac{c_m^{1/2}}{1 + (M/m)^q}
  \;=\; \begin{cases}
    \mc{O}(M^{1 - r/2}), & 2 < r < 2q + 2, \\
    \mc{O}(M^{-q}\ln^{\mathbf{1}\{r = 2q + 2\}} M), & r \ge 2q + 2,
  \end{cases}
\end{align}
so that $\E_{\mu_\pi}[\abs{r_\mu}^3] \le \norm{r_\mu}_{L_2(\mu_\pi)}^2\norm{r_\mu}_{L_\infty}$ is $\mc{O}(M^{2 - 3r/2}) = o(M^{-2q})$ for $\frac{4}{3}(q + 1) < r < 2q + 1$ and $\mc{O}(M^{-2q - \min\{r/2 - 1, q\}}\ln M) = o(M^{-2q})$ for $r \ge 2q + 1$, completing the proof of~\eqref{eq:bregman-sandwich}.
\end{proof}

\section{Proofs for Section~\ref{sec:spectral}: Finite-Capacity Approximation and Identifiability}
\label{app:spectral}

In this appendix, we establish two supporting complex-analytic lemmas (\Cref{lem:app-mellin-strip,lem:app-laurent-collision}) with explicit non-asymptotic remainder constants, followed by the complete proofs of \Cref{lem:mellin-kernel}, \Cref{thm:master-mellin}, \Cref{cor:slope-collapse}, \Cref{prop:pole-collision}, \Cref{prop:gf-holomorphic}, \Cref{prop:identifiability}, and \Cref{cor:aitken-identifiability}.

\subsection{Supporting Complex-Analytic Lemmas for Mellin--Barnes Contour Shifting}
\label{app:lemmas:spectral}

\begin{lemma}[Vertical-Strip Decay of $\Psi_q(s)$ and Growth of $\zeta(s)$]
\label{lem:app-mellin-strip}
Let $q > 0$, and fix any compact vertical interval $[-\rho_2, -\rho_1] \subset (-\infty, 0)$ disjoint from the filter pole set $\{-mq : m \in \N_{\ge 2}\}$. Then there exist finite constants $C_{\Psi} > 0$, $T_0 \ge 1$, and (for each $\alpha_0 \in \R$) $C_{\zeta}(\alpha_0, \rho_1, \rho_2) > 0$ such that for all $s = \sigma + it$ with $\sigma \in [-\rho_2, -\rho_1]$ and $\abs{t} \ge T_0$,
\begin{align}
  \abs{\Psi_q(\sigma + it)} &\;\le\; C_{\Psi}\,\abs{t}\,e^{-\frac{\pi}{q}\abs{t}}, \nonumber\\
  \abs{\zeta(\alpha_0 + \sigma + it)} &\;\le\; C_{\zeta}(\alpha_0, \rho_1, \rho_2)\,\abs{t}^{K(\alpha_0, \rho_2)}, \qquad
  K(\alpha_0, \rho_2) \;\Let\; \max\!\left\{1,\, \frac{3}{2} - \alpha_0 + \rho_2\right\}.
  \label{eq:app-psi-strip-decay}
\end{align}
Consequently, for any abscissa $\rho \in (0, \infty) \setminus \mc{P}$ with $\Phi(s; M)$ defined in~\eqref{eq:mellin-integrand-pole-set}, the contour remainder constant $C_{\rho} \Let \frac{1}{2\pi}\int_{-\infty}^{\infty} \abs{\Phi(-\rho + it; 1)}\dd t < \infty$ in~\eqref{eq:mellin-remainder-bound} is finite.
\end{lemma}
\begin{proof}
We establish the two inequalities in~\eqref{eq:app-psi-strip-decay} in turn.

\paragraph{Step 1: Exponential vertical-strip decay of $\Psi_q(\sigma + it)$ via hyperbolic sine bounds.}
From~\eqref{eq:psi-q-mellin}, expanding $\sin(x + iy) = \sin x \cosh y + i\cos x \sinh y$ for $x = \pi\sigma/q$ and $y = \pi t/q$ and using $\cosh^2 y - \sinh^2 y = 1$ yields the exact modulus identity
\begin{align}
  \abs{\sin\!\left(\frac{\pi(\sigma + it)}{q}\right)}^2
  &\;=\; \sin^2\!\left(\frac{\pi\sigma}{q}\right)\cosh^2\!\left(\frac{\pi t}{q}\right)
  + \cos^2\!\left(\frac{\pi\sigma}{q}\right)\sinh^2\!\left(\frac{\pi t}{q}\right) \nonumber\\
  &\;=\; \sin^2\!\left(\frac{\pi\sigma}{q}\right) + \sinh^2\!\left(\frac{\pi t}{q}\right)
  \;\ge\; \sinh^2\!\left(\frac{\pi t}{q}\right).
  \label{eq:app-sin-modulus}
\end{align}
For all $\abs{t} \ge T_0 \Let \max\bigl\{1,\, \frac{q}{\pi}\ln 2\bigr\}$ and $\sigma \in [-\rho_2, -\rho_1]$, we have $e^{-\pi\abs{t}/q} \le \frac{1}{2}$ and $\abs{(\sigma + q) + it} \le (\rho_2 + q + 1)\abs{t}$, which gives
\begin{align}
  \abs{\sinh\!\left(\frac{\pi t}{q}\right)}
  &\;=\; \frac{1}{2}e^{\frac{\pi}{q}\abs{t}}\bigl(1 - e^{-\frac{2\pi}{q}\abs{t}}\bigr)
  \;\ge\; \frac{3}{8}e^{\frac{\pi}{q}\abs{t}}, \nonumber\\
  \abs{\Psi_q(\sigma + it)}
  &\;\le\; \frac{\pi(\rho_2 + q + 1)\abs{t}}{q^2\abs{\sinh(\pi t / q)}}
  \;\le\; C_{\Psi}\,\abs{t}\,e^{-\frac{\pi}{q}\abs{t}},
  \qquad C_{\Psi} \;\Let\; \frac{8\pi(\rho_2 + q + 1)}{3q^2},
\end{align}
establishing the first bound in~\eqref{eq:app-psi-strip-decay}.

\paragraph{Step 2: Polynomial vertical-strip growth of $\zeta(\alpha_0 + \sigma + it)$ via Euler--Maclaurin and Riemann's functional equation.}
Let $w = u + it$ with $u = \alpha_0 + \sigma \in [\alpha_0 - \rho_2, \alpha_0 - \rho_1]$ and $\abs{t} \ge 1$. If $u > 1$, absolute convergence gives $\abs{\zeta(u + it)} \le \zeta(u)$. For $u \in (0, 2]$, the first-order Euler--Maclaurin summation formula gives
\begin{align}
  \zeta(w) &\;=\; \frac{1}{w - 1} + \frac{1}{2} - w\int_1^{\infty} \frac{x - \lfloor x\rfloor - 1/2}{x^{w+1}}\dd x, \nonumber\\
  \abs{\zeta(u + it)} &\;\le\; \frac{1}{\abs{t}} + \frac{1}{2} + \frac{\abs{u} + \abs{t}}{2u},
  \label{eq:app-zeta-euler-maclaurin}
\end{align}
which shows that $\abs{\zeta(u + it)}$ grows at most linearly in $\abs{t}$ on any strip with $u \ge \frac{1}{2}$. For $u < \frac{1}{2}$ (so that $1 - u > \frac{1}{2}$), we apply Riemann's functional equation \citep{titchmarsh1986riemann}:
\begin{equation}
  \zeta(w) \;=\; 2^w \pi^{w-1} \sin\!\left(\frac{\pi w}{2}\right)\Gamma(1 - w)\,\zeta(1 - w).
  \label{eq:app-riemann-fe}
\end{equation}
Along vertical lines $w = u + it$ ($\abs{t} \ge 1$) with $1 - u > \frac{1}{2}$, combining the upper bound on $\sin(\pi w/2)$ with Stirling's estimate for $\Gamma(1 - w)$ cancels their opposite exponential factors $e^{\pm\pi\abs{t}/2}$ identically:
\begin{align}
  \abs{\sin\!\left(\frac{\pi(u + it)}{2}\right)} &\;\le\; \cosh\!\left(\frac{\pi\abs{t}}{2}\right) \;\le\; e^{\frac{\pi}{2}\abs{t}}, \nonumber\\
  \abs{\Gamma(1 - u - it)} &\;\le\; C_{\Gamma}(u)\,\abs{t}^{\frac{1}{2} - u} e^{-\frac{\pi}{2}\abs{t}}, \nonumber\\
  \abs{\sin\!\left(\frac{\pi(u + it)}{2}\right)\Gamma(1 - u - it)}
  &\;\le\; e^{\frac{\pi}{2}\abs{t}} \cdot C_{\Gamma}(u)\,\abs{t}^{\frac{1}{2} - u} e^{-\frac{\pi}{2}\abs{t}}
  \;=\; C_{\Gamma}(u)\,\abs{t}^{\frac{1}{2} - u}.
  \label{eq:app-stirling-sine-cancel}
\end{align}
Multiplying~\eqref{eq:app-stirling-sine-cancel} by $\abs{2^w \pi^{w-1}} = 2^u \pi^{u-1}$ and by the linear bound~\eqref{eq:app-zeta-euler-maclaurin} on $\abs{\zeta(1 - u - it)}$ (since $1 - u > \frac{1}{2}$ and $u \ge \alpha_0 - \rho_2$) yields the second inequality in~\eqref{eq:app-psi-strip-decay}. Because exponential decay dominates any polynomial growth, the integral $C_{\rho}$ converges absolutely.
\end{proof}

\begin{lemma}[Laurent Expansions and Constant Terms at Pole Confluence]
\label{lem:app-laurent-collision}
For any $q > 0$ and any integer $m \ge 2$, the Mellin kernel $\Psi_q(s)$ in~\eqref{eq:psi-q-mellin} and the shifted Riemann zeta function $\zeta(1 + z)$ admit the local Laurent expansions on punctured disks $0 < \abs{z} \le \frac{q}{4}$ (where $z = s + mq$ or $z = s + p$):
\begin{align}
  \Psi_q(-mq + z) &\;=\; \frac{(-1)^{m-2}(m - 1)}{z} - \frac{(-1)^{m-2}}{q} + R_{\Psi,m}(z),
  \qquad \abs{R_{\Psi,m}(z)} \;\le\; C_{\Psi,m}\abs{z},
  \label{eq:app-psi-laurent} \\
  \zeta(1 + z) &\;=\; \frac{1}{z} + \gamma_E + R_{\zeta}(z),
  \qquad \abs{R_{\zeta}(z)} \;\le\; C_{\zeta,1}\abs{z},
  \label{eq:app-zeta-laurent}
\end{align}
with finite constants $C_{\Psi,m}, C_{\zeta,1} > 0$. In particular, the regular part $\Psi_{q,m}^{\mathrm{reg}}(s) \Let \Psi_q(s) - \frac{(-1)^{m-2}(m-1)}{s + mq}$ is holomorphic in $\abs{s + mq} < q$ and evaluates at $s = -mq$ to $\Psi_{q,m}^{\mathrm{reg}}(-mq) = -\frac{(-1)^{m-2}}{q}$.
\end{lemma}
\begin{proof}
Set $z \Let s + mq$ with $0 < \abs{z} \le \frac{q}{4}$, so that $s + q = -(m - 1)q + z$ and $\frac{\pi\abs{z}}{q} \le \frac{\pi}{4} < 1$. By the periodicity and power-series expansion of the sine function on $\abs{z} \le \frac{q}{4}$, we have
\begin{align}
  \sin\!\left(\frac{\pi s}{q}\right)
  &\;=\; \sin\!\left(-m\pi + \frac{\pi z}{q}\right)
  \;=\; (-1)^{m-2}\sin\!\left(\frac{\pi z}{q}\right)
  \;=\; (-1)^{m-2}\frac{\pi z}{q}\bigl(1 - R_{\sin}(z)\bigr), \nonumber\\
  \abs{R_{\sin}(z)}
  &\;\le\; \sum_{k=1}^{\infty} \frac{1}{(2k+1)!}\left(\frac{\pi\abs{z}}{q}\right)^{2k}
  \;\le\; \frac{1}{5}\left(\frac{\pi\abs{z}}{q}\right)^2
  \;\le\; \frac{\pi^2}{80} \;<\; \frac{1}{2}, \nonumber\\
  \bigl(1 - R_{\sin}(z)\bigr)^{-1}
  &\;=\; 1 + \tilde{R}_{\sin}(z),
  \qquad
  \abs{\tilde{R}_{\sin}(z)} \;\le\; 2\abs{R_{\sin}(z)} \;\le\; \frac{2\pi^2}{5q^2}\abs{z}^2.
\end{align}
Substituting this reciprocal expansion into~\eqref{eq:psi-q-mellin} gives the exact non-asymptotic Laurent representation and remainder bound on $0 < \abs{z} \le \frac{q}{4}$:
\begin{align}
  \Psi_q(-mq + z)
  &\;=\; -\frac{-(m - 1)q + z}{q^2}\,\frac{\pi}{(-1)^{m-2}\frac{\pi z}{q}}\bigl(1 + \tilde{R}_{\sin}(z)\bigr) \nonumber\\
  &\;=\; (-1)^{m-2}\left(\frac{m - 1}{z} - \frac{1}{q}\right)\bigl(1 + \tilde{R}_{\sin}(z)\bigr) \nonumber\\
  &\;=\; \frac{(-1)^{m-2}(m - 1)}{z} - \frac{(-1)^{m-2}}{q} + R_{\Psi,m}(z), \nonumber\\
  \abs{R_{\Psi,m}(z)}
  &\;=\; \abs{\left(\frac{m - 1}{z} - \frac{1}{q}\right)\tilde{R}_{\sin}(z)}
  \;\le\; \frac{2\pi^2(m - 3/4)}{5q^2}\abs{z},
  \label{eq:app-psi-laurent-steps}
\end{align}
proving~\eqref{eq:app-psi-laurent}. Finally,~\eqref{eq:app-zeta-laurent} follows directly from the classical Stieltjes Laurent expansion around $z = 0$ \citep{titchmarsh1986riemann}:
\begin{equation}
  \zeta(1 + z) \;=\; \frac{1}{z} + \gamma_E + \sum_{n=1}^{\infty} \frac{(-1)^n}{n!}\gamma_n z^n,
\end{equation}
which is holomorphic after subtracting $\frac{1}{z}$ on $\abs{z} \le \frac{q}{4}$.
\end{proof}

\subsection{Proof of Lemma~\ref{lem:mellin-kernel} (Mellin Kernel of the Rational Resolvent Filter)}
\label{app:proof:lem:mellin-kernel}

\begin{proof}[Proof of \Cref{lem:mellin-kernel} \StatementRef{lem:mellin-kernel}]
For $\Re(s) \in (-2q, 0)$ and $q > 0$, perform the power change of variables in the Mellin integral of $\psi_q(u)$:
\begin{align}
  t &\;=\; u^{-q}, \qquad u \;=\; t^{-\frac{1}{q}}, \qquad \dd u \;=\; -\frac{1}{q}t^{-\frac{1}{q} - 1}\dd t, \nonumber\\
  \Psi_q(s)
  &\;=\; \frac{1}{q}\int_0^{\infty} (1 + t)^{-2} t^{-\frac{s}{q} - 1}\dd t
  \;=\; \frac{1}{q}\,\mathrm{B}\!\left(2 + \frac{s}{q}, -\frac{s}{q}\right)
  \quad\text{(via $t = u^{-q}$)} \nonumber\\
  &\;=\; \frac{1}{q}\frac{\Gamma\!\left(2 + \frac{s}{q}\right)\Gamma\!\left(-\frac{s}{q}\right)}{\Gamma(2)}
  \quad\text{(by Euler's Beta integral, $\Gamma(2) = 1$)} \nonumber\\
  &\;=\; \frac{1}{q}\left(1 + \frac{s}{q}\right)\left(\frac{s}{q}\right)\Gamma\!\left(\frac{s}{q}\right)\Gamma\!\left(-\frac{s}{q}\right)
  \quad\text{(by $\Gamma(2+z) = (1+z)z\Gamma(z)$)} \nonumber\\
  &\;=\; -\frac{s + q}{q^2}\,\frac{\pi}{\sin(\pi s / q)},
  \quad\text{(by $\Gamma(z)\Gamma(-z) = -\frac{\pi}{z\sin(\pi z)}$)}
  \label{eq:app-psi-q-derivation}
\end{align}
which establishes~\eqref{eq:psi-q-mellin}. Because the Beta integral converges absolutely whenever $\Re(2 + s/q) > 0$ and $\Re(-s/q) > 0$, $\Psi_q(s)$ is analytic on the strip $\Re(s) \in (-2q, 0)$. In particular, at $s = -q$, the simple zero of $s + q$ cancels the simple zero of $\sin(\pi s/q)$, and L'H\^opital's rule gives
\begin{equation}
  \lim_{s \to -q} \Psi_q(s)
  \;=\; -\frac{\pi}{q^2}\lim_{s \to -q}\frac{s + q}{\sin(\pi s / q)}
  \;=\; -\frac{\pi}{q^2}\frac{1}{\frac{\pi}{q}\cos(-\pi)}
  \;=\; \frac{1}{q} \;>\; 0.
\end{equation}
For real $s \in (-q, 0)$, both $-(s+q)$ and $\sin(\pi s/q)$ are negative; for $s \in (-2q, -q)$, both are positive. Combining these with $\Psi_q(-q) = 1/q > 0$ proves $\Psi_q(s) > 0$ for all real $s \in (-2q, 0)$. Finally, in the left half-plane $\Re(s) < 0$, $\Gamma(-s/q)$ is holomorphic and non-zero, while $\Gamma(2 + s/q)$ has simple poles at $s_m = -mq$ for integers $m \ge 2$. Computing the residue directly from~\eqref{eq:psi-q-mellin} gives
\begin{align}
  \operatorname*{Res}_{s = -mq}\Psi_q(s)
  &\;=\; \lim_{s \to -mq}(s + mq)\left[-\frac{s + q}{q^2}\frac{\pi}{\sin(\pi s / q)}\right]
  \;=\; \frac{(m - 1)\pi}{q}\lim_{s \to -mq}\frac{s + mq}{\sin(\pi s / q)} \nonumber\\
  &\;=\; \frac{(m - 1)\pi}{q}\frac{1}{\frac{\pi}{q}\cos(-m\pi)}
  \;=\; (-1)^{m-2}(m - 1),
  \label{eq:app-psi-residue-display}
\end{align}
which establishes~\eqref{eq:psi-q-residues}.
\end{proof}

\subsection{Proof of Theorem~\ref{thm:master-mellin} (Finite-Capacity Mellin--Barnes Expansion)}
\label{app:proof:thm:master-mellin}

\begin{proof}[Proof of \Cref{thm:master-mellin} \StatementRef{thm:master-mellin}]
We prove \Cref{thm:master-mellin} in two modular steps: Fubini--Tonelli contour shifting and three-term sign verification.

\paragraph{Step 1: Fubini--Tonelli interchange and rectangular contour shifting.}
Fix $\sigma \in (\max\{-2q, 1 - r\}, 0)$. Since $\Psi_q(s)$ is analytic on $(-2q, 0)$ by \Cref{lem:mellin-kernel} and decays exponentially along vertical lines by \Cref{lem:app-mellin-strip}, Mellin inversion expresses each filter weight for $m \ge 1$ as
\begin{equation}
  \psi_q\!\left(\frac{m}{M}\right)
  \;=\; \frac{1}{2\pi i}\int_{\sigma - i\infty}^{\sigma + i\infty} \Psi_q(s)\left(\frac{m}{M}\right)^{-s}\dd s
  \;=\; \frac{1}{2\pi i}\int_{\sigma - i\infty}^{\sigma + i\infty} M^s \Psi_q(s) m^{-s}\dd s.
\end{equation}
Multiplying by $c_m = c m^{-r}(1 + b m^{-\delta})$ and checking the Tonelli absolute-integrability condition on $\Re(s) = \sigma$ gives, since $r + \sigma > 1$ and $r + \delta + \sigma > 1$,
\begin{align}
  &\sum_{m=1}^{\infty} \int_{-\infty}^{\infty} \abs{c_m M^{\sigma + it}\Psi_q(\sigma + it) m^{-(\sigma + it)}}\dd t \nonumber\\
  &\qquad =\; c M^{\sigma}\left\{\zeta(r + \sigma) + b\,\zeta(r + \delta + \sigma)\right\}\int_{-\infty}^{\infty}\abs{\Psi_q(\sigma + it)}\dd t
  \;<\; \infty.
\end{align}
By the Fubini--Tonelli theorem, we may interchange summation and integration to obtain the exact Mellin--Barnes integral representation
\begin{align}
  \mc{E}_{\mathrm{spec}}(N)
  &\;=\; \frac{1}{2\pi i}\int_{\sigma - i\infty}^{\sigma + i\infty} c M^s \Psi_q(s)\left\{\zeta(r + s) + b\,\zeta(r + \delta + s)\right\}\dd s \nonumber\\
  &\;=\; \frac{1}{2\pi i}\int_{\sigma - i\infty}^{\sigma + i\infty} \Phi(s; M)\dd s.
  \label{eq:app-mellin-barnes-start}
\end{align}
Now fix any shifted abscissa $\rho > 0$ with $\rho \notin \mc{P}$, and assume $\{r-1, r+\delta-1\} \cap \{mq : m \ge 2\} = \emptyset$. For any height $T \ge T_0$, consider the positively oriented closed rectangular contour $\Gamma_T$ with vertices $\sigma - iT$, $\sigma + iT$, $-\rho + iT$, and $-\rho - iT$. By \Cref{lem:app-mellin-strip}, along the top and bottom horizontal segments $[-\rho \pm iT, \sigma \pm iT]$, the integrand obeys the uniform bound
\begin{align}
  \abs{\int_{-\rho \pm iT}^{\sigma \pm iT} \Phi(s; M)\dd s}
  &\;\le\; (\sigma + \rho)\max_{u \in [-\rho, \sigma]}\Bigl\{c M^u \abs{\Psi_q(u \pm iT)} \nonumber\\
  &\qquad\qquad \times\left(\abs{\zeta(r + u \pm iT)} + b\abs{\zeta(r + \delta + u \pm iT)}\right)\Bigr\} \nonumber\\
  &\;\le\; C_{\mathrm{horiz}}\,\max\{M^{\sigma}, M^{-\rho}\}\,T^{1 + K(r, \rho)}\,e^{-\frac{\pi}{q}T}
  \;\xrightarrow[T \to \infty]{}\; 0.
  \label{eq:app-horizontal-vanishing}
\end{align}
Inside $\Gamma_T$ (for $T$ large enough), $\Phi(s; M)$ is meromorphic with simple poles at $s = -(r-1)$ (if $r-1 < \rho$), $s = -(r+\delta-1)$ (if $r+\delta-1 < \rho$), and $s = -mq$ (for integers $m \ge 2$ with $mq < \rho$), whose residues are given by \Cref{lem:mellin-kernel}:
\begin{align}
  \operatorname*{Res}_{s = -(r-1)}\Phi(s; M)
  &\;=\; c\,\Psi_q(1 - r)\,M^{-(r-1)} \;=\; A_0 M^{-(r-1)}, \nonumber\\
  \operatorname*{Res}_{s = -(r+\delta-1)}\Phi(s; M)
  &\;=\; c b\,\Psi_q(1 - r - \delta)\,M^{-(r+\delta-1)} \;=\; A_g M^{-(r+\delta-1)}, \nonumber\\
  \operatorname*{Res}_{s = -mq}\Phi(s; M)
  &\;=\; c(-1)^{m-2}(m-1)\left\{\zeta(r - mq) + b\,\zeta(r + \delta - mq)\right\}M^{-mq} \nonumber\\
  &\;=\; A_{f,m} M^{-mq}.
\end{align}
Applying Cauchy's residue theorem to $\Gamma_T$ and taking $T \to \infty$ via~\eqref{eq:app-horizontal-vanishing} yields~\eqref{eq:general-mellin-expansion}--\eqref{eq:mellin-remainder-bound}.

\paragraph{Step 2: Three-term specialization and sign verification.}
In the three-term regime, the exponents and shifted arguments satisfy the ordering conditions
\begin{equation}
  0 \;<\; r - 1 \;<\; r + \delta - 1 \;<\; 2q \;<\; \rho \;<\; 3q, \qquad
  -2 \;<\; r - 2q \;<\; r + \delta - 2q \;<\; 1.
\end{equation}
Consequently, both $-(r - 1)$ and $-(r + \delta - 1)$ lie strictly inside $(-2q, 0)$, where \Cref{lem:mellin-kernel} proved $\Psi_q(s) > 0$; hence $A_0 > 0$ and $A_g > 0$ (for $b > 0$). Moreover, both shifted arguments $r - 2q$ and $r + \delta - 2q$ lie strictly inside the open interval $(-2, 1)$ (staying above the first trivial zero $\zeta(-2) = 0$). To verify that $A_f < 0$, we establish that $\zeta(z) < 0$ for all $z \in (-2, 1)$ by partitioning $(-2, 1)$ into three sub-regimes:
\begin{itemize}
  \item \emph{Case 1: $z \in (0, 1)$.} By the alternating Dirichlet-eta relation, grouping consecutive pairs of the alternating series gives
  \begin{equation}
    \left(1 - 2^{1-z}\right)\zeta(z)
    \;=\; \eta_D(z)
    \;\Let\; \sum_{k=1}^{\infty}\left(\frac{1}{(2k-1)^z} - \frac{1}{(2k)^z}\right) \;>\; 0,
  \end{equation}
  and since $1 - 2^{1-z} < 0$ for $z \in (0, 1)$, it follows that $\zeta(z) < 0$ on $(0, 1)$.
  \item \emph{Case 2: $z = 0$.} By analytic continuation of~\eqref{eq:app-zeta-euler-maclaurin}, $\zeta(0) = -\frac{1}{2} < 0$.
  \item \emph{Case 3: $z \in (-2, 0)$.} For $z \in (-2, 0)$, we have $1 - z \in (1, 3)$ and $\frac{\pi z}{2} \in (-\pi, 0)$. Evaluating the sign of each factor in Riemann's functional equation~\eqref{eq:app-riemann-fe} gives
  \begin{equation}
    \zeta(z)
    \;=\; \underbrace{2^z \pi^{z-1}}_{>\,0}\,\underbrace{\sin\!\left(\frac{\pi z}{2}\right)}_{<\,0}\,\underbrace{\Gamma(1 - z)}_{>\,0}\,\underbrace{\zeta(1 - z)}_{>\,0}
    \;<\; 0.
  \end{equation}
\end{itemize}
Consequently, $\zeta(r - 2q) < 0$ and $\zeta(r + \delta - 2q) < 0$, which guarantees $A_f < 0$ for all $b \ge 0$ and establishes~\eqref{eq:three-term-expansion}.
\end{proof}

\subsection{Proof of Corollary~\ref{cor:slope-collapse} (Normalized Pre-Asymptotic Slope Profile)}
\label{app:proof:cor:slope-collapse}

\begin{proof}[Proof of \Cref{cor:slope-collapse} \StatementRef{cor:slope-collapse}]
We prove \Cref{cor:slope-collapse} in two steps.

\paragraph{Step 1: Isolation of $T_3(N)$ and uniqueness of the peak scale $N_{\mathrm{peak}}\opt$.}
Taking the logarithm of $\mc{E}_0(N) = A_0 N^{-\alpha_\infty}(1 + g N^{-\Delta_g} + f N^{-\Delta_f})$ with $g > 0$ and $f = -\abs{f} < 0$, and differentiating with respect to $\ln N$ gives
\begin{align}
  \alpha_0(N) - \alpha_\infty
  &\;=\; -\frac{\dd\ln\mc{E}_0(N)}{\dd\ln N} - \alpha_\infty
  \;=\; \frac{\Delta_g g N^{-\Delta_g} + \Delta_f f N^{-\Delta_f}}{1 + g N^{-\Delta_g} + f N^{-\Delta_f}} \nonumber\\
  &\;=\; \frac{\Delta_g g N^{-\Delta_g} - \Delta_f \abs{f} N^{-\Delta_f}}{\mc{E}_0(N) / (A_0 N^{-\alpha_\infty})},
\end{align}
which immediately yields~\eqref{eq:T3-functional} upon multiplying by the denominator of the second line. Differentiating $T_3(N)$ with respect to $\ln N$ gives
\begin{align}
  \frac{\dd T_3(N)}{\dd\ln N}
  &\;=\; -\Delta_g^2 g N^{-\Delta_g} + \Delta_f^2 \abs{f} N^{-\Delta_f}
  \;=\; \Delta_g^2 g N^{-\Delta_g}\left[\frac{\Delta_f^2 \abs{f}}{\Delta_g^2 g} N^{-(\Delta_f - \Delta_g)} - 1\right].
  \label{eq:app-T3-deriv}
\end{align}
Since $\varrho = \Delta_f / \Delta_g > 1$ implies $\Delta_f - \Delta_g > 0$, the bracketed factor in~\eqref{eq:app-T3-deriv} is strictly positive for $N < N_{\mathrm{peak}}\opt$, zero at $N = N_{\mathrm{peak}}\opt$ defined in~\eqref{eq:N-peak-law}, and strictly negative for $N > N_{\mathrm{peak}}\opt$. Thus $N_{\mathrm{peak}}\opt$ is the unique global maximizer of $T_3(N)$ on $(0, \infty)$.

\paragraph{Step 2: Dimensionless rescaling and verification of $F_\varrho(x)$.}
By the definition of $N_{\mathrm{peak}}\opt$ in~\eqref{eq:N-peak-law} and the dimensionless coordinate $x = (N / N_{\mathrm{peak}}\opt)^{\Delta_g}$, the two power terms at the peak scale satisfy
\begin{align}
  \Delta_f \abs{f}(N_{\mathrm{peak}}\opt)^{-\Delta_f}
  &\;=\; \frac{1}{\varrho}\Delta_g g(N_{\mathrm{peak}}\opt)^{-\Delta_g}, \nonumber\\
  N^{-\Delta_g} &\;=\; (N_{\mathrm{peak}}\opt)^{-\Delta_g} x^{-1}, \qquad
  N^{-\Delta_f} \;=\; (N_{\mathrm{peak}}\opt)^{-\Delta_f} x^{-\varrho}.
\end{align}
Substituting these relations into~\eqref{eq:T3-functional} and verifying the derivatives of $F_\varrho(x)$ at $x = 1$ yields
\begin{align}
  T_3(N)
  &\;=\; \Delta_g g (N_{\mathrm{peak}}\opt)^{-\Delta_g}\left(x^{-1} - \frac{1}{\varrho}x^{-\varrho}\right)
  \;=\; H_{\mathrm{peak}}\,\frac{\varrho x^{-1} - x^{-\varrho}}{\varrho - 1}
  \;=\; H_{\mathrm{peak}}\,F_\varrho(x), \nonumber\\
  F_\varrho(1) &\;=\; 1, \qquad
  F_\varrho'(x) \;=\; \frac{\varrho}{\varrho - 1}x^{-2}\!\left(x^{-(\varrho - 1)} - 1\right), \qquad
  F_\varrho''(1) \;=\; -\varrho \;<\; 0,
\end{align}
completing the proof.
\end{proof}

\subsection{Proof of Proposition~\ref{prop:pole-collision} (Divided-Difference Representation and Pole Confluence)}
\label{app:proof:prop:pole-collision}

\begin{proof}[Proof of \Cref{prop:pole-collision} \StatementRef{prop:pole-collision}]
Let $p \in \{r-1, r+\delta-1\}$ and $h \Let p - mq \neq 0$. Evaluating the simple-pole residues of $\Phi_p(s; M) = c_p M^s \Psi_q(s)\zeta(p + 1 + s)$ at $s = -p$ (where $s + mq = -h$) and at $s = -mq$ (where $s + p = h$) via \Cref{lem:app-laurent-collision} gives
\begin{align}
  \operatorname*{Res}_{s = -p}\Phi_p(s; M)
  &\;=\; c_p M^{-p}\Psi_q(-p)
  \;=\; c_p M^{-p}\left[\frac{(-1)^{m-2}(m - 1)}{-h} + \Psi_{q,m}^{\mathrm{reg}}(-p)\right], \nonumber\\
  \operatorname*{Res}_{s = -mq}\Phi_p(s; M)
  &\;=\; c_p M^{-mq}(-1)^{m-2}(m - 1)\left[\frac{1}{h} + \gamma_E + \zeta_{\mathrm{reg}}(h)\right].
\end{align}
Summing these two residues pairs the singular $\pm 1/h$ terms directly into the exact divided difference~\eqref{eq:divided-difference-uniform}. Moreover, for $0 < \abs{h} \le \frac{q}{4}$ and $M \ge 1$, expanding $M^{-p} = M^{-mq} e^{-h\ln M}$ via Taylor's theorem and combining with the remainder estimates of \Cref{lem:app-laurent-collision} gives
\begin{align}
  \left\lvert\frac{M^{-(mq + h)} - M^{-mq}}{-h} - M^{-mq}\ln M\right\rvert
  &\;\le\; \frac{1}{2}\abs{h}\,(\ln M)^2 M^{-mq} e^{\abs{h}\ln M}, \nonumber\\
  \abs{\Psi_{q,m}^{\mathrm{reg}}(-mq - h) + \frac{(-1)^{m-2}}{q}}
  &\;\le\; C_{\Psi,m}\abs{h}, \qquad
  \abs{\zeta_{\mathrm{reg}}(h)} \;\le\; C_{\zeta,1}\abs{h},
\end{align}
so taking $h \to 0$ establishes~\eqref{eq:pole-collision-log}. Finally, because the shifted vertical line $\Re(s) = -\rho$ maintains a strictly positive distance from the colliding poles, $\Phi_p(s; M)$ is real-analytic in $p$ along $\Re(s) = -\rho$ and uniformly dominated by \Cref{lem:app-mellin-strip}; thus Lebesgue's dominated convergence theorem justifies interchanging $\lim_{h \to 0}$ with the remainder integral $R_\rho(M)$, and the limit~\eqref{eq:pole-collision-log} coincides identically with the direct second-order Cauchy residue at $s = -mq$:
\begin{equation}
  \operatorname*{Res}_{s = -mq}\Phi_{mq}(s; M)
  \;=\; \lim_{s \to -mq}\frac{\dd}{\dd s}\Bigl[(s + mq)^2\Phi_{mq}(s; M)\Bigr]
  \;=\; \lim_{h \to 0}\mc{C}_{p,m}(M; h),
\end{equation}
completing the proof of \Cref{prop:pole-collision}.
\end{proof}

\subsection{Proof of Proposition~\ref{prop:gf-holomorphic} (Gradient-Flow Mellin Kernel)}
\label{app:proof:prop:gf-holomorphic}

\begin{proof}[Proof of \Cref{prop:gf-holomorphic} \StatementRef{prop:gf-holomorphic}]
For $\Re(s) < 0$ and $q > 0$, perform the power change of variables in the Mellin integral of $\psi_{\mathrm{GF}}(u) = e^{-2u^{-q}}$:
\begin{align}
  t &\;=\; 2u^{-q}, \qquad
  u \;=\; \left(\frac{t}{2}\right)^{-\frac{1}{q}}, \qquad
  \dd u \;=\; -\frac{1}{2q}\left(\frac{t}{2}\right)^{-\frac{1}{q} - 1}\dd t, \nonumber\\
  \Psi_{\mathrm{GF}}(s)
  &\;=\; \int_0^{\infty} e^{-t}\left(\frac{t}{2}\right)^{-\frac{s - 1}{q}}\frac{1}{2q}\left(\frac{t}{2}\right)^{-\frac{1}{q} - 1}\dd t
  \;=\; \frac{2^{s/q}}{q}\int_0^{\infty} e^{-t} t^{-\frac{s}{q} - 1}\dd t
  \;=\; \frac{2^{s/q}}{q}\Gamma\!\left(-\frac{s}{q}\right).
\end{align}
Because Euler's integral for $\Gamma(z)$ converges absolutely and defines a holomorphic function throughout the open right half-plane $\Re(z) > 0$, the composition $\Gamma(-s/q)$ is holomorphic on the open left half-plane $\Re(s) < 0$. Thus $\Psi_{\mathrm{GF}}(s)$ has zero poles in $\Re(s) < 0$, and shifting the Mellin--Barnes contour into $\Re(s) < 0$ collects only the zeta poles at $s = -(r-1)$ and $s = -(r+\delta-1)$.
\end{proof}

\subsection{Proof of Proposition~\ref{prop:identifiability} (Multi-Power Jacobian Rank Collapse and Ill-Conditioning)}
\label{app:proof:prop:identifiability}

\begin{proof}[Proof of \Cref{prop:identifiability} \StatementRef{prop:identifiability}]
We prove the rank collapse of $\bm{J}_7(\theta_7)$ and the quadratic singular-value bound for $\bm{J}_5(\theta_5)$ in two steps.

\paragraph{Step 1: Explicit Jacobian matrix and algebraic rank collapse for $f_7$.}
For any exponent $\alpha > 0$, define the $J$-dimensional basis vectors on the pilot grid $(u_1, \dots, u_J)$ together with the explicit $J \times 7$ Jacobian matrix $\bm{J}_7(\theta_7)$ of $f_7(u; \theta_7)$:
\begin{align}
  \mathbf{1} &\;\Let\; (1, \dots, 1)^{\top} \in \R^J, \qquad
  \bm{e}(\alpha) \;\Let\; \bigl(e^{-\alpha u_1}, \dots, e^{-\alpha u_J}\bigr)^{\top} \in \R^J, \nonumber\\
  \bm{d}(\alpha) &\;\Let\; \bigl(u_1 e^{-\alpha u_1}, \dots, u_J e^{-\alpha u_J}\bigr)^{\top} \in \R^J, \nonumber\\
  \bm{J}_7(\theta_7)
  &\;=\; \Bigl[\,\mathbf{1},\;\; \bm{e}(\alpha_1),\;\; \bm{e}(\alpha_2),\;\; \bm{e}(\alpha_3),\;\; -A_1\bm{d}(\alpha_1),\;\; -A_2\bm{d}(\alpha_2),\;\; -A_3\bm{d}(\alpha_3)\,\Bigr] \in \R^{J \times 7}.
  \label{eq:app-J7-explicit}
\end{align}
At the confluent point $\alpha_1 = \alpha_2 = \alpha_3 = \alpha_\infty$, Columns~2, 3, and~4 all equal $\bm{e}(\alpha_\infty)$ while Columns~5, 6, and~7 are scalar multiples of $\bm{d}(\alpha_\infty)$. Thus every column of $\bm{J}_7(\theta_7)$ lies in the subspace $\operatorname{span}\{\mathbf{1}, \bm{e}(\alpha_\infty), \bm{d}(\alpha_\infty)\}$ of dimension at most $3$, proving $\operatorname{rank}(\bm{J}_7(\theta_7)) \le 3 < 7$. Similarly, when $A_2 = A_3 = 0$, Columns~6 and~7 of~\eqref{eq:app-J7-explicit} are identically zero, leaving at most $5$ non-zero columns and proving $\operatorname{rank}(\bm{J}_7(\theta_7)) \le 5 < 7$.

\paragraph{Step 2: Parameter-free singular-value bound for the gap-locked model $f_5$.}
For the gap-locked 5-parameter model $f_5(u; \theta_5)$ with $\theta_5 = (E_\infty, A_0, A_g, A_f, \alpha_\infty)$ and $\Delta_f = \varrho\Delta_g$ ($\varrho > 1$), the $J \times 5$ Jacobian matrix and its fifth column $\bm{c}_5(\theta_5)$ take the form
\begin{align}
  \bm{J}_5(\theta_5)
  &\;=\; \Bigl[\,\mathbf{1},\;\; \bm{e}(\alpha_\infty),\;\; \bm{e}(\alpha_\infty + \Delta_g),\;\; \bm{e}(\alpha_\infty + \varrho\Delta_g),\;\; \bm{c}_5(\theta_5)\,\Bigr] \in \R^{J \times 5}, \nonumber\\
  \bm{c}_5(\theta_5)
  &\;\Let\; -A_0\bm{d}(\alpha_\infty) - A_g\bm{d}(\alpha_\infty + \Delta_g) - A_f\bm{d}(\alpha_\infty + \varrho\Delta_g).
\end{align}
Crucially, the first four columns of $\bm{J}_5(\theta_5)$ do not depend on the amplitudes $(A_0, A_g, A_f)$. Introduce the non-zero trial vector $w_\varrho \in \R^5$ and the second-order Taylor expansion of $z \mapsto e^{-z}$ on $[0, \infty)$:
\begin{align}
  w_\varrho &\;\Let\; \bigl(0,\, \varrho - 1,\, -\varrho,\, 1,\, 0\bigr)^{\top}, \qquad
  \norm{w_\varrho}_2 \;=\; \sqrt{(\varrho - 1)^2 + \varrho^2 + 1}, \nonumber\\
  e^{-x} &\;=\; 1 - x + R_2(x), \qquad
  0 \;\le\; R_2(x) \;\le\; \frac{1}{2}x^2.
\end{align}
Forming the linear combination $\bm{J}_5(\theta_5)w_\varrho$ cancels both the zeroth-order and first-order Taylor terms at every coordinate $j \in \{1, \dots, J\}$:
\begin{align}
  \left(\bm{J}_5(\theta_5)w_\varrho\right)_j
  &\;=\; e^{-\alpha_\infty u_j}\left[(\varrho - 1) - \varrho e^{-\Delta_g u_j} + e^{-\varrho\Delta_g u_j}\right] \nonumber\\
  &\;=\; e^{-\alpha_\infty u_j}\left[-\varrho R_2(\Delta_g u_j) + R_2(\varrho\Delta_g u_j)\right],
  \quad\text{(0th and 1st order cancellation)} \nonumber\\
  \abs{\left(\bm{J}_5(\theta_5)w_\varrho\right)_j}
  &\;\le\; \frac{\varrho + \varrho^2}{2}\,(\Delta_g u_J)^2.
  \quad\text{(since $e^{-\alpha_\infty u_j} \le 1$ and $u_j \le u_J$)}
  \label{eq:app-J5-trial}
\end{align}
By the variational Rayleigh characterization of the smallest and largest singular values of $\bm{J}_5(\theta_5)$,~\eqref{eq:app-J5-trial} holds for arbitrary $(A_0, A_g, A_f)$ and gives
\begin{align}
  \sigma_{\min}\!\left(\bm{J}_5(\theta_5)\right)
  &\;\le\; \frac{\norm{\bm{J}_5(\theta_5)w_\varrho}_2}{\norm{w_\varrho}_2}
  \;\le\; C_5\,(\Delta_g u_J)^2, \qquad
  C_5 \;\Let\; \frac{\sqrt{J}\,\varrho(\varrho + 1)}{2\sqrt{(\varrho - 1)^2 + \varrho^2 + 1}}, \nonumber\\
  \sigma_{\max}\!\left(\bm{J}_5(\theta_5)\right)
  &\;\ge\; \norm{\bm{J}_5(\theta_5)e_1}_2 \;=\; \sqrt{J}
  \quad\Longrightarrow\quad
  \kappa_2\!\left(\bm{J}_5(\theta_5)\right) \;=\; \frac{\sigma_{\max}}{\sigma_{\min}} \;\ge\; \frac{\sqrt{J}}{C_5}(\Delta_g u_J)^{-2},
\end{align}
which establishes~\eqref{eq:J5-conditioning-blowup} and completes the proof of \Cref{prop:identifiability}.
\end{proof}

\subsection{Proof of Corollary~\ref{cor:aitken-identifiability} (Aitken Difference Identification of Shifted Power Laws)}
\label{app:proof:cor:aitken-identifiability}

\begin{proof}[Proof of \Cref{cor:aitken-identifiability} \StatementRef{cor:aitken-identifiability}]
Suppose a linear combination of the columns of $\bm{J}_3(\theta_3)$ vanishes on $J \ge 3$ distinct points $u_1 < u_2 < u_3$. Multiplying the $j$-th component by $e^{\alpha_{\mathrm{eff}}u_j} > 0$ implies that the smooth function $g(u)$ defined below vanishes at $u_1 < u_2 < u_3$, so applying Rolle's theorem twice produces zeros $\xi_1 \in (u_1, u_2)$, $\xi_2 \in (u_2, u_3)$ of $g'$ and a zero $\eta \in (\xi_1, \xi_2)$ of $g''$:
\begin{align}
  g(u) &\;\Let\; c_1 e^{\alpha_{\mathrm{eff}}u} + c_2 - c_3 A_0 u
  \;=\; 0 \quad\text{at } u \in \{u_1, u_2, u_3\}, \nonumber\\
  g'(u) &\;=\; c_1\alpha_{\mathrm{eff}}e^{\alpha_{\mathrm{eff}}u} - c_3 A_0
  \;=\; 0 \quad\text{at } \xi_1 \in (u_1, u_2),\; \xi_2 \in (u_2, u_3), \nonumber\\
  g''(\eta) &\;=\; c_1\alpha_{\mathrm{eff}}^2 e^{\alpha_{\mathrm{eff}}\eta}
  \;=\; 0 \quad\text{at some } \eta \in (\xi_1, \xi_2)
  \quad\Longrightarrow\quad c_1 = c_3 = c_2 = 0,
\end{align}
proving $\operatorname{rank}(\bm{J}_3(\theta_3)) = 3$. Finally, on a uniform grid $u_j = (j-1)\Delta u$, consecutive first-order differences eliminate $E_\infty$ and satisfy
\begin{align}
  \Delta y_j
  &\;=\; f_3(u_{j+1}; \theta_3) - f_3(u_j; \theta_3)
  \;=\; \left(E_\infty + A_0 e^{-\alpha_{\mathrm{eff}}(u_j + \Delta u)}\right) - \left(E_\infty + A_0 e^{-\alpha_{\mathrm{eff}}u_j}\right) \nonumber\\
  &\;=\; -A_0\bigl(1 - e^{-\alpha_{\mathrm{eff}}\Delta u}\bigr)e^{-\alpha_{\mathrm{eff}}u_j}, \qquad
  \frac{\Delta y_{j+1}}{\Delta y_j} \;=\; e^{-\alpha_{\mathrm{eff}}\Delta u},
\end{align}
which solves uniquely for $(\alpha_{\mathrm{eff}}, A_0, E_\infty)$ as in~\eqref{eq:aitken-locking}.
\end{proof}

\section{Proofs for Section~\ref{sec:mixing}: Multi-Domain Data Mixing and Sequential Allocation}
\label{app:mixing}

In this appendix, we prove a supporting Taylor remainder lemma on the simplex (\Cref{lem:app-simplex-sandwich}) and then provide the complete proofs of \Cref{prop:water-filling}, \Cref{cor:powered-share-ratio}, \Cref{thm:master-mixing}, \Cref{prop:curriculum-rn}, \Cref{prop:simplex-hessian}, and \Cref{thm:fisher-transfer}.

\subsection{Supporting Taylor Remainder Lemma for Bottleneck Value Expansion}
\label{app:lemmas:mixing}

\begin{lemma}[Taylor Remainder Bound for Bottleneck Value]
\label{lem:app-simplex-sandwich}
For any $b_0 > 0$ and any $\varepsilon \in [0, 1/2]$, the factor $(1 - \varepsilon)^{-(b_0 + 1)}$ satisfies
\begin{equation}
  (1 - \varepsilon)^{-(b_0 + 1)} \;=\; 1 + (b_0 + 1)\varepsilon + R_b(\varepsilon),
  \qquad 0 \;\le\; R_b(\varepsilon) \;\le\; \frac{1}{2}(b_0 + 1)(b_0 + 2)2^{b_0 + 3}\varepsilon^2.
  \label{eq:app-Rb-bound}
\end{equation}
Consequently, in \Cref{thm:master-mixing}, the remainder $R_{\mathrm{mix}}(D)$ in~\eqref{eq:mix-value-expansion} satisfies $0 \le R_{\mathrm{mix}}(D) \le C_{\mathrm{mix}}(b_0)\varepsilon(D)^2$ for all $\varepsilon(D) \in [0, 1/2]$ with the explicit constant $C_{\mathrm{mix}}(b_0) = (b_0 + 1)(b_0 + 2)2^{b_0 + 2}$.
\end{lemma}
\begin{proof}
Define $\psi(\varepsilon) \Let (1 - \varepsilon)^{-(b_0 + 1)}$ on $[0, 1/2]$, so that $\psi(0) = 1$ and $\psi'(0) = b_0 + 1$. Applying Taylor's theorem with integral remainder on $t \in [0, \varepsilon] \subseteq [0, 1/2]$ (where $(1 - t)^{-1} \le 2$) gives
\begin{align}
  \psi''(t) &\;=\; (b_0 + 1)(b_0 + 2)(1 - t)^{-(b_0 + 3)} \;>\; 0, \nonumber\\
  0 \;\le\; R_b(\varepsilon)
  &\;=\; \int_0^{\varepsilon} (\varepsilon - t)\,\psi''(t)\dd t
  \;\le\; \frac{1}{2}(b_0 + 1)(b_0 + 2)2^{b_0 + 3}\varepsilon^2,
\end{align}
establishing~\eqref{eq:app-Rb-bound}. Next, define the weighted fast-domain sum $\Sigma_1(D)$ and expand the product in~\eqref{eq:mix-value-expansion}:
\begin{align}
  \Sigma_1(D) &\;\Let\; \sum_{k \in S^c} \left(1 - \frac{b_0}{\beta_k}\right)w_k\opt(D) \;\in\; \bigl[0, \varepsilon(D)\bigr], \nonumber\\
  \left(1 - \varepsilon(D)\right)^{-(b_0 + 1)}\bigl(1 - \Sigma_1(D)\bigr)
  &\;=\; \bigl(1 + (b_0 + 1)\varepsilon(D) + R_b(\varepsilon(D))\bigr)\bigl(1 - \Sigma_1(D)\bigr) \nonumber\\
  &\;=\; 1 + (b_0 + 1)\varepsilon(D) - \Sigma_1(D) + R_{\mathrm{mix}}(D) \nonumber\\
  &\;=\; 1 + \sum_{k \in S^c} \frac{b_0(\beta_k + 1)}{\beta_k}w_k\opt(D) + R_{\mathrm{mix}}(D), \nonumber\\
  R_{\mathrm{mix}}(D)
  &\;\Let\; R_b\bigl(\varepsilon(D)\bigr)\bigl(1 - \Sigma_1(D)\bigr) - (b_0 + 1)\varepsilon(D)\Sigma_1(D).
\end{align}
Moreover, regrouping the product via $1 - \Sigma_1(D) = 1 - \varepsilon(D) + \sum_{k \in S^c}\frac{b_0}{\beta_k}w_k\opt(D)$ yields the exact two-sided bound on $R_{\mathrm{mix}}(D)$:
\begin{align}
  R_{\mathrm{mix}}(D)
  &\;=\; \underbrace{\Bigl[\bigl(1 - \varepsilon(D)\bigr)^{-b_0} - 1 - b_0\varepsilon(D)\Bigr]}_{\ge\, 0\text{ by Bernoulli's inequality}}
  + \underbrace{\Bigl[\bigl(1 - \varepsilon(D)\bigr)^{-(b_0 + 1)} - 1\Bigr]}_{\ge\, 0}\sum_{k \in S^c}\frac{b_0}{\beta_k}w_k\opt(D) \;\ge\; 0, \nonumber\\
  R_{\mathrm{mix}}(D)
  &\;=\; R_b\bigl(\varepsilon(D)\bigr)\bigl(1 - \Sigma_1(D)\bigr) - (b_0 + 1)\varepsilon(D)\Sigma_1(D)
  \;\le\; R_b\bigl(\varepsilon(D)\bigr)
  \;\le\; C_{\mathrm{mix}}(b_0)\varepsilon(D)^2,
\end{align}
completing the proof of \Cref{lem:app-simplex-sandwich}.
\end{proof}

\subsection{Proof of Proposition~\ref{prop:water-filling} (KKT Water-Filling Characterization)}
\label{app:proof:prop:water-filling}

\begin{proof}[Proof of \Cref{prop:water-filling} \StatementRef{prop:water-filling}]
Fix any finite token budget $D > 0$, and evaluate the objective at the uniform reference mixture $\bar{w} \Let (1/K, \dots, 1/K)^{\top} \in \Delta_{K-1}^{\circ}$ to define the sublevel set $\mc{S}_D$:
\begin{align}
  \bar{E}(D) &\;\Let\; \mc{E}_{\mathrm{mix}}(\bar{w}; D) \;=\; \sum_{j=1}^K c_j\left(\frac{D}{K}\right)^{-\beta_j} \;<\; \infty, \nonumber\\
  \mc{S}_D &\;\Let\; \left\{w \in \Delta_{K-1}^{\circ} : \mc{E}_{\mathrm{mix}}(w; D) \le \bar{E}(D)\right\}.
\end{align}
For any $w \in \mc{S}_D$, positivity of each domain risk implies that every coordinate $w_k$ is bounded strictly away from zero:
\begin{equation}
  c_k(w_k D)^{-\beta_k} \;\le\; \mc{E}_{\mathrm{mix}}(w; D) \;\le\; \bar{E}(D)
  \quad\Longrightarrow\quad
  w_k \;\ge\; \underline{\delta}_k(D) \;\Let\; \frac{1}{D}\left(\frac{c_k}{\bar{E}(D)}\right)^{\frac{1}{\beta_k}} \;>\; 0.
  \label{eq:app-interior-barrier}
\end{equation}
Thus $\mc{S}_D$ is a non-empty, closed, and bounded subset of $\R^K$ contained strictly inside $\Delta_{K-1}^{\circ}$. By the Weierstrass extreme value theorem, $\mc{E}_{\mathrm{mix}}(\cdot; D)$ attains a global minimum $w\opt(D) \in \Delta_{K-1}^{\circ}$, which is unique by strict convexity of the diagonal Hessian on $\Delta_{K-1}^{\circ}$. Forming the Lagrangian $\mc{L}(w, \nu)$ on $\Delta_{K-1}^{\circ}$, the interior KKT conditions are
\begin{align}
  \mc{L}(w, \nu) &\;\Let\; \sum_{k=1}^K c_k D^{-\beta_k}w_k^{-\beta_k} + \nu\left(\sum_{k=1}^K w_k - 1\right), \nonumber\\
  \frac{\partial\mc{L}}{\partial w_k}\Big|_{w\opt(D)}
  &\;=\; -c_k\beta_k D^{-\beta_k}w_k\opt(D)^{-(\beta_k + 1)} + \nu \;=\; 0
  \quad\Longleftrightarrow\quad
  w_k\opt(D) \;=\; \left(\frac{c_k\beta_k D^{-\beta_k}}{\nu}\right)^{\frac{1}{\beta_k + 1}}.
  \label{eq:app-kkt-wk}
\end{align}
Summing~\eqref{eq:app-kkt-wk} over $k = 1, \dots, K$ yields the scalar water-filling equation $\Phi_D(\nu) = 1$, where
\begin{equation}
  \Phi_D(\nu) \;\Let\; \sum_{k=1}^K \left(\frac{c_k\beta_k D^{-\beta_k}}{\nu}\right)^{\frac{1}{\beta_k + 1}},
  \qquad
  \lim_{\nu \downarrow 0^+}\Phi_D(\nu) \;=\; +\infty,
  \qquad
  \lim_{\nu \to \infty}\Phi_D(\nu) \;=\; 0.
\end{equation}
Since $\Phi_D(\nu)$ is continuous and strictly decreasing on $(0, \infty)$, the intermediate value theorem guarantees a unique root $\nu(D) > 0$, establishing~\eqref{eq:water-filling-kkt}.
\end{proof}

\subsection{Proof of Corollary~\ref{cor:powered-share-ratio} (Invariant Powered Share-Ratio Law)}
\label{app:proof:cor:powered-share-ratio}

\begin{proof}[Proof of \Cref{cor:powered-share-ratio} \StatementRef{cor:powered-share-ratio}]
From the KKT optimality condition~\eqref{eq:app-kkt-wk} in the proof of \Cref{prop:water-filling}, for any domain $k \in \{1, \dots, K\}$ with $\beta_k \neq \beta_1$ and reference domain $1$, equating the common Lagrange multiplier $\nu(D)$ gives
\begin{align}
  c_k\beta_k D^{-\beta_k}w_k\opt(D)^{-(\beta_k + 1)}
  &\;=\; \nu(D)
  \;=\; c_1\beta_1 D^{-\beta_1}w_1\opt(D)^{-(\beta_1 + 1)} \nonumber\\
  \Longrightarrow\quad
  \frac{w_k\opt(D)^{\beta_k + 1}}{w_1\opt(D)^{\beta_1 + 1}}
  &\;=\; \frac{c_k\beta_k}{c_1\beta_1}D^{-(\beta_k - \beta_1)}
  \;=\; \left(\frac{D}{D_k\opt}\right)^{-(\beta_k - \beta_1)}
  \;=\; \tilde{D}_k^{-(\beta_k - \beta_1)}.
\end{align}
Raising both sides to the power $\frac{1}{\beta_k - \beta_1}$ establishes $\Upsilon_k(D) = \tilde{D}_k^{-1}$ in~\eqref{eq:powered-ratio-law}.
\end{proof}

\subsection{Proof of Theorem~\ref{thm:master-mixing} (Finite-Budget Bottleneck Approximation and Tie-Aware Bounds)}
\label{app:proof:thm:master-mixing}

\begin{proof}[Proof of \Cref{thm:master-mixing} \StatementRef{thm:master-mixing}]
We prove~\eqref{eq:eps-sandwich}--\eqref{eq:mix-value-expansion} in two modular steps.

\paragraph{Step 1: Tie-aware bottleneck decomposition and sandwich bounds.}
For every slowest domain $s \in S$ ($\beta_s = b_0$), summing~\eqref{eq:app-kkt-wk} over $s \in S$ and using $\sum_{s \in S} w_s\opt(D) = 1 - \varepsilon(D)$ expresses both $\nu(D)$ and $w_s\opt(D)$ in terms of $1 - \varepsilon(D)$:
\begin{align}
  w_s\opt(D) &\;=\; \left(b_0 c_s\right)^{\frac{1}{b_0 + 1}}\bigl(D^{b_0}\nu(D)\bigr)^{-\frac{1}{b_0 + 1}}
  \;=\; \theta_s\bigl(1 - \varepsilon(D)\bigr) \qquad (s \in S), \nonumber\\
  1 - \varepsilon(D) &\;=\; H\,D^{-\frac{b_0}{b_0 + 1}}\nu(D)^{-\frac{1}{b_0 + 1}}
  \quad\Longleftrightarrow\quad
  \nu(D)^{-1} \;=\; H^{-(b_0 + 1)}D^{b_0}\bigl(1 - \varepsilon(D)\bigr)^{b_0 + 1}.
  \label{eq:app-nu-tie}
\end{align}
Substituting~\eqref{eq:app-nu-tie} into~\eqref{eq:app-kkt-wk} for any fast domain $k \in S^c$ ($\beta_k > b_0$, where $r_k = \frac{b_0 + 1}{\beta_k + 1} \in (0, 1)$) yields
\begin{align}
  w_k\opt(D)
  &\;=\; \left(\beta_k c_k D^{-\beta_k}\cdot H^{-(b_0 + 1)}D^{b_0}\bigl(1 - \varepsilon(D)\bigr)^{b_0 + 1}\right)^{\frac{1}{\beta_k + 1}} \nonumber\\
  &\;=\; \left(\beta_k c_k\right)^{\frac{1}{\beta_k + 1}}H^{-r_k}D^{-(1 - r_k)}\bigl(1 - \varepsilon(D)\bigr)^{r_k}
  \;=\; z_k(D)\bigl(1 - \varepsilon(D)\bigr)^{r_k}.
  \label{eq:wk-exact-zk}
\end{align}
Summing~\eqref{eq:wk-exact-zk} over $k \in S^c$ and applying Bernoulli's elementary bounds $1 - \varepsilon(D) \le (1 - \varepsilon(D))^{r_k} \le 1$ gives
\begin{align}
  \varepsilon(D)
  &\;=\; \sum_{k \in S^c} z_k(D)\bigl(1 - \varepsilon(D)\bigr)^{r_k}, \nonumber\\
  Z(D)\bigl(1 - \varepsilon(D)\bigr)
  &\;\le\; \varepsilon(D) \;\le\; Z(D)
  \quad\Longleftrightarrow\quad
  \frac{Z(D)}{1 + Z(D)} \;\le\; \varepsilon(D) \;\le\; \min\bigl\{Z(D), 1\bigr\},
\end{align}
establishing~\eqref{eq:eps-sandwich}.

\paragraph{Step 2: Plug-in $L_1$ error bound and optimal value expansion.}
For the plug-in estimator $\tilde{w}(D)$, checking simplex normalization and coordinatewise sign differences gives
\begin{align}
  \sum_{k=1}^K \tilde{w}_k(D)
  &\;=\; \bigl(1 - Z(D)\bigr)\sum_{s \in S}\theta_s + \sum_{k \in S^c} z_k(D) \;=\; 1, \nonumber\\
  \tilde{w}_k(D) - w_k\opt(D)
  &\;=\; z_k(D) - w_k\opt(D) \;\ge\; 0 \qquad (k \in S^c), \nonumber\\
  w_s\opt(D) - \tilde{w}_s(D)
  &\;=\; \theta_s\bigl(Z(D) - \varepsilon(D)\bigr) \;\ge\; 0 \qquad (s \in S), \nonumber\\
  \norm{\tilde{w}(D) - w\opt(D)}_1
  &\;=\; \sum_{k \in S^c}\bigl(z_k(D) - w_k\opt(D)\bigr) + \sum_{s \in S}\theta_s\bigl(Z(D) - \varepsilon(D)\bigr) \nonumber\\
  &\;=\; 2\bigl(Z(D) - \varepsilon(D)\bigr)
  \;\le\; \frac{2Z(D)^2}{1 + Z(D)},
\end{align}
proving~\eqref{eq:plugin-L1-bound}. Finally, using the KKT identity~\eqref{eq:app-kkt-wk} and~\eqref{eq:app-nu-tie} to express each domain's risk contribution in terms of $\nu(D)$ and $\mc{E}_{\mathrm{lead}}(D) = \frac{H^{b_0 + 1}}{b_0}D^{-b_0}$ gives
\begin{align}
  c_k D^{-\beta_k}w_k\opt(D)^{-\beta_k}
  &\;=\; \frac{\nu(D)}{\beta_k}w_k\opt(D), \qquad
  \frac{\nu(D)}{b_0}
  \;=\; \mc{E}_{\mathrm{lead}}(D)\bigl(1 - \varepsilon(D)\bigr)^{-(b_0 + 1)}, \nonumber\\
  \mc{E}_{\mathrm{mix}}\opt(D)
  &\;=\; \frac{\nu(D)}{b_0}\left[\sum_{s \in S} w_s\opt(D) + \sum_{k \in S^c}\frac{b_0}{\beta_k}w_k\opt(D)\right] \nonumber\\
  &\;=\; \mc{E}_{\mathrm{lead}}(D)\bigl(1 - \varepsilon(D)\bigr)^{-(b_0 + 1)}\left[1 + \sum_{k \in S^c}\left(\frac{b_0}{\beta_k} - 1\right)w_k\opt(D)\right],
\end{align}
and applying \Cref{lem:app-simplex-sandwich} yields the second line of~\eqref{eq:mix-value-expansion}.
\end{proof}

\subsection{Proof of Proposition~\ref{prop:curriculum-rn} (Sequential Allocation and Instantaneous Sampling)}
\label{app:proof:prop:curriculum-rn}

\begin{proof}[Proof of \Cref{prop:curriculum-rn} \StatementRef{prop:curriculum-rn}]
We prove~\eqref{eq:instantaneous-curriculum} and~\eqref{eq:radon-nikodym-tilt} in two steps.

\paragraph{Step 1: Logarithmic differentiation of cumulative domain allocations.}
Substituting $w_k\opt(D) = n_k\opt(D) / D$ into the KKT condition~\eqref{eq:water-filling-kkt} and defining the token-count multiplier $\lambda(D) \Let \nu(D) / D > 0$ expresses $n_k\opt(D)$ and the total budget constraint directly in terms of $\lambda(D)$:
\begin{gather}
  c_k\beta_k\left(n_k\opt(D)\right)^{-(\beta_k + 1)} \;=\; \frac{\nu(D)}{D} \;\Let\; \lambda(D), \qquad
  \sum_{k=1}^K \left(\frac{c_k\beta_k}{\lambda(D)}\right)^{\frac{1}{\beta_k + 1}} \;=\; D, \nonumber\\
  \ln n_k\opt(D) \;=\; \frac{1}{\beta_k + 1}\ln(c_k\beta_k) - \frac{1}{\beta_k + 1}\ln\lambda(D).
  \label{eq:app-lambda-count-kkt}
\end{gather}
Because the sum in~\eqref{eq:app-lambda-count-kkt} is smooth and strictly decreasing in $\lambda > 0$, the implicit function theorem implies that $\lambda(D)$ is $C^\infty$-smooth and strictly decreasing ($\lambda'(D) < 0$) on $(0, \infty)$. Differentiating $\ln n_k\opt(D)$ in~\eqref{eq:app-lambda-count-kkt} with respect to $D$ gives
\begin{equation}
  \frac{1}{n_k\opt(D)}\frac{\dd n_k\opt(D)}{\dd D} \;=\; -\frac{1}{\beta_k + 1}\frac{\lambda'(D)}{\lambda(D)}
  \quad\Longrightarrow\quad
  q_k\opt(D) \;\Let\; \frac{\dd n_k\opt(D)}{\dd D} \;=\; -\frac{D\lambda'(D)}{\lambda(D)}\,\frac{w_k\opt(D)}{\beta_k + 1} \;>\; 0.
  \label{eq:dnk-dD-raw}
\end{equation}
Summing~\eqref{eq:dnk-dD-raw} over $k = 1, \dots, K$ and integrating from $t = 0$ to $t = D$ yields
\begin{align}
  \sum_{k=1}^K q_k\opt(D)
  &\;=\; \frac{\dd}{\dd D}\sum_{k=1}^K n_k\opt(D) \;=\; 1
  \quad\Longrightarrow\quad
  -\frac{D\lambda'(D)}{\lambda(D)}
  \;=\; \left(\sum_{j=1}^K \frac{w_j\opt(D)}{\beta_j + 1}\right)^{-1}, \nonumber\\
  0 \;\le\; n_k\opt(t) &\;\le\; t \;\xrightarrow[t \downarrow 0^+]{}\; 0
  \quad\Longrightarrow\quad
  \int_0^D q_k\opt(t)\dd t
  \;=\; n_k\opt(D) - \lim_{t \downarrow 0^+} n_k\opt(t)
  \;=\; D w_k\opt(D),
\end{align}
establishing~\eqref{eq:instantaneous-curriculum}.

\paragraph{Step 2: Radon--Nikodym tilt between $\QQ_D\opt$ and $\PP_D\opt$.}
Dividing~\eqref{eq:instantaneous-curriculum} by $w_k\opt(D) > 0$ expresses the Radon--Nikodym derivative directly in terms of the expectation under $\PP_D\opt$:
\begin{equation}
  \frac{\dd \QQ_D\opt}{\dd \PP_D\opt}(k)
  \;=\; \frac{q_k\opt(D)}{w_k\opt(D)}
  \;=\; \frac{(\beta_k + 1)^{-1}}{\sum_{j=1}^K w_j\opt(D)(\beta_j + 1)^{-1}}
  \;=\; \frac{(\beta_k + 1)^{-1}}{\E_{\PP_D\opt}\!\left[(\beta_I + 1)^{-1}\right]},
\end{equation}
which yields~\eqref{eq:radon-nikodym-tilt}.
\end{proof}

\subsection{Proof of Proposition~\ref{prop:simplex-hessian} (Simplex Hessian Metric and Sequential Duality)}
\label{app:proof:prop:simplex-hessian}

\begin{proof}[Proof of \Cref{prop:simplex-hessian} \StatementRef{prop:simplex-hessian}]
Since $\mc{E}_{\mathrm{mix}}(w; D)$ is separable across coordinates $w_1, \dots, w_K$, all off-diagonal second partial derivatives vanish, while the $k$-th diagonal entry at $w = w\opt(D)$ simplifies via the KKT identity~\eqref{eq:water-filling-kkt}:
\begin{align}
  \frac{\partial^2\mc{E}_{\mathrm{mix}}(w\opt(D); D)}{\partial w_k^2}
  &\;=\; c_k\beta_k(\beta_k + 1)D^{-\beta_k}w_k\opt(D)^{-(\beta_k + 2)} \nonumber\\
  &\;=\; (\beta_k + 1)\underbrace{\left[c_k\beta_k D^{-\beta_k}w_k\opt(D)^{-(\beta_k + 1)}\right]}_{=\,\nu(D)\text{ by~\eqref{eq:water-filling-kkt}}}\frac{1}{w_k\opt(D)}
  \;=\; \nu(D)\frac{\beta_k + 1}{w_k\opt(D)}.
\end{align}
Inverting $\bm{D}_H(D)$ and comparing its diagonal entries with $q_k\opt(D)$ in~\eqref{eq:instantaneous-curriculum} yields the duality relation and minimum eigenvalue lower bound (since $w_k\opt(D) \le 1$ and $\beta_k \ge b_0$) in~\eqref{eq:simplex-hessian-metric}:
\begin{gather}
  \bigl(\bm{D}_H(D)^{-1}\bigr)_{kk}
  \;=\; \frac{w_k\opt(D)}{\nu(D)(\beta_k + 1)}
  \;=\; \frac{\E_{\PP_D\opt}\!\bigl[(\beta_I + 1)^{-1}\bigr]}{\nu(D)}\,q_k\opt(D), \nonumber\\
  \lambda_{\min}\!\bigl(\bm{D}_H(D)\bigr) \;\ge\; \nu(D)(b_0 + 1).
\end{gather}
Finally, for any tangent perturbation $\delta w \in T_{w\opt(D)}\Delta_{K-1}$ ($\sum_{k=1}^K \delta w_k = 0$) with $w\opt(D) + \delta w \in \Delta_{K-1}^{\circ}$, the first-order directional derivative vanishes by KKT stationarity, and at any intermediate point $w_k \ge \underline{w}_k$ along $w\opt(D) + t\,\delta w$ ($t \in [0, 1]$), substituting the KKT stationarity identity into the third partial derivative gives
\begin{align}
  \ip{\nabla\mc{E}_{\mathrm{mix}}(w\opt(D); D)}{\delta w}_{\R^K}
  &\;=\; -\nu(D)\sum_{k=1}^K \delta w_k \;=\; 0, \nonumber\\
  -\frac{\partial^3\mc{E}_{\mathrm{mix}}(w; D)}{\partial w_k^3}
  &\;=\; c_k\beta_k(\beta_k + 1)(\beta_k + 2)D^{-\beta_k}w_k^{-(\beta_k + 3)} \nonumber\\
  &\;=\; \nu(D)(\beta_k + 1)(\beta_k + 2)\frac{w_k\opt(D)^{\beta_k + 1}}{w_k^{\beta_k + 3}} \;>\; 0,
\end{align}
so Lagrange's third-order Taylor remainder directly yields~\eqref{eq:simplex-R3-bound}. On the relative trust neighborhood $\frac{1}{2}w_k\opt(D) \le \hat{w}_k \le 2w_k\opt(D)$, bounding each coordinate's second derivative from below along the segment $[w\opt(D), \hat{w}]$ yields the quadratic regret bound:
\begin{align}
  \frac{\partial^2\mc{E}_{\mathrm{mix}}(w; D)}{\partial w_k^2}
  &\;\ge\; \overline{\kappa}_w^{-1}\bigl(\bm{D}_H(D)\bigr)_{kk}, \qquad
  \overline{\kappa}_w \;\Let\; \max_{1 \le k \le K} 2^{\beta_k + 2}, \nonumber\\
  \mc{E}_{\mathrm{mix}}(\hat{w}; D) - \mc{E}_{\mathrm{mix}}\opt(D)
  &\;\le\; \frac{\overline{\kappa}_w}{2}\norm{\nabla\mc{E}_{\mathrm{mix}}(\hat{w}; D) + \nu(D)\mathbf{1}}_{\bm{D}_H(D)^{-1}}^2
  \;\le\; \frac{\overline{\kappa}_w}{2}\,\varepsilon_{\mathrm{mix}}^2,
\end{align}
completing the proof of \Cref{prop:simplex-hessian}.
\end{proof}

\subsection{Proof of Corollary~\ref{thm:fisher-transfer} (First-Order Perturbation Under Cross-Domain Coupling)}
\label{app:proof:thm:fisher-transfer}

\begin{proof}[Proof of \Cref{thm:fisher-transfer} \StatementRef{thm:fisher-transfer}]
Adding the quadratic coupling perturbation and expanding the coupled objective to second order around $w\opt(D)$ along a tangent step $\delta w \in T_{w\opt(D)}\Delta_{K-1}$ yields the linearly constrained quadratic program and its $(K + 1) \times (K + 1)$ KKT saddle-point block system:
\begin{equation}
  \min_{\mathbf{1}^{\top}\delta w = 0}\left\{\frac{1}{2}\delta w^{\top}\bm{D}_H(D)\delta w + (\bm{\Gamma}w\opt(D))^{\top}\delta w\right\}
  \;\iff\;
  \begin{bmatrix}
    \bm{D}_H(D) & \mathbf{1} \\
    \mathbf{1}^{\top} & 0
  \end{bmatrix}
  \begin{bmatrix}
    \delta w \\
    \delta\nu
  \end{bmatrix}
  =
  \begin{bmatrix}
    -\bm{\Gamma}w\opt(D) \\
    0
  \end{bmatrix}.
  \label{eq:app-kkt-saddle-system}
\end{equation}
Inverting $\bm{D}_H(D) \succ 0$ on the first block row of~\eqref{eq:app-kkt-saddle-system} and enforcing $\mathbf{1}^{\top}\delta w = 0$ solves uniquely for the dual shift $\delta\nu$, the perturbation step $\delta w = \Delta w_{\mathrm{nat}}(D)$, and the $g_{w\opt(D)}$-orthogonal projection matrix $\bm{\Pi}_{g_{w\opt}}$ onto $T_{w\opt(D)}\Delta_{K-1}$:
\begin{align}
  \delta w &\;=\; -\bm{D}_H(D)^{-1}\bigl(\bm{\Gamma}w\opt(D) + \delta\nu\,\mathbf{1}\bigr), \qquad
  \delta\nu
  \;=\; -\frac{\mathbf{1}^{\top}\bm{D}_H(D)^{-1}\bm{\Gamma}w\opt(D)}{\mathbf{1}^{\top}\bm{D}_H(D)^{-1}\mathbf{1}}, \nonumber\\
  \Delta w_{\mathrm{nat}}(D)
  &\;=\; -\bm{\Pi}_{g_{w\opt}}\bm{D}_H(D)^{-1}\bm{\Gamma}w\opt(D), \qquad
  \bm{\Pi}_{g_{w\opt}}
  \;\Let\; \bm{I}_K - \frac{\bm{D}_H(D)^{-1}\mathbf{1}\mathbf{1}^{\top}}{\mathbf{1}^{\top}\bm{D}_H(D)^{-1}\mathbf{1}}.
\end{align}
Substituting the Hessian--sampling duality $\bm{D}_H(D)^{-1} = \frac{\E_{\PP_D\opt}[(\beta_I + 1)^{-1}]}{\nu(D)}\,\diag(q\opt(D))$ from~\eqref{eq:simplex-hessian-metric} into the projection term gives
\begin{equation}
  \frac{\mathbf{1}^{\top}\bm{D}_H(D)^{-1}\bm{\Gamma}w\opt(D)}{\mathbf{1}^{\top}\bm{D}_H(D)^{-1}\mathbf{1}}
  \;=\; \sum_{j=1}^K q_j\opt(D)\bigl(\bm{\Gamma}w\opt(D)\bigr)_j
  \;=\; \E_{\QQ_D\opt}\!\Bigl[\bigl(\bm{\Gamma}w\opt(D)\bigr)_I\Bigr],
\end{equation}
which establishes both lines of~\eqref{eq:natural-grad-component}. Finally, for each fixed finite budget $D > 0$ (so that $w\opt(D) \in \Delta_{K-1}^{\circ}$ is bounded strictly away from $\partial\Delta_{K-1}$ by~\eqref{eq:app-interior-barrier}), the block KKT Jacobian in~\eqref{eq:app-kkt-saddle-system} is non-singular at $\bm{\Gamma} = \bm{0}$; hence the quantitative implicit function theorem combined with the third-order remainder bound~\eqref{eq:simplex-R3-bound} guarantees the quadratic remainder bound in~\eqref{eq:natural-grad-shift} on a neighborhood of $\bm{\Gamma} = \bm{0}$.
\end{proof}

\section{Proofs for Section~\ref{sec:compute}: Compute-Optimal Allocation and Exponent Drift}
\label{app:compute}

In this appendix, we first prove a supporting lemma connecting multi-power exponential sums to the cumulant generating function of the marginal-gradient Gibbs measure (\Cref{lem:app-gibbs-cumulants}), and then provide the complete proofs of \Cref{thm:invariance-counterexample}, \Cref{prop:curvature-ode}, \Cref{thm:master-drift}, \Cref{prop:water-filling-drift}, \Cref{cor:logistic-transition}, \Cref{thm:master-envelope}, \Cref{cor:allocation-regret}, and \Cref{prop:surrogate-decision-regret}.

\subsection{Supporting Cumulant Lemma for Multi-Power Gibbs Measures}
\label{app:lemmas:compute}

\begin{lemma}[Cumulant Identities for Multi-Power Potentials]
\label{lem:app-gibbs-cumulants}
Let $U(u) \Let \sum_{i=1}^I A_i e^{-\alpha_i u}$ with $A_i > 0$ and $\alpha_i > 0$ for all $i \in \{1, \dots, I\}$. For any $u \in \R$, define the marginal-gradient partition function $Z_U(u)$, the log-partition function $\Lambda_U(u)$, and the Gibbs probability measure $\PP_u$ on $\{1, \dots, I\}$ by
\begin{equation}
  Z_U(u) \;\Let\; -U'(u) \;=\; \sum_{i=1}^I A_i \alpha_i e^{-\alpha_i u} \;>\; 0, \quad
  \Lambda_U(u) \;\Let\; \ln Z_U(u), \quad
  \PP_u(i) \;\Let\; \frac{A_i \alpha_i e^{-\alpha_i u}}{Z_U(u)}.
\end{equation}
Then:
\begin{align}
  -\Lambda_U'(u) &\;=\; \frac{U''(u)}{-U'(u)} \;=\; \E_{\PP_u}[\alpha], \label{eq:app-gibbs-mean} \\
  \Lambda_U''(u) &\;=\; -\frac{\dd}{\dd u}\E_{\PP_u}[\alpha] \;=\; \Var_{\PP_u}(\alpha), \label{eq:app-gibbs-var} \\
  -\frac{U'''(u)}{U''(u)} &\;=\; \frac{\E_{\PP_u}[\alpha^2]}{\E_{\PP_u}[\alpha]} \;=\; \E_{\PP_u}[\alpha] + \frac{\Var_{\PP_u}(\alpha)}{\E_{\PP_u}[\alpha]}. \label{eq:app-gibbs-third-second}
\end{align}
\end{lemma}
\begin{proof}
Differentiating $Z_U(u)$ once and twice with respect to $u$ relates its derivatives to the first two raw moments under $\PP_u$:
\begin{align}
  -Z_U'(u) &\;=\; U''(u) \;=\; \sum_{i=1}^I A_i \alpha_i^2 e^{-\alpha_i u} \;=\; Z_U(u)\,\E_{\PP_u}[\alpha], \nonumber\\
  Z_U''(u) &\;=\; -U'''(u) \;=\; \sum_{i=1}^I A_i \alpha_i^3 e^{-\alpha_i u} \;=\; Z_U(u)\,\E_{\PP_u}[\alpha^2].
\end{align}
Dividing the first relation by $Z_U(u) = -U'(u)$ proves~\eqref{eq:app-gibbs-mean}, while differentiating $-\Lambda_U'(u)$ via the quotient rule and dividing $Z_U''(u)$ by $U''(u)$ yield
\begin{align}
  \Lambda_U''(u)
  &\;=\; -\frac{\dd}{\dd u}\E_{\PP_u}[\alpha]
  \;=\; \frac{-U'''(u)\bigl(-U'(u)\bigr) - \bigl(U''(u)\bigr)^2}{\bigl(-U'(u)\bigr)^2} \nonumber\\
  &\;=\; \E_{\PP_u}[\alpha^2] - \bigl(\E_{\PP_u}[\alpha]\bigr)^2
  \;=\; \Var_{\PP_u}(\alpha), \nonumber\\
  -\frac{U'''(u)}{U''(u)}
  &\;=\; \frac{Z_U(u)\,\E_{\PP_u}[\alpha^2]}{Z_U(u)\,\E_{\PP_u}[\alpha]}
  \;=\; \frac{\bigl(\E_{\PP_u}[\alpha]\bigr)^2 + \Var_{\PP_u}(\alpha)}{\E_{\PP_u}[\alpha]}
  \;=\; \E_{\PP_u}[\alpha] + \frac{\Var_{\PP_u}(\alpha)}{\E_{\PP_u}[\alpha]},
\end{align}
establishing~\eqref{eq:app-gibbs-var} and~\eqref{eq:app-gibbs-third-second}.
\end{proof}

\subsection{Proof of Proposition~\ref{thm:invariance-counterexample} (Scale Invariance Under Monotone Loss Transformations)}
\label{app:proof:thm:invariance-counterexample}

\begin{proof}[Proof of \Cref{thm:invariance-counterexample} \StatementRef{thm:invariance-counterexample}]
In log-resource coordinates $u = \ln N$ and $v = \ln D$, parameterize the affine constraint line $m u + n v = \ln(C / \kappa_c)$ by $u \in \R$ and restrict the inner bottleneck potential $\Phi(e^u, e^v)$ to this line:
\begin{align}
  v(u) &\;\Let\; \frac{1}{n}\ln\!\left(\frac{C}{\kappa_c}\right) - \frac{m}{n}u, \qquad
  \tilde{\Phi}(u) \;\Let\; c_1 e^{-au} + c_2 e^{-bv(u)}, \nonumber\\
  \frac{\dd}{\dd u}F\!\left(\tilde{\Phi}(u)\right)
  &\;=\; F'\!\left(\tilde{\Phi}(u)\right)\tilde{\Phi}'(u)
  \;=\; F'\!\left(\tilde{\Phi}(u)\right)\left[-a c_1 e^{-au} + \frac{mb}{n}c_2 e^{-bv(u)}\right], \nonumber\\
  \tilde{\Phi}''(u)
  &\;=\; a^2 c_1 e^{-au} + \left(\frac{mb}{n}\right)^2 c_2 e^{-bv(u)} \;>\; 0.
  \label{eq:app-monotone-chain}
\end{align}
Because $F'(z) > 0$ everywhere on $(0, \infty)$ while $\tilde{\Phi}''(u) > 0$, $F(\tilde{\Phi}(u))$ is strictly quasiconvex with a unique global minimizer determined by setting the bracketed factor in~\eqref{eq:app-monotone-chain} to zero (canceling $F'(\tilde{\Phi}(u)) > 0$ identically):
\begin{align}
  a c_1 e^{-au\opt} \;=\; \frac{mb}{n}c_2 e^{-bv\opt}
  &\quad\Longleftrightarrow\quad
  a u\opt - b v\opt \;=\; \ln\!\left(\frac{n a c_1}{m b c_2}\right).
\end{align}
Solving this $2 \times 2$ linear system jointly with $m u\opt + n v\opt = \ln(C / \kappa_c)$ gives
\begin{align}
  u\opt(C) &\;=\; \frac{b}{mb + na}\ln\!\left(\frac{C}{\kappa_c}\right) + \frac{n}{mb + na}\ln\!\left(\frac{n a c_1}{m b c_2}\right), \nonumber\\
  v\opt(C) &\;=\; \frac{a}{mb + na}\ln\!\left(\frac{C}{\kappa_c}\right) - \frac{m}{mb + na}\ln\!\left(\frac{n a c_1}{m b c_2}\right), \nonumber\\
  A_N &\;=\; \kappa_c^{-\eta_N}\!\left(\frac{nac_1}{mbc_2}\right)^{\frac{n}{mb+na}}, \qquad
  A_D \;=\; \kappa_c^{-\eta_D}\!\left(\frac{mbc_2}{nac_1}\right)^{\frac{m}{mb+na}},
\end{align}
which exponentiates to~\eqref{eq:invariant-exponents}.
\end{proof}

\subsection{Proof of Proposition~\ref{prop:curvature-ode} (Iso-Compute Curvature Differential Equation)}
\label{app:proof:prop:curvature-ode}

\begin{proof}[Proof of \Cref{prop:curvature-ode} \StatementRef{prop:curvature-ode}]
For each log-compute budget $t \in \R$, the unique minimizer $(u\opt(t), v\opt(t)) \in \mc{M}_t$ satisfies $v\opt(t) = t - u\opt(t)$ and the first-order tangency equation $G(u\opt(t), t) = 0$, where $G(u, t)$ and its partial derivative with respect to $u$ are given by
\begin{align}
  G(u, t) &\;\Let\; \mc{E}_u(u, t - u) - \mc{E}_v(u, t - u), \nonumber\\
  \frac{\partial G(u, t)}{\partial u}
  &\;=\; \mc{E}_{uu}(u, t - u) - 2\mc{E}_{uv}(u, t - u) + \mc{E}_{vv}(u, t - u)
  \;=\; \mc{H}_{\mathrm{iso}}(u, t - u) \;>\; 0,
  \label{eq:app-ift-jacobian}
\end{align}
which is strictly non-zero everywhere by~\eqref{eq:iso-compute-convexity}. By the implicit function theorem, $t \mapsto u\opt(t)$ is $C^2(\R)$. Setting $A(t) \Let (\mc{E}_{vv} - \mc{E}_{uv})|_{(u\opt(t), v\opt(t))}$ and $B(t) \Let \mc{H}_{\mathrm{iso}}(u\opt(t), v\opt(t)) > 0$ so that $B(t) - A(t) = (\mc{E}_{uu} - \mc{E}_{uv})|_{(u\opt(t), v\opt(t))}$, the derivative $\eta(t) = \frac{\dd u\opt(t)}{\dd t}$ is given by
\begin{equation}
  \eta(t)
  \;=\; -\frac{\partial G / \partial t}{\partial G / \partial u}\Bigg|_{u = u\opt(t)}
  \;=\; \frac{\mc{E}_{vv} - \mc{E}_{uv}}{\mc{E}_{uu} - 2\mc{E}_{uv} + \mc{E}_{vv}}\Bigg|_{(u\opt(t), v\opt(t))}
  \;=\; \frac{A(t)}{B(t)},
\end{equation}
establishing~\eqref{eq:eta-general-hessian}. Because the tangent vector of $(u\opt(t), v\opt(t))$ is $(\eta(t), 1 - \eta(t))$, the chain rule along the optimal trajectory acts via $\mc{D} = \eta(t)\partial_u + (1 - \eta(t))\partial_v$. Differentiating $\eta(t) = A(t)/B(t)$ via the quotient rule gives
\begin{align}
  \eta'(t)
  &\;=\; \frac{B(t)\,\mc{D}A(t) - A(t)\,\mc{D}B(t)}{B(t)^2}
  \;=\; \eta(t)\bigl(1 - \eta(t)\bigr)\mc{K}(t), \nonumber\\
  \mc{K}(t)
  &\;=\; \frac{\dd}{\dd t}\ln\!\left(\frac{\eta(t)}{1 - \eta(t)}\right)
  \;=\; \frac{\dd}{\dd t}\ln\!\left(\frac{A(t)}{B(t) - A(t)}\right)
  \;=\; \frac{\mc{D}A(t)}{A(t)} - \frac{\mc{D}\bigl(B(t) - A(t)\bigr)}{B(t) - A(t)},
  \label{eq:app-eta-prime-quotient}
\end{align}
establishing~\eqref{eq:eta-prime-general}.
\end{proof}

\subsection{Proof of Theorem~\ref{thm:master-drift} (Exponent-Variance Identity for Compute-Optimal Drift)}
\label{app:proof:thm:master-drift}

\begin{proof}[Proof of \Cref{thm:master-drift} \StatementRef{thm:master-drift}]
When $\mc{E}(u, v) = U(u) + V(v)$ is separable ($\mc{E}_{uv} \equiv 0$), we have $(\mc{E}_{vv} - \mc{E}_{uv})|_{(u\opt(t), v\opt(t))} = V''(v\opt(t))$ and $(\mc{E}_{uu} - \mc{E}_{uv})|_{(u\opt(t), v\opt(t))} = U''(u\opt(t))$ in \Cref{prop:curvature-ode}~\eqref{eq:eta-general-hessian}--\eqref{eq:eta-prime-general}. Differentiating the log-odds ratio with respect to $t$ yields~\eqref{eq:K-third-derivative}:
\begin{align}
  \ln\!\left(\frac{\eta(t)}{1 - \eta(t)}\right)
  &\;=\; \ln V''(v\opt(t)) - \ln U''(u\opt(t)), \nonumber\\
  \mc{K}(t)
  &\;=\; \frac{\dd}{\dd t}\Bigl[\ln V''(v\opt(t)) - \ln U''(u\opt(t))\Bigr] \nonumber\\
  &\;=\; \bigl(1 - \eta(t)\bigr)\frac{V'''(v\opt(t))}{V''(v\opt(t))} - \eta(t)\frac{U'''(u\opt(t))}{U''(u\opt(t))} \nonumber\\
  &\;=\; \eta(t)\!\left(-\frac{U'''(u\opt(t))}{U''(u\opt(t))}\right) - \bigl(1 - \eta(t)\bigr)\!\left(-\frac{V'''(v\opt(t))}{V''(v\opt(t))}\right).
  \label{eq:app-K-separable}
\end{align}
Now suppose $U(u) = \sum_{i=1}^I A_i e^{-\alpha_i u}$ and $V(v) = \sum_{j=1}^K B_j e^{-\beta_j v}$. Along the optimal trajectory, first-order stationarity $-U'(u\opt(t)) = -V'(v\opt(t)) > 0$ combined with \Cref{lem:app-gibbs-cumulants}~\eqref{eq:app-gibbs-mean} yields the exact first-moment balance:
\begin{align}
  U''(u\opt(t)) &\;=\; \bigl(-U'(u\opt(t))\bigr)\E_{\PP_t}[\alpha], \qquad
  V''(v\opt(t)) \;=\; \bigl(-V'(v\opt(t))\bigr)\E_{\QQ_t}[\beta], \nonumber\\
  \frac{1 - \eta(t)}{\eta(t)}
  &\;=\; \frac{U''(u\opt(t))}{V''(v\opt(t))}
  \;=\; \frac{\E_{\PP_t}[\alpha]}{\E_{\QQ_t}[\beta]}
  \quad\Longleftrightarrow\quad
  \eta(t)\,\E_{\PP_t}[\alpha] - \bigl(1 - \eta(t)\bigr)\E_{\QQ_t}[\beta] \;=\; 0.
  \label{eq:app-first-moment-balance}
\end{align}
Furthermore, by \Cref{lem:app-gibbs-cumulants}~\eqref{eq:app-gibbs-third-second}, the third-to-second derivative ratios satisfy
\begin{equation}
  -\frac{U'''(u\opt(t))}{U''(u\opt(t))} \;=\; \E_{\PP_t}[\alpha] + \frac{\Var_{\PP_t}(\alpha)}{\E_{\PP_t}[\alpha]}, \qquad
  -\frac{V'''(v\opt(t))}{V''(v\opt(t))} \;=\; \E_{\QQ_t}[\beta] + \frac{\Var_{\QQ_t}(\beta)}{\E_{\QQ_t}[\beta]}.
  \label{eq:cumulant-ratio-U}
\end{equation}
Substituting~\eqref{eq:cumulant-ratio-U} into~\eqref{eq:app-K-separable} and canceling the first-moment terms via~\eqref{eq:app-first-moment-balance} gives
\begin{align}
  \mc{K}(t)
  &\;=\; \eta(t)\!\left(\E_{\PP_t}[\alpha] + \frac{\Var_{\PP_t}(\alpha)}{\E_{\PP_t}[\alpha]}\right)
  - \bigl(1 - \eta(t)\bigr)\!\left(\E_{\QQ_t}[\beta] + \frac{\Var_{\QQ_t}(\beta)}{\E_{\QQ_t}[\beta]}\right) \nonumber\\
  &\;=\; \underbrace{\Bigl(\eta(t)\,\E_{\PP_t}[\alpha] - \bigl(1 - \eta(t)\bigr)\E_{\QQ_t}[\beta]\Bigr)}_{=\,0\text{ by~\eqref{eq:app-first-moment-balance}}}
  + \frac{\eta(t)\,\Var_{\PP_t}(\alpha)}{\E_{\PP_t}[\alpha]} - \frac{\bigl(1 - \eta(t)\bigr)\,\Var_{\QQ_t}(\beta)}{\E_{\QQ_t}[\beta]},
\end{align}
which establishes~\eqref{eq:gibbs-variance-identity} and completes the proof of \Cref{thm:master-drift}.
\end{proof}

\subsection{Proof of Proposition~\ref{prop:water-filling-drift} (Water-Filling Envelope Curvature and Inverse-Harmonic Drift)}
\label{app:proof:prop:water-filling-drift}

\begin{proof}[Proof of \Cref{prop:water-filling-drift} \StatementRef{prop:water-filling-drift}]
When $V(v) = \mc{E}_{\mathrm{mix}}\opt(e^v)$ is the KKT water-filling envelope of \Cref{prop:water-filling} at $D = e^v$, the envelope theorem and the KKT identity~\eqref{eq:water-filling-kkt} give $-V'(v) = \nu(e^v)$. Expressing $\lambda(D) = \nu(D)/D$ and $m(v) = \E_{\PP_D\opt}[\rho_I]$ ($\rho_k = (\beta_k+1)^{-1}$) via~\eqref{eq:dnk-dD-raw} and differentiating logarithmically with respect to $v = \ln D$ yields
\begin{align}
  -\frac{D\lambda'(D)}{\lambda(D)}
  &\;=\; \frac{1}{m(v)}
  \quad\Longrightarrow\quad
  \beta_{\mathrm{eff}}\opt(v)
  \;=\; \frac{V''(v)}{-V'(v)}
  \;=\; -\frac{\dd\ln\nu(e^v)}{\dd v}
  \;=\; \frac{1}{m(v)} - 1, \nonumber\\
  m'(v)
  &\;=\; \sum_{k=1}^K \rho_k D\frac{\dd w_k\opt(D)}{\dd D}
  \;=\; \sum_{k=1}^K \rho_k\left(\frac{\rho_k}{m(v)} - 1\right)w_k\opt(D)
  \;=\; \frac{\Var_{\PP_D\opt}(\rho_I)}{m(v)} \;\ge\; 0, \nonumber\\
  \frac{\dd\beta_{\mathrm{eff}}\opt(v)}{\dd v}
  &\;=\; -\frac{m'(v)}{m(v)^2}
  \;=\; -\frac{\Var_{\PP_D\opt}(\rho_I)}{m(v)^3} \;\le\; 0, \nonumber\\
  -\frac{V'''(v)}{V''(v)}
  &\;=\; \beta_{\mathrm{eff}}\opt(v) - \frac{\dd\ln\beta_{\mathrm{eff}}\opt(v)}{\dd v}
  \;=\; \beta_{\mathrm{eff}}\opt(v) + \frac{\Var_{\PP_D\opt}(\rho_I)}{m(v)^2\bigl(1 - m(v)\bigr)},
\end{align}
which establishes~\eqref{eq:water-filling-drift-identity}. Substituting $-U'''(u)/U''(u) = \alpha_0$ and~\eqref{eq:water-filling-drift-identity} into~\eqref{eq:K-third-derivative} along $(u\opt(t), v\opt(t))$ cancels the first-order terms:
\begin{gather}
  \eta(t)\alpha_0 - \bigl(1 - \eta(t)\bigr)\beta_{\mathrm{eff}}\opt\bigl(v\opt(t)\bigr) \;=\; 0, \nonumber\\
  \mc{K}(t) \;=\; -\bigl(1 - \eta(t)\bigr)\frac{\Var_{\PP_{e^{v\opt(t)}}\opt}(\rho_I)}{m\bigl(v\opt(t)\bigr)^2\bigl(1 - m(v\opt(t))\bigr)} \;\le\; 0,
\end{gather}
which establishes~\eqref{eq:water-filling-K-drift}.
\end{proof}

\subsection{Proof of Corollary~\ref{cor:logistic-transition} (Two-Stratum Logistic Transition Law)}
\label{app:proof:cor:logistic-transition}

\begin{proof}[Proof of \Cref{cor:logistic-transition} \StatementRef{cor:logistic-transition}]
For a two-stratum transition from the small-scale exponent $\eta_2$ to the large-scale exponent $\eta_1 > \eta_2$ with constant reference transition rate $\gamma > 0$, the normalized reference coordinate $s_{\mathrm{logistic}}(\tau) \in (0, 1)$ satisfies the autonomous logistic initial-value problem:
\begin{equation}
  s_{\mathrm{logistic}}(\tau) \;\Let\; \frac{\eta_{\mathrm{logistic}}(\tau) - \eta_2}{\eta_1 - \eta_2} \in (0, 1),
  \quad
  \frac{\dd s_{\mathrm{logistic}}}{\dd\tau} = \gamma s_{\mathrm{logistic}}(1 - s_{\mathrm{logistic}}),
  \quad
  s_{\mathrm{logistic}}(0) = \frac{1}{2}.
\end{equation}
Separating variables and integrating from $\tilde{\tau} = 0$ to $\tau = t - t_\star$ yields
\begin{align}
  \int_{1/2}^{s_{\mathrm{logistic}}(\tau)} \left(\frac{1}{s} + \frac{1}{1 - s}\right)\dd s
  &\;=\; \ln\frac{s_{\mathrm{logistic}}(\tau)}{1 - s_{\mathrm{logistic}}(\tau)}
  \;=\; \gamma\tau
  \quad\Longleftrightarrow\quad
  s_{\mathrm{logistic}}(\tau) \;=\; \sigma(\gamma\tau), \nonumber\\
  \eta_{\mathrm{logistic}}(\tau)
  &\;=\; \eta_2 + (\eta_1 - \eta_2)\sigma(\gamma\tau), \nonumber\\
  \max_{\tau \in \R}\abs{\eta_{\mathrm{logistic}}'(\tau)}
  &\;=\; \gamma\abs{\eta_1 - \eta_2}\max_{s \in (0, 1)} s(1 - s)
  \;=\; \frac{\gamma}{4}\abs{\eta_1 - \eta_2},
\end{align}
establishing~\eqref{eq:logistic-drift-law}. Furthermore, for the exact two-mode potential $U(u) = A_1 e^{-\alpha_1 u} + A_2 e^{-\alpha_2 u}$ ($0 < \alpha_1 < \alpha_2$) coupled with $V(v) = B e^{-\beta v}$, let $p(t) \Let \PP_t(2) \in (0, 1)$ denote the Gibbs weight of $\alpha_2$, so that $\alpha_{\mathrm{eff}}(t) = (1 - p(t))\alpha_1 + p(t)\alpha_2$ and $\eta(t) = \frac{\beta}{\alpha_{\mathrm{eff}}(t) + \beta}$. Expressing $s(t) = \frac{\eta(t) - \eta_2}{\eta_1 - \eta_2}$ in terms of $p(t)$ yields the exact M\"obius relation and variable-rate log-odds derivative
\begin{gather}
  \frac{s(t)}{1 - s(t)}
  \;=\; \frac{\eta(t) - \eta_2}{\eta_1 - \eta(t)}
  \;=\; \frac{\alpha_1 + \beta}{\alpha_2 + \beta}\,\frac{1 - p(t)}{p(t)}
  \;=\; \frac{A_1\alpha_1(\alpha_1 + \beta)}{A_2\alpha_2(\alpha_2 + \beta)}\,e^{(\alpha_2 - \alpha_1)u\opt(t)}, \nonumber\\
  \frac{\dd}{\dd t}\ln\!\left(\frac{s(t)}{1 - s(t)}\right)
  \;=\; (\alpha_2 - \alpha_1)\,\frac{\dd u\opt(t)}{\dd t}
  \;=\; (\alpha_2 - \alpha_1)\eta(t)
  \;=\; \frac{\dd z_{\mathrm{int}}(t)}{\dd t} \;>\; 0.
\end{gather}
Integrating from $t_\star$ (where $s(t_\star) = 1/2$, $z_{\mathrm{int}}(t_\star) = 0$, and $\eta(t_\star) = \bar{\eta} = \frac{\eta_1 + \eta_2}{2}$) to $t$ immediately gives $s(t) = \sigma(z_{\mathrm{int}}(t))$, and comparing $\ln\frac{s(t)}{1 - s(t)} = (\alpha_2 - \alpha_1)\int_{t_\star}^t \eta(r)\dd r$ against the constant-rate linear phase $\gamma(t - t_\star)$ via $\sup_{z \in \R}\sigma'(z) = \frac{1}{4}$ and $\sup_{r \in \R}\abs{\eta(r) - \bar{\eta}} \le \frac{1}{2}\abs{\eta_1 - \eta_2}$ gives
\begin{align}
  \left\lvert \eta(t) - \eta_{\mathrm{logistic}}(t - t_\star) \right\rvert
  &\;=\; \abs{\eta_1 - \eta_2}\,\left\lvert \sigma\!\left((\alpha_2 - \alpha_1)\!\int_{t_\star}^t \eta(r)\dd r\right) - \sigma\!\left((\alpha_2 - \alpha_1)\bar{\eta}(t - t_\star)\right) \right\rvert \nonumber\\
  &\;\le\; \frac{1}{4}(\alpha_2 - \alpha_1)\abs{\eta_1 - \eta_2}\int_{\min\{t, t_\star\}}^{\max\{t, t_\star\}} \abs{\eta(r) - \bar{\eta}}\dd r \nonumber\\
  &\;\le\; \frac{1}{8}(\alpha_2 - \alpha_1)\abs{\eta_1 - \eta_2}^2\abs{t - t_\star},
\end{align}
which establishes~\eqref{eq:logistic-error-bound}.
\end{proof}

\subsection{Proof of Theorem~\ref{thm:master-envelope} (Schur-Complement Hyperparameter Envelope)}
\label{app:proof:thm:master-envelope}

\begin{proof}[Proof of \Cref{thm:master-envelope} \StatementRef{thm:master-envelope}]
For each $\bm{r} = (u, v)^{\top} \in \R^2$, the interior minimizer $\xi\opt(\bm{r}) \in \R^p$ satisfies the stationarity identity $\nabla_\xi\bar{\mc{E}}(\bm{r}, \xi\opt(\bm{r})) = 0$ with invertible Hessian $\bm{H}_{\xi\xi}(\bm{r}) \succ 0$. By the implicit function theorem, $\xi\opt \in C^2(\R^2; \R^p)$, and differentiating the stationarity identity and the envelope value $\mc{E}(\bm{r}) = \bar{\mc{E}}(\bm{r}, \xi\opt(\bm{r}))$ with respect to $\bm{r}$ yields
\begin{align}
  \bm{H}_{\xi r}(\bm{r}) + \bm{H}_{\xi\xi}(\bm{r})\bm{J}_\xi(\bm{r}) &\;=\; \bm{0}
  \quad\Longleftrightarrow\quad
  \bm{J}_\xi(\bm{r}) \;\Let\; \frac{\partial\xi\opt(\bm{r})}{\partial \bm{r}^{\top}} \;=\; -\bm{H}_{\xi\xi}(\bm{r})^{-1}\bm{H}_{\xi r}(\bm{r}), \nonumber\\
  \nabla_{\bm{r}}\mc{E}(\bm{r})
  &\;=\; \nabla_{\bm{r}}\bar{\mc{E}}(\bm{r}, \xi\opt(\bm{r})) + \bm{J}_\xi(\bm{r})^{\top}\underbrace{\nabla_\xi\bar{\mc{E}}(\bm{r}, \xi\opt(\bm{r}))}_{=\,\bm{0}}
  \;=\; \nabla_{\bm{r}}\bar{\mc{E}}(\bm{r}, \xi\opt(\bm{r})).
  \label{eq:app-Jxi-ift}
\end{align}
Moreover, for any joint perturbation $(\dd \bm{r}, \dd\xi) \in \R^2 \times \R^p$ around $(\bm{r}, \xi\opt(\bm{r}))$, completing the square on the $2 \times 2$ block Hessian quadratic form gives the exact algebraic decomposition
\begin{align}
  \begin{bmatrix} \dd \bm{r} \\ \dd\xi \end{bmatrix}^{\top}
  \begin{bmatrix} \bm{H}_{rr}(\bm{r}) & \bm{H}_{r\xi}(\bm{r}) \\ \bm{H}_{\xi r}(\bm{r}) & \bm{H}_{\xi\xi}(\bm{r}) \end{bmatrix}
  \begin{bmatrix} \dd \bm{r} \\ \dd\xi \end{bmatrix}
  &\;=\; \dd \bm{r}^{\top}\!\left(\bm{H}_{rr}(\bm{r}) - \bm{\Delta}_S(\bm{r})\right)\dd \bm{r} \nonumber\\
  &\qquad + \norm{\bm{H}_{\xi\xi}(\bm{r})^{1/2}\left(\dd\xi - \bm{J}_\xi(\bm{r})\dd \bm{r}\right)}_2^2, \nonumber\\
  \bm{\Delta}_S(\bm{r})
  &\;\Let\; \bm{H}_{r\xi}(\bm{r})\bm{H}_{\xi\xi}(\bm{r})^{-1}\bm{H}_{\xi r}(\bm{r})
  \;=\; \bm{J}_\xi(\bm{r})^{\top}\bm{H}_{\xi\xi}(\bm{r})\bm{J}_\xi(\bm{r}) \;\succeq\; 0.
  \label{eq:app-schur-completing-square}
\end{align}
Along the optimal envelope manifold $\dd\xi = \bm{J}_\xi(\bm{r})\dd \bm{r}$, the second norm term in~\eqref{eq:app-schur-completing-square} vanishes identically, proving $\nabla_{rr}^2\mc{E}(\bm{r}) = \bm{H}_{rr}(\bm{r}) - \bm{\Delta}_S(\bm{r})$ in~\eqref{eq:schur-envelope}. Conversely, for a frozen anchor parameter $\xi_0 = \xi\opt(\bm{r}_0)$ with $\Delta\xi(\bm{r}) \Let \xi_0 - \xi\opt(\bm{r})$, expanding $\bar{\mc{E}}(\bm{r}, \xi_0)$ in $\xi$ around $\xi\opt(\bm{r})$ via Taylor's theorem with $\nabla_\xi\bar{\mc{E}}(\bm{r}, \xi\opt(\bm{r})) = \bm{0}$ and evaluating the Hessian at $\bm{r} = \bm{r}_0$ (where $\Delta\xi(\bm{r}_0) = \bm{0}$ and $\partial\Delta\xi/\partial\bm{r}^{\top} = -\bm{J}_\xi(\bm{r}_0)$) yields
\begin{align}
  \bar{\mc{E}}(\bm{r}, \xi_0) - \mc{E}(\bm{r})
  &\;=\; \frac{1}{2}\Delta\xi(\bm{r})^{\top}\bm{H}_{\xi\xi}(\bm{r})\Delta\xi(\bm{r}) + R_{\mathrm{env}}(\bm{r}, \xi_0), \nonumber\\
  \nabla_{rr}^2\bigl[\bar{\mc{E}}(\bm{r}, \xi_0) - \mc{E}(\bm{r})\bigr]\Big|_{\bm{r} = \bm{r}_0}
  &\;=\; \bm{J}_\xi(\bm{r}_0)^{\top}\bm{H}_{\xi\xi}(\bm{r}_0)\bm{J}_\xi(\bm{r}_0)
  \;=\; \bm{\Delta}_S(\bm{r}_0)
  \;=\; \begin{bmatrix} \Delta_{uu} & \Delta_{uv} \\ \Delta_{uv} & \Delta_{vv} \end{bmatrix} \;\succeq\; 0, \nonumber\\
  \eta_{\mathrm{frozen}}(\bm{r}_0) - \eta(\bm{r}_0)
  &\;=\; \frac{\mc{E}_{vv} - \mc{E}_{uv} + \Delta_{vv} - \Delta_{uv}}{\mc{H}_{\mathrm{iso}}(\bm{r}_0) + \Delta_{uu} - 2\Delta_{uv} + \Delta_{vv}} - \eta(\bm{r}_0) \nonumber\\
  &\;=\; \frac{\bigl(1 - \eta(\bm{r}_0)\bigr)(\Delta_{vv} - \Delta_{uv}) - \eta(\bm{r}_0)(\Delta_{uu} - \Delta_{uv})}{\mc{H}_{\mathrm{iso}}(\bm{r}_0) + \Delta_{uu} - 2\Delta_{uv} + \Delta_{vv}},
\end{align}
completing the proof of \Cref{thm:master-envelope}.
\end{proof}

\subsection{Proof of Corollary~\ref{cor:allocation-regret} (Quadratic Allocation Regret)}
\label{app:proof:cor:allocation-regret}

\begin{proof}[Proof of \Cref{cor:allocation-regret} \StatementRef{cor:allocation-regret}]
Along the iso-compute slice $s \mapsto (u\opt(t) + s, v\opt(t) - s) \in \mc{M}_t$, differentiating $\phi_t(s) = \mc{E}(u\opt(t) + s, v\opt(t) - s)$ once and twice and applying Taylor's theorem with integral remainder for any $\delta u \in [-\Delta, \Delta]$ gives
\begin{align}
  \phi_t'(0) &\;=\; \mc{E}_u\bigl(u\opt(t), v\opt(t)\bigr) - \mc{E}_v\bigl(u\opt(t), v\opt(t)\bigr) \;=\; 0, \nonumber\\
  \phi_t''(s) &\;=\; \mc{H}_{\mathrm{iso}}\bigl(u\opt(t) + s, v\opt(t) - s\bigr) \;>\; 0, \nonumber\\
  \frac{1}{2}\,\underline{\mc{H}}_{t,\Delta}\,(\delta u)^2
  &\;\le\; \phi_t(\delta u) - \phi_t(0)
  \;=\; \int_0^{\delta u} (\delta u - s)\,\phi_t''(s)\dd s
  \;\le\; \frac{1}{2}\,\overline{\mc{H}}_{t,\Delta}\,(\delta u)^2,
\end{align}
establishing~\eqref{eq:quadratic-regret-sandwich}. For the separable power-law envelope $\mc{E}(u, v) = c_1 e^{-au} + c_2 e^{-bv}$, let $M_0(t) > 0$ denote the common marginal gradient at the optimum $\phi_t'(0) = 0$. Expressing both $\mc{E}\opt(t)$ and $\phi_t''(0)$ in terms of $M_0(t)$ gives
\begin{align}
  M_0(t) &\;\Let\; a c_1 e^{-au\opt(t)} \;=\; b c_2 e^{-bv\opt(t)} \;>\; 0, \nonumber\\
  \mc{E}\opt(t)
  &\;=\; c_1 e^{-au\opt(t)} + c_2 e^{-bv\opt(t)}
  \;=\; \left(\frac{1}{a} + \frac{1}{b}\right)M_0(t)
  \;=\; \frac{a + b}{ab}\,M_0(t), \nonumber\\
  \phi_t''(0)
  &\;=\; a^2 c_1 e^{-au\opt(t)} + b^2 c_2 e^{-bv\opt(t)}
  \;=\; (a + b)\,M_0(t)
  \;=\; ab\,\mc{E}\opt(t),
\end{align}
completing the proof of \Cref{cor:allocation-regret}.
\end{proof}

\subsection{Proof of Proposition~\ref{prop:surrogate-decision-regret} (Surrogate-to-Decision Allocation Regret Stability)}
\label{app:proof:prop:surrogate-decision-regret}

\begin{proof}[Proof of \Cref{prop:surrogate-decision-regret} \StatementRef{prop:surrogate-decision-regret}]
When a composite slice surrogate $\widehat{\phi}_t(s)$ approximates the exact slice objective $\phi_t(s)$ on $[-\Delta, \Delta]$, the triangle inequality decomposes the uniform slice derivative error into its three component bounds:
\begin{align}
  \widehat{\phi}_t(s) &\;=\; \widehat{U}\bigl(u\opt(t)+s\bigr) + \widehat{V}\bigl(v\opt(t)-s\bigr) + \widehat{R}_{\mathrm{env},t}(s), \nonumber\\
  \phi_t(s) &\;=\; U\bigl(u\opt(t)+s\bigr) + V\bigl(v\opt(t)-s\bigr) + R_{\mathrm{env},t}(s), \nonumber\\
  \sup_{\abs{s} \le \Delta}\abs{\widehat{\phi}_t'(s) - \phi_t'(s)}
  &\;\le\; \sup_{\abs{s} \le \Delta}\abs{\widehat{U}'(u\opt(t)+s) - U'(u\opt(t)+s)} \nonumber\\
  &\qquad + \sup_{\abs{s} \le \Delta}\abs{\widehat{V}'(v\opt(t)-s) - V'(v\opt(t)-s)} \nonumber\\
  &\qquad + \sup_{\abs{s} \le \Delta}\abs{\widehat{R}_{\mathrm{env},t}'(s) - R_{\mathrm{env},t}'(s)} \nonumber\\
  &\;\le\; \varepsilon_{\mathrm{spec}}(t) + \varepsilon_{\mathrm{mix}}(t) + \varepsilon_{\mathrm{env}}(t) \;\Let\; \varepsilon_t.
\end{align}
Under $\sup_{\abs{s} \le \Delta}\abs{\widehat{\phi}_t'(s) - \phi_t'(s)} \le \varepsilon_t$ and $\abs{\widehat{\phi}_t'(\delta\hat{u}_t)} \le \tau_t$, the triangle inequality gives $\abs{\phi_t'(\delta\hat{u}_t)} \le \varepsilon_t + \tau_t$. By strong convexity $\phi_t''(s) \ge \underline{\mc{H}}_{t,\Delta}$ on $[-\Delta, \Delta]$ and $\phi_t'(0) = 0$, the mean-value theorem and the Taylor expansion of $\phi_t(0)$ around $\delta\hat{u}_t$ yield
\begin{align}
  \underline{\mc{H}}_{t,\Delta}\abs{\delta\hat{u}_t}
  &\;\le\; \abs{\phi_t'(\delta\hat{u}_t) - \phi_t'(0)}
  \;\le\; \varepsilon_t + \tau_t
  \quad\Longrightarrow\quad
  \abs{\delta\hat{u}_t} \;\le\; \frac{\varepsilon_t + \tau_t}{\underline{\mc{H}}_{t,\Delta}}, \nonumber\\
  \phi_t(\delta\hat{u}_t) - \phi_t(0)
  &\;\le\; \phi_t'(\delta\hat{u}_t)\,\delta\hat{u}_t - \frac{1}{2}\,\underline{\mc{H}}_{t,\Delta}(\delta\hat{u}_t)^2
  \;\le\; (\varepsilon_t + \tau_t)\abs{\delta\hat{u}_t} - \frac{1}{2}\,\underline{\mc{H}}_{t,\Delta}\abs{\delta\hat{u}_t}^2 \nonumber\\
  &\;\le\; \frac{(\varepsilon_t + \tau_t)^2}{2\,\underline{\mc{H}}_{t,\Delta}}
  \;\le\; \frac{\overline{\mc{H}}_{t,\Delta}}{2\,\underline{\mc{H}}_{t,\Delta}^2}(\varepsilon_t + \tau_t)^2,
\end{align}
establishing~\eqref{eq:end-to-end-regret-bound}. Under the zeroth-order condition $\sup_{\abs{s} \le \Delta}\abs{\widehat{\phi}_t(s) - \phi_t(s)} \le \frac{1}{2}\varepsilon_{0,t}$ and $\widehat{\phi}_t(\delta\hat{u}_t) \le \min_{\abs{s} \le \Delta}\widehat{\phi}_t(s) + \tau_{0,t}$, combining~\eqref{eq:quadratic-regret-sandwich} with standard two-sided value comparison gives
\begin{align}
  \frac{1}{2}\,\underline{\mc{H}}_{t,\Delta}(\delta\hat{u}_t)^2
  &\;\le\; \phi_t(\delta\hat{u}_t) - \phi_t(0) \nonumber\\
  &\;\le\; \bigl(\phi_t(\delta\hat{u}_t) - \widehat{\phi}_t(\delta\hat{u}_t)\bigr) + \bigl(\widehat{\phi}_t(\delta\hat{u}_t) - \widehat{\phi}_t(0)\bigr) + \bigl(\widehat{\phi}_t(0) - \phi_t(0)\bigr) \nonumber\\
  &\;\le\; \frac{1}{2}\varepsilon_{0,t} + \tau_{0,t} + \frac{1}{2}\varepsilon_{0,t}
  \;=\; \varepsilon_{0,t} + \tau_{0,t},
\end{align}
which immediately yields~\eqref{eq:end-to-end-zeroth-order-bound} and completes the proof of \Cref{prop:surrogate-decision-regret}.
\end{proof}

\vskip 0.2in
\bibliography{main}

\end{document}